\documentclass[]{article}
\usepackage{geometry}
\usepackage{amsmath}     
\usepackage{authblk}
\usepackage{amssymb}     
\usepackage{mathtools}   
\usepackage{algorithm}
\usepackage{algorithmic}
\usepackage{orcidlink}
\usepackage{float}       
\usepackage{threeparttable} 
\usepackage{graphicx}    
\usepackage{adjustbox}   
\usepackage{multirow}    
\usepackage{appendix}
\usepackage{subcaption}
\usepackage{hyperref}
\usepackage{xcolor}
\usepackage{booktabs}
\usepackage{placeins}
\usepackage{makecell}

\newcommand{\omegaVector}{\boldsymbol{\omega}}

\title{An Adaptive Random Fourier Features approach Applied to Learning Stochastic Differential Equations}
\author[1]{Owen Douglas}
\author[1]{Aku Kammonen\,\orcidlink{0000-0002-8458-0852}}
\author[2]{Anamika Pandey\,\orcidlink{0000-0001-8644-8540}}
\author[1,2,3]{Ra\'{u}l~Tempone\,\orcidlink{0000-0003-1967-4446}}
\affil[1]{Computer, Electrical and Mathematical Sciences and Engineering,
	4700 King Abdullah University of Science and Technology (KAUST),
	Thuwal 23955-6900, Kingdom of Saudi Arabia.}
\affil[2]{Chair of Mathematics for Uncertainty Quantification, RWTH Aachen University, 52062 Aachen, Germany.} 
\affil[3]{Alexander von Humboldt Professor in Mathematics for Uncertainty Quantification, RWTH Aachen University, 52062 Aachen, Germany.}

\newcommand{\keywords}[1]{\par\noindent\textbf{Keywords:} #1}
\newcommand{\subjclass}[2][2020]{\par\noindent\textbf{Mathematics Subject Classification #1:} #2}
\date{\today}

\begin{document}

\maketitle

\begin{abstract}
This work proposes a training algorithm based on adaptive random Fourier features (ARFF) with Metropolis sampling and resampling \cite{kammonen2024adaptiverandomfourierfeatures} for learning drift and diffusion components of stochastic differential equations from snapshot data. Specifically, this study considers It\^{o} diffusion processes and a likelihood-based loss function derived from the Euler-Maruyama integration introduced in \cite{Dietrich2023} and \cite{dridi2021learningstochasticdynamicalsystems}. 

This work evaluates the proposed method against benchmark problems presented in \cite{Dietrich2023}, including polynomial examples, underdamped Langevin dynamics, a stochastic susceptible-infected-recovered model, and a stochastic wave equation. Across all cases, the ARFF-based approach matches or surpasses the performance of conventional Adam-based optimization in both loss minimization and convergence speed. These results highlight the potential of ARFF as a compelling alternative for data-driven modeling of stochastic dynamics.
\end{abstract}

\section{Introduction} 
The efficient identification of  dynamical systems from data is a fundamental challenge in many scientific and engineering domains. Classical parameter estimation techniques for stochastic differential equations (SDEs) - including maximum likelihood estimation, the method of moments, and Bayesian inference \cite{BerntSDE}, \cite{Sorensen2009}, have widespread applications in physics \cite{risken1996fokker}, \cite{vankampen2007stochastic}, finance \cite{Yacine_finance}, \cite{james2000interest} and biology \cite{Schwartz_biology}. Despite their utility, these methods impose strong model assumptions, demand substantial analytical effort, and often become computationally intractable for complex or high-dimensional systems.

Recent advances in machine learning have offer new options for data-driven modelling of dynamical systems \cite{RAISSI2019}. Deep learning frameworks, such as residual networks, neural ordinary differential equations \cite{neural_ODEs}, and neural partial differential equations (PDEs) \cite{pde-net, Karniadakis_PINNs}, demonstrate significant promise in approximating complex dynamical systems. The success of these approaches has increased interest in learning stochastic dynamics from data. 

Several recent studies have investigated the application of neural networks to approximate the drift and diffusion terms of SDEs. Notable contributions include neural SDEs \cite{neural_SDEs}, which model stochastic processes using deep latent Gaussian models, where drift and diffusion functions are parametrized via neural networks. Another work \cite{SDE_Net} proposed SDE-Net, that explicitly integrates neural networks to model drift dynamics while capturing uncertainty via a learned diffusion process. Furthermore, additional studies \cite{Dietrich2023} and \cite{dridi2021learningstochasticdynamicalsystems} have trained drift and diffusion networks that minimize the likelihood-based loss functions derived from the Euler-Maruyama (EM) scheme.

These approaches highlight the increasing potential of deep learning in data-driven modelling of stochastic systems. However, these methods rely on gradient-based optimization techniques, which can be computationally expensive and suffer from slow convergence in high-dimensional spaces. Moreover, the choice of network architectures and training schemes can significantly influence the generalizability of these models.

Random Fourier features (RFF) \cite{rahimi_recht} offer a scalable approach to approximating kernel-based methods, allowing efficient complex-function learning. Unlike traditional deep learning approaches, RFF methods sample feature weights from a predefined distribution, typically Gaussian, and optimize only the remaining parameters. 

Adaptive extensions of RFF \cite{kammonen2020adaptiverandomfourierfeatures, kammonen2024adaptiverandomfourierfeatures, kammonen2021smallergeneralizationerrorderived} dynamically adjust the sampling distribution to improve the approximation quality and convergence speed. These methods have found success solving regression and classification problems and approximating solutions to PDEs. Given these advantages, adaptive RFF methods offer a compelling prospect the learning of stochastic systems from data.

\vspace{1em}

Section \ref{sec:ARFF algorithm} introduces an adaptive RFF (ARFF) algorithm with Metropolis sampling and resampling, building on the work in \cite{kammonen2024adaptiverandomfourierfeatures}. Section \ref{sec:mathprobset} formulates the mathematical problem setting, deriving the loss minimisation objective from the EM discretization of the target SDE, a time-independent It\^{o} diffusion process. This section also details the structure of the networks $f_\theta$ and $\Sigma_{\theta'}$ employed to approximate the drift and diffusion dynamics respectively. Section \ref{sec:training_algorithm} presents the proposed novel SDE training algorithm, describing how the ARFF training procedure is implemented to learn the drift and diffusion dynamics. Next, Section \ref{sec:computational_experiments} establishes the experimental framework, describing the benchmark datasets employed to evaluate the proposed approach against existing Adam-based optimization. Section \ref{sec:results} compiles the results of the experiments, commenting on any significant findings. Section \ref{sec:further_comparisions} introduces Finally, Section \ref{sec:summary} summarizes the results, highlighting the benefits and acknowledging the limitations of the proposed algorithm in the context of data-driven SDE learning.

\section{Adaptive Random Fourier Features Training Algorithm}\label{sec:ARFF algorithm}
The ARFF algorithm with Metropolis sampling was first introduced in \cite{kammonen2020adaptiverandomfourierfeatures} and was later combined with a resampling step in \cite{kammonen2024adaptiverandomfourierfeatures} for stability with respect to parameter choices. 

Consider the data $(x_n, y_n)\in\mathbb{R}^D\times\mathbb{R}^{D'}$, $n=1,\dots,N$ where $x_n$ is a random variable from an unknown probability density function $\hat\rho$ and assumes that an underlying function $g:\mathbb{R}^D\to\mathbb{R}^{D'}$ exists such that $g(x_n)=y_n, n=1,\dots,N$. Then the ARFF algorithm aims to solve
\begin{equation}
\label{eq:regularized_loss}
\min_{\beta \in \mathcal{N}_K} \left\{ \mathbb{E}_{\hat{\rho}}[|y_n - \beta(x_n)|^2] + \lambda \sum_{k=1}^K |\hat{\beta}_k|^2 \right\},
\end{equation}
where
\begin{equation*}
\label{eq:network_def}
\mathcal{N}_K := \left\{ \beta(x) = \sum_{k=1}^K \hat{\beta}_k e^{\mathrm{i}\omega_k\cdot x} \right\},
\end{equation*}
and $\lambda>0$ is a Tikhonov parameter.

In \cite{kammonen2020adaptiverandomfourierfeatures}, following \cite{UASig-Barron-IEEE-1993}, the authors derive the bound
\begin{equation*}
\min_{\beta \in \mathcal{N}_K} \left\{ \mathbb{E}_{\hat{\rho}}[|y_n - \beta(x_n)|^2] + \lambda \sum_{k=1}^K |\hat{\beta}_k|^2 \right\} \leq \frac{1+\lambda}{K}\mathbb{E}_p\left[\frac{|\hat{g}(\omega)|^2}{(2\pi)^Dp^2(\omega)}\right],
\end{equation*}
where $\omega\sim p$ for some probability density function $p$ and $\hat{g}$ denotes the Fourier transform of the function $g$.

Further they show that
\begin{equation*}
    p_*(\omega) := p(\omega) = \frac{|\hat g(\omega)|}{||\hat g||_{L^1(\mathbb{R}^D)}}
\end{equation*}
is the optimal distribution to sample from in the sense that it minimizes the constant $\mathbb{E}_p\left[\frac{|\hat{g}(\omega)|^2}{(2\pi)^Dp^2(\omega)}\right]$.

The ARFF algorithm aims to asymptotically sample from $p_*$ and for fixed samples $\omega_k, k =1,\dots,K$ the problem \eqref{eq:regularized_loss} reduces to a linear least squares problem in $\hat\beta_k, k=1,\dots,K$ which has the normal equations
\begin{equation}
    \label{eq:normal_equations_ARFF}
    (S^HS + \lambda N I)\hat{\beta} = S^Hy,
\end{equation}
where $S\in \mathbb{C}^{N\times K}$ has the components $S_{n,k} = e^{\mathrm{i}\omega_k\cdot x_n}, n=1,\dots,N, k=1,\dots,K$, $y = [y_1,\dots,y_n]^T$, and $I\in \mathbb{R}^{K\times K}$ is the identity matrix.

The ARFF training procedure we use, outlined in Algorithm \ref{alg:ARFF}, is adapted from Algorithm 4 in \cite{kammonen2024adaptiverandomfourierfeatures}, with a few key differences: i) an early stopping criterion is included based on validation error stagnation, ii) the computation of the effective sample size is omitted, and iii) resampling and the Metropolis test are always performed.

\begin{algorithm}[ht!]
  \caption{Adaptive random Fourier features with Metropolis sampling and resampling}
  \label{alg:ARFF}
  \begin{algorithmic}
    \STATE {\bfseries Input:} $\{(x_n, y_n)\}_{n=1}^{N}$
    \STATE {\bfseries Output:} $x \mapsto \sum_{k=1}^{K} \hat{\beta}_k e^{i \omega_k \cdot x}$ 
    \STATE {\bfseries Algorithm Parameters:} 
      \STATE \hspace{\algorithmicindent} $M_{\text{max}}$ \hspace{\algorithmicindent} (maximum number of iterations),
      \STATE \hspace{\algorithmicindent} $\delta$ \hspace{\algorithmicindent} (proposal step length),
      \STATE \hspace{\algorithmicindent} $\gamma$ \hspace{\algorithmicindent} (exponent in the Metropolis test),
      \STATE \hspace{\algorithmicindent} $\lambda$ \hspace{\algorithmicindent} (Tikhonov parameter)
    \STATE~ 
    \STATE $\omega \gets$ zero vector in $\mathbb{R}^{K d}$
    \STATE $\boldsymbol{\hat{\beta}} \gets$ minimizer of problem (\ref{eq:normal_equations_ARFF}) given $\omega$
    
    \FOR{$i = 1$ {\bfseries to} $M_{\text{max}}$}
    \IF{resampling}
      \STATE $\mathbf{\check{p}} \gets |\boldsymbol{\hat{\beta}}| / \sum_{k=1}^{K} |\hat{\beta}_k|$
        \STATE $\mathbf{j} \gets \text{vector of $K$ independent samples from $1,\dots,K$ with PMF $\mathbf{\check{p}}$}$
        \STATE $\omegaVector_{1,\dots,K} \gets \omegaVector_\mathbf{j}$

      \ENDIF
      \STATE $r_N \gets$ standard normal random vector in $\mathbb{R}^{K d}$
      \STATE $\omega^* \gets \omega + \delta r_N$ \hspace{\algorithmicindent} \COMMENT{random walk Metropolis proposal}
      \STATE $\boldsymbol{\hat{\beta}^*} \gets$ minimizer of problem (\ref{eq:normal_equations_ARFF}) given $\omega^*$
      
      \FOR{$k = 1$ {\bfseries to} $K$}
        \STATE $r_U \gets$ sample from uniform distribution on $[0,1]$
        \IF{$|\hat{\beta}_k^*| / |\hat{\beta}_k|^\gamma > r_U$ {\COMMENT{Metropolis test}}} 
        \STATE \hspace{\algorithmicindent} $\omega_k \gets \omega_k^*$
        \ENDIF
      \ENDFOR
      
      \STATE $\boldsymbol{\hat{\beta}}\gets$ minimizer of problem (\ref{eq:normal_equations_ARFF}) with adaptive $\omega$ 
      \STATE
      \IF{validation error stagnates\textsuperscript{1}}
      \STATE \textbf{break}
      \ENDIF
    \ENDFOR

    \STATE $x \mapsto \sum_{k=1}^{K} \hat{\beta}_k e^{i \omega_k \cdot x}$

  \end{algorithmic}
  \vspace{0.5cm}
  \footnotesize{\textsuperscript{1}The validation error is computed as the mean squared error between the predicted outputs and corresponding target values, using a subset of the input data reserved for validation. Stagnation occurs when the moving average of the five most recent validation errors fails to decrease over five consecutive iterations.}
\end{algorithm}

\section{Mathematical Problem Setting }\label{sec:mathprobset} 

This work aims to identify an effective SDE that approximates the coarse-grained dynamics of a system based on snapshot data. The assumed form of the SDE is an Itô diffusion process
\begin{equation}
\label{eq:ito diffusion SDE}
    \mathrm{d}x_t = f(x_t) \, \mathrm{d}t + \sigma(x_t) \, \mathrm{d}W_t,
\end{equation}
where $x_t \in \mathbb{R}^D$ is the system state at time $t$, $f: \mathbb{R}^D \to \mathbb{R}^D$ is the drift function, $\sigma: \mathbb{R}^D \to \mathbb{R}^{D \times D}$ is the diffusion function. These functions are assumed to satisfy the appropriate regularity condition (see Chapter 5 in \cite{BerntSDE}), that ensure the existence of a solution to Eq. (\eqref{eq:ito diffusion SDE}). The process $W_t$ is a D-dimensional Wiener with independent increments, such that $W_{t} - W_{s} \sim \mathcal{N}(0, t-s)$ for $0\leq s < t$.

Training data comprises of short-term observations of the system, in the form
\begin{equation*}
\label{eq:dataset}
    \mathcal{D} = \{(x_0^{(n)}, x_1^{(n)}, h^{(n)})\}_{n=1}^N,
\end{equation*}
where $x_0^{(n)}$ is the initial point and $x_1^{(n)}$ is the evolved state after time $h^{(n)} > 0$.

\subsection*{Euler-Maruyama Approximation and Transition Density}

Using the EM scheme, we approximate the forward evolution of the SDE over a short time $h$ by
\begin{equation}
\label{eq:euler maruyama}
    x_1 \approx x_0 + h f(x_0) + \sigma(x_0) \, \delta W, \quad \delta W \sim \mathcal{N}(0, h I_D).
\end{equation}
Thus, the conditional distribution of $x_1$ given $x_0$ and $h$ is approximated as
\begin{equation*}
    x_1 \mid x_0 \sim \mathcal{N}(x_0 + h f(x_0),\, h\Sigma(x_0)), \quad \text{where } \Sigma(x_0) = \sigma(x_0)\sigma(x_0)^\top.
\end{equation*}

This gives a tractable likelihood model based on numerical simulation suitable for statistical learning.

\subsection*{Learning Objective}

This work introduces two parametrized models. First, the drift model \( f_\theta: \mathbb{R}^D \to \mathbb{R}^D \) is defined as
\begin{equation*}
    f_\theta(x) = \Re\left(\sum_{k=1}^K \hat{\beta}_k e^{\mathrm{i} \omega_k \cdot x}\right),
\end{equation*}
with \( \hat{\beta}_k \in \mathbb{C}^D \) and \( \omega_k \in \mathbb{R}^D \). 

Second, the diffusion covariance model \( \Sigma_{\theta'}: \mathbb{R}^D \to \mathbb{R}^{D \times D} \) is defined as a symmetric matrix function with the lower-triangular components given by
\begin{equation*}
    \Big({\Sigma_{\theta'}(x)}\Big)_{(i,j)} = \left({\hat{\Sigma}_{\theta'}(x)}\right)_{(i(i-1)/2 + j)} \quad \text{for } i \in \{1, \dots, D\} \text{ and } j \in \{1, \dots, i\},
    \label{eq:diff_func_transform}
\end{equation*}
where \( \hat{\Sigma}_{\theta'}: \mathbb{R}^D \to \mathbb{R}^{D(D+1)/2} \) is defined as
\begin{equation*}
    \hat{\Sigma}_{\theta'}(x) = \Re\left(\sum_{k=1}^{K'} \hat{\beta}'_k e^{\mathrm{i} \omega'_k \cdot x}\right),
\end{equation*}
with \( \hat{\beta}'_k \in \mathbb{C}^{D(D+1)/2} \) and \( \omega'_k \in \mathbb{R}^D \).

The full parameter sets are
\begin{equation*}
    \theta \in \Theta = \Theta_{\hat{\beta}} \times \Theta_\omega \subseteq \mathbb{C}^{D \times K'} \times \mathbb{R}^{D \times K'},
\end{equation*}
\begin{equation*}
    \theta' \in \Theta' = \Theta_{\hat{\beta}'} \times \Theta_{\omega'} \subseteq \mathbb{C}^{D(D+1)/2 \times K} \times \mathbb{R}^{D \times K}.
\end{equation*}

Following \cite{Dietrich2023} and \cite{dridi2021learningstochasticdynamicalsystems}, the negative log-likelihood loss for a single observation is
\begin{equation}
\label{eq:loss}
    \mathcal{L}(\theta, \theta' \mid x_0, x_1, h) = 
    \frac{1}{2} (x_1 - x_0 - h f_\theta(x_0))^\top (h\Sigma_{\theta'}(x_0))^{-1} (x_1 - x_0 - h f_\theta(x_0)) 
    + \frac{1}{2} \log |h\Sigma_{\theta'}(x_0)| + \frac{D}{2} \log(2\pi).
\end{equation}

The total loss over the dataset is
\begin{equation}
    \mathcal{L}_{\text{total}}(\theta, \theta') = \mathbb{E}_{\hat{\rho}} \left[\mathcal{L}(\theta, \theta' \mid x_0, x_1, h)\right],
    \label{eq:total_loss}
\end{equation}
where $\hat{\rho}$ denotes the empirical density of the dataset $\mathcal{D}$.

The objective is to minimize the total loss (\ref{eq:total_loss}) w.r.t the models $f_\theta$ and $\Sigma_{\theta'}$. 

Minimizing the loss w.r.t the diffusion $\sigma$ is equivalent to minimizing it w.r.t the diffusion covariance $\Sigma$. However, since the mapping from $\sigma$ to $\Sigma$ is many-to-one, there can exist infinitely many solutions for $\sigma$ that minimise (\ref{eq:total_loss}). Hence, information about the structure of $\sigma$ is necessary to recover a unique solution. If the true diffusion is known to be symmetric, $\sigma$ can be calculated as the principal matrix square root of $\Sigma$. If it is lower-triangular, a Cholesky decomposition yields the solution. 

\section{Drift and Diffusion Training Algorithm}\label{sec:training_algorithm} 

This section describes how the ARFF training algorithm (Algorithm \ref{alg:ARFF}) is implemented to minimize the objective function (\ref{eq:total_loss}) w.r.t the drift and diffusion models $f_\theta$ and $\Sigma_{\theta'}$ given the data $\mathcal{D}$.

\subsection{Drift Training}
For a fixed diffusion model $\theta' = \tilde{\theta}'$, the objective function can be bounded by a minimisation over the drift network $f_\theta$ 
\begin{equation}
\begin{aligned}
    \min _{f_\theta, \Sigma_{\theta'}}\left\{ \mathcal{L}_{total}\left(\theta, \theta'\right) \right\} & \leq \min _{f_\theta \in \mathcal{N}_K}\left\{ \mathcal{L}_{total}\left(\theta, \theta'\right)\Big|_{\theta'=\tilde{\theta}'} \right\} \\
    &= \min _{f_\theta \in \mathcal{N}_K}\left\{ \mathbb{E}_{\hat{\rho}} \left[ \frac{1}{2}(x_1 - x_0 - h f_\theta(x_0))^\top (h\Sigma_{\theta'}(x_0))^{-1} (x_1 - x_0 - h f_\theta(x_0)) \right.\right. \\
    &\qquad \left.\left. + \frac{1}{2}\log\left| h\Sigma_{\theta'}(x_0) \right| \right] + \frac{D}{2}\log(2\pi) \right\} \\
    &= \min _{f_\theta \in \mathcal{N}_K} \frac{1}{2}\left\{ \mathbb{E}_{\hat{\rho}} \left[ \left| (h\Sigma_{\theta'}(x_0))^{-1/2} (x_1 - x_0 - h f_\theta(x_0)) \right|^2 \right] \right\} + C \\
    &\leq \frac{1}{2} \mathbb{E}_{\hat{\rho}} \left[ h\lVert \Sigma_{\theta'}(x_0)^{-1/2} \rVert^2 \right] \min _{f_\theta \in \mathcal{N}_K}\left\{ \mathbb{E}_{\hat{\rho}} \left[ \left| h^{-1}(x_1 - x_0) - f_\theta(x_0) \right|^2 \right] \right\} + C,
\end{aligned}
\label{eq:drift problem}
\end{equation}
where $C$ collects the terms independent of $f_\theta$ and $\lVert \cdot \rVert$ denotes the Frobenius norm.

Isolating the minimization problem and introducing a Tikhonov regularization parameter $\lambda \ge 0$, renders a least-square problem in the form of (\ref{eq:regularized_loss}),
\begin{equation}
    \min _{f_\theta \in \mathcal{N}_K}\left\{ \mathbb{E}_{\hat{\rho}} \left[ | h^{-1}(x_1 - x_0) - f_\theta(x_0) |^2\right] + \lambda \sum_{k=1}^K |\hat{\beta}_k|^2 \right\}.
    \label{eq:drift_minimisation_problem}
\end{equation}
The network $f_\theta$ is trained on the dataset $\{(x_0^{(n)}, h_{(n)}^{-1}(x_1^{(n)} - x_0^{(n)}))\}_{n=1}^N $ using Algorithm \ref{alg:ARFF}.

\subsection{Diffusion Training}

With the knowledge of the drift estimate $f_\theta$, the optimal diffusion covariance that minimizes the Gaussian log-likelihood loss (\ref{eq:loss}) for each observed triplet $\{(x_0^{(n)}, x_1^{(n)}, h^{(n)})\}_{n=1}^N$ is calculated. Then a network is trained using Algorithm \ref{alg:ARFF} that approximates these closed-form solutions. 

The pointwise minimization problems
\begin{equation}
    \min _{\Sigma_{(n)} \in \mathbb{R}^{D \times D}}\left\{ \frac{1}{2} (x_1^{(n)} - x_0^{(n)} - h_{(n)} f_\theta(x_0^{(n)}))^\top (h_{(n)}\Sigma_{(n)})^{-1} (x_1^{(n)} - x_0^{(n)} - h_{(n)} f_\theta(x_0^{(n)})) + \frac{1}{2} \log |h\Sigma_{(n)}| + \frac{D}{2} \log(2\pi) \right\},
    \label{eq:pointwise_diff_loss}
\end{equation}
have the solutions
\begin{equation}
    \Sigma^*_{(n)} = h_{(n)}^{-1}(x_1^{(n)} - x_0^{(n)} - h_{(n)} f_\theta(x_0^{(n)}))(x_1^{(n)} - x_0^{(n)} - h_{(n)} f_\theta(x_0^{(n)}))^\top, \quad \text{for } n=1,\dots,N. 
    \label{eq:pointwise_diff_minima}
\end{equation}
Algorithm \ref{alg:ARFF} only supports the learning of vector-to-vector functions. Since $\Sigma^*_{(n)}$ are symmetric matrices, the lower-triangular elements are vectorized into $\hat{\Sigma}^*_{(n)} \in \mathbb{R}^{D(D+1)/2}$ where 
\begin{equation}
    {\hat{\Sigma}^*} _{{(i(i-1)/2 + j)}} =  \Sigma^*_{(i,j)} \quad \text{for } i \in \{1, \dots, D\} \text{ and } j \in \{1, \dots, i\}.
    \label{eq:diff_transform}
\end{equation}
Then, the least-squares problem becomes
\begin{equation*}
    \min _{\hat{\Sigma}_{\theta'} \in \mathcal{N}_{K'}}\left\{ \mathbb{E}_{\hat{\rho}} \left[ | \hat{\Sigma}^* - \hat{\Sigma}_{\theta'}(x_0) |^2\right] + \lambda' \sum_{k=1}^{K'} |\hat{\beta}'_k|^2 \right\},
\end{equation*}
where $\lambda' \ge 0$ is a Tikhonov regularization parameter.

The network $\hat{\Sigma}_{\theta'}(x_0)$ is trained on the dataset $\{(x_0^{(n)}, \hat{\Sigma}^*_{(n)})\}_{n=1}^N $ using Algorithm \ref{alg:ARFF}. After training, the matrix-valued function $\Sigma_{\theta'}(x_0)$ is recovered by reversing the lower-triangular vectorization (\ref{eq:diff_func_transform}). 

Algorithm \ref{alg:ARFF_SDE} summarizes the drift and diffusion training procedures. 

\begin{algorithm}[ht!]
\caption{Adaptive random Fourier features training for drift and diffusion}\label{alg:ARFF_SDE}
\begin{algorithmic}
\STATE \textbf{Input:} $\{(x_1^{(n)}, x_0^{(n)}, h_{(n)})\}_{n=1}^N$
\STATE \textbf{Output:} Drift; $f_\theta(x_0)$, Diffusion; $\Sigma_{\theta'}(x_0)$
\STATE
\STATE $f_\theta(x_0) = \sum_{k=1}^{K} \hat{\beta}_k e^{i w_k \cdot x_0} \gets $ Algorithm \ref{alg:ARFF} with \textbf{Input:} $\{(x_0^{(n)}, h^{-1}_{(n)}( x_1^{(n)} - x_0^{(n)}))\}_{n=1}^N$. 
\STATE $\Sigma^*_{(n)} \gets$ Minimizer (\ref{eq:pointwise_diff_minima}) of problem (\ref{eq:pointwise_diff_loss}). 
\STATE $\hat{\Sigma}^*_{(n)} \gets$ Vectorize lower-triangular elements of $\Sigma^*_{(n)}$ (\ref{eq:diff_transform}).
\STATE $\hat{\Sigma}_{\theta'}(x_0) = \sum_{k=1}^{K'} \hat{\beta}'_k e^{i w'_k \cdot x_0} \gets $ Algorithm \ref{alg:ARFF} with \textbf{Input:} $\{(x_0^{(n)}, \hat{\Sigma}^*_{(n)})\}_{n=1}^N$. 
\STATE $\Sigma_{\theta'}(x_0) \gets $ Reshape $\hat{\Sigma}_{\theta'}(x_0)$ back into symmetric matrix (\ref{eq:diff_func_transform}). 

\end{algorithmic}
\end{algorithm}

The inputs to Algorithm \ref{alg:ARFF} are normalized and denormalized before and after each training sequence.

\section{Computational Experiments}\label{sec:computational_experiments}
This section describes a series of computational experiments that compare Algorithm \ref{alg:ARFF_SDE} and the Adam optimizer when learning SDEs. The specific target SDEs are taken directly from   \cite{Dietrich2023}. They include simple polynomial examples and a series of benchmarks from the literature including an underdamped Langevin equation, a Stochastic Susceptible Infected Recovered (SIR) model and a stochastic (wave) PDE (SPDE).

\subsection{Polynomial: Experiments 1 to 4} 

In these experiments, training datasets are generated using polynomial drift and diffusion functions in one, two and three dimensions. The drift and diffusion functions exist up to cubic and linear degrees respectively. Experiment 4 includes symmetric and lower-triangular (LT) positive-definite diffusion matrices.
Training datasets are generated according to Table \ref{tab:polynomial_training_data}. The initial distributions of the $x_0$ data points are uniform across the specified domains.  

\begin{table}[ht!]
\begin{threeparttable}
\begin{adjustbox}{center}
\begin{tabular}{|c|c|c|c|c|c|c|}
\hline
Exp.& Dim. & True Drift $f(x)$          & True Diffusion $\sigma(x)$ & \begin{tabular}[c]{@{}c@{}}Step\\ Size $h$\end{tabular} & $\mathcal{D}_{x_0}$& \# Points\\ \hline
1& 1    & $0.5x$                & $0.1$                   & 0.1                                                     & {[}-1,1{]}                                     & 10000                                                      \\
2& 2    & $-x$                  & $0.05x + 0.005$       & 0.01                                                    & {[}1,2{]}                                      & 10000                                                      \\
3a& 1    & $-2x^3 + 4x - 1.5$  & $0.05x + 0.5$         & 0.01                                                    & {[}-2,2{]}                                     & 10000                                                      \\
3b& 2    & $-2x^3 + 4x - 1.5$  & $0.05x + 0.5$         & 0.01                                                    & {[}-2,2{]}& 10000                                                      \\
4a& 3    & $-x$                  & \footnotesize $\begin{bmatrix} 0.02056 & 0.03502 & 0.02678 \\ 0.03502 & 0.06356 & 0.02982 \\ 0.02678 & 0.02982 & 0.12454 \end{bmatrix}$                           & 0.01& {[}-1,1{]}& 10000\\
4b& 3    & $-x$                  & \footnotesize $\begin{bmatrix} 0.09506 & 0 & 0 \\ 0.04639 & 0.15817 & 0 \\ 0.04338 & 0.07507 & 0.00853 \end{bmatrix}$                        & 0.01& {[}-1,1{]}& 10000\\ \hline
\end{tabular}
\end{adjustbox}
\end{threeparttable}
\caption{Summary of the training data for the polynomial experiments 1 to 4.}
\label{tab:polynomial_training_data}
\end{table}

\subsection{Underdamped Langevin Equations: Experiment 5}
This experiment considers the underdamped Langevin equation, a second-order SDE commonly used in physics to model stochastic particle dynamics. In this second-order system, noise is applied only to the velocity variables $v_t$ as follows
\begin{equation}
\begin{aligned}
    \mathrm{d}x_t &= v_t \, \mathrm{d}t, \\
    \mathrm{d}v_t &= f(x_t, v_t) \, \mathrm{d}t + \sigma(x_t, v_t) \, \mathrm{d}W_t.
\end{aligned}
\label{eq:langevin}
\end{equation}
Thus, the density of $x_t$ is no longer Gaussian, indicating that a loss function other than (\ref{eq:loss}) is required to approximate the drift $f$ and diffusion $\sigma$. Equation (\ref{eq:langevin}) can be approximated via the (symplectic) EM scheme 
\begin{equation}
\begin{aligned}
    x_1 &= x_0 + v_0 h, \\
    v_1 &= v_0 + f(x_1, v_0) h + \sigma(x_1, v_0) \delta W_0.
\end{aligned}
\label{eq:langevin_EM}
\end{equation}
This work only assumes access to snapshot data $((x_0, v_0), (x_1, v_1), h)$; hence,  $x_1$ in (\ref{eq:langevin_EM}) can be considered a fixed parameter for the functions $f$ and $\sigma$ over a time span of $h$. Then, $v_1$ is again normally distributed, conditioned on $x_1$, $v_0$ and step size $h$. Thus, the following loss function can be minimized to learn for $f_\theta$ and $\Sigma_{\theta'}$
\begin{align*}
    \mathcal{L}(\theta, \theta' \mid x_1, v_0, v_1, h) =\ 
    &\frac{1}{2} (v_1 - v_0 - h f_\theta(x_1, v_0))^\top (h\Sigma_{\theta'}(x_1, v_0))^{-1} (v_1 - v_0 - h f_\theta(x_1, v_0)) \\
    &+ \frac{1}{2} \log |h\Sigma_{\theta'}(x_1, v_0)| + \frac{D}{2} \log(2\pi).
\end{align*}

For this experiment, training data is generated according to Table \ref{tab:Langevin_training_data}, where the initial distributions of $v_0$ and $x_0$ data points are uniform across the respective domains. 

\begin{table}[ht!]
\begin{threeparttable}
\begin{adjustbox}{center}
\begin{tabular}{|c|c|c|c|c|c|c|c|c|}
\hline
Exp. & $v_0$ Dim. & $x_0$ Dim. & \begin{tabular}[c]{@{}c@{}}True Drift\\ $f(x, v)$\end{tabular} & \begin{tabular}[c]{@{}c@{}}True Diffusion\\ $\sigma(x, v)$\end{tabular} & \begin{tabular}[c]{@{}c@{}}Step\\ Size $h$\end{tabular} & $\mathcal{D}_{v_0}$& $\mathcal{D}_{x_0}$& \# Points\\ \hline
5 & 1 & 1 & $-x^3 - x + 0.5v$ & $\sqrt{0.1}$ & 0.01 & {[}-2.5,2.5{]} & {[}-2.5,2.5{]} & 10000 \\ \hline
\end{tabular}
\end{adjustbox}
\end{threeparttable}
\caption{Summary of the training data for the underdamped Langevin experiment.}
\label{tab:Langevin_training_data}
\end{table}

\subsection{Stochastic SIR Model: Experiment 6}
This section explores the ability of the model to learn the mean-field SDE from data generated from fine-scale particle simulations.  This work considers an epidemic Susceptible-Infected-Recovered-Susceptible (SIRS) model \cite{Dietrich2023} derived from Gillespie's stochastic simulation algorithm (SSA) \cite{Gillespie_Alg} that is used to simulate a well-stirred, spatially homogeneous system. For the SIRS model, SSA tracks three integer-valued stochastic variables: the number of susceptible individuals, $n_0$, the number of infected individuals, $n_1$, and the number of recovered individuals, $n_2$.  Their concentrations are calculated as $\tilde{x}_p = n_p/N$, for $p = 0, 1, 2$, where $N$ is the total population size. 

At each time step $h$, the transition rates are computed as $r_1 = 4k_1\tilde{x}_0\tilde{x}_1$, $r_2 = k_2\tilde{x}_1$, and $r_3 = k_3\tilde{x}_2$. Then an event $I + S \to I + I$, $I \to R$, or $R \to S$ is selected with a probability proportional to its corresponding rate. The values of $n_p$ are updated accordingly. The time step is sampled as $h = -\log(U_1)/(r_1 + r_2 + r_3)$, where $U_1 \sim \mathrm{Uniform}(0,1)$. This simulation procedure is repeated until the trajectory reaches a predefined final time.

The training data is generated along several trajectories initiated with $\tilde{x}_0$, $\tilde{x}_1$ and $\tilde{x}_2$ sampled randomly from the two-simplex, such that $N\tilde{x}_p$ is an integer, and incremented by SSA. Trajectory data is formatted as $\mathcal{D}=\{(x_1^{(n)}, x_0^{(n)}, h_{(n)})\}_{n=1}^N$ where $x_0^{(n)} = (\tilde{x}_0^{(n)}, \tilde{x}_2^{(n)})\in \mathbb{R}^2$, and $x_1^{(n)}$ represents the state after $x_0^{(n)}$ evolves under SIRS with time step $h_{(n)}$. Note that $\tilde{x}_1$ is redundant because $\tilde{x}_0 + \tilde{x}_1 + \tilde{x}_2 = 1$. The experimental data is generated according to the parameters in Table \ref{tab:SIR_training_data}.
\begin{table}[ht!]
\begin{threeparttable}
\begin{adjustbox}{center}
\begin{tabular}{|c|c|c|c|c|}
\hline
Exp. & $k$ & Max Time& \# Trajectories & Mean \# Points\\ \hline
6 & $k_1$, $k_2$, $k_3$ = 1, 1, 0 & 4& 250 & 22942\\ \hline
\end{tabular}
\end{adjustbox}
\end{threeparttable}
\caption{Summary of the training data for the susceptible-infected-recovered (SIR) experiment.}
\label{tab:SIR_training_data}
\end{table}

Since $k_3$ = 0, the SIRS model reduces to the SIR model.

SSA simulates individual events in a Markov jump process. For a large population size $N$, the mean-field population behavior can be modeled using a system of Langevin-type SDEs \cite{Gillespie_SDE}
\begin{equation*}
\begin{gathered}
    \mathrm{d}\tilde{x}_0 = (-r_1 + r_3)\mathrm{d}t - \sqrt{\frac{r_1}{N}} \mathrm{d}W_1(t) + \sqrt{\frac{r_3}{N}} \mathrm{d}W_3(t), \\
    \mathrm{d}\tilde{x}_1 = (r_1 - r_2)\mathrm{d}t + \sqrt{\frac{r_1}{N}} \mathrm{d}W_1(t) - \sqrt{\frac{r_2}{N}} \mathrm{d}W_2(t), \\
    \mathrm{d}\tilde{x}_2 = (r_2 - r_3)\mathrm{d}t + \sqrt{\frac{r_2}{N}} \mathrm{d}W_2(t) - \sqrt{\frac{r_3}{N}} \mathrm{d}W_3(t),
\end{gathered}
\label{eq:SIR_SDE}
\end{equation*}

where $W_p(t)$ are independent Wiener processes such that $dW_p(t) \sim \mathcal{N}(0,dt)$. For the SIR model, this formulation simplifies to a system with drift and diffusion functions: 
\begin{equation}
f(x) = 
\begin{bmatrix} 
-r_1 \\ 
r_2 
\end{bmatrix}, \quad 
\sigma(x) = 
\begin{bmatrix}
\sqrt{\frac{r_1}{N}} & 0 \\
0 & \sqrt{\frac{r_2}{N}}
\end{bmatrix}.
\label{eq:SIR_drift_diff}
\end{equation}
Data points generated using SSA according to Table \ref{tab:SIR_training_data} can be applied to Algorithm \ref{alg:ARFF_SDE} to learn drift and diffusion functions that approximate (\ref{eq:SIR_drift_diff}) for large $N$.

\subsection{Stochastic Wave Equation: Experiment 7}

Stochastic partial differential equations extend SDEs to infinite-dimensional spaces, capturing system dynamics where finite-size effects remain significant. Unlike finite-dimensional SDEs, SPDEs lack universally applicable higher-order numerical schemes akin to EM \cite{walsh2006numerical}. However, certain SPDEs can be reformulated into existing numerical schemes while preserving critical dynamics. This work reintroduces one such example from \cite{Dietrich2023}; a stochastic wave equation with deterministic driving $f(x)$ and stochastic forcing $\sigma(x)$, given by

\begin{equation}
\begin{gathered}
    \frac{\partial^{2}u}{\partial t^{2}} = \frac{\partial^{2}u}{\partial x^{2}} + f(x) + \sigma(x)\mathrm{d}W_{t}, \quad x \in \mathbb{R},\ t > 0, \\
    u(x, 0) = u_{0}(x), \quad \frac{\partial u}{\partial t}(x, 0) = v_{0}(x), \quad x \in \mathbb{R}.
\end{gathered}
\label{eq:wave_eq}
\end{equation}

To solve (\ref{eq:wave_eq}), a small step size $h > 0$ is taken to discretize space and time into grid points $x_{i} = ih$, $t_{j} = jh$. The following numerical scheme approximates the solution to the stochastic wave equation on this staggered grid

\begin{equation}
\begin{gathered}
u_{i,-1} = u(x_{i}, t_{-1}) = \frac{1}{2}(u_{0}(x_{i-1}) + u_{0}(x_{i+1})) - hv_{0}(x_{i}), \quad i \text{ odd}; \\
u_{i,1} = u_{0}(x_{i-1}) + u_{0}(x_{i+1}) - u_{i,-1} + \frac{h^{2}}{2}f(x_{i}) + \frac{1}{2}\sigma(x_{i})\delta W_{0}(h^{2}/2), \quad i \text{ odd}; \\
u_{i,j+1} = u_{i+1,j} + u_{i-1,j} - u_{i,j-1} + h^{2}f(x_{i}) + \frac{1}{2}\sigma(x_{i})\delta W_{0}(h^{2}), \quad (i(j+1)) \text{ even},\ j \geq 2.
\end{gathered}
\label{eq:wave_scheme}
\end{equation}

The scheme in (\ref{eq:wave_scheme}) can be reformulated to match the EM scheme (\ref{eq:euler maruyama}) by defining

\begin{equation*}
\tilde{h} := \frac{1}{2}h^{2}, \quad \tilde{u}_{i,j} := \frac{1}{2}(u_{i+1,j} + u_{i-1,j}), \quad \tilde{u}_{i,j+1} := \frac{1}{2}(u_{i,j+1} + u_{i,j-1}).
\end{equation*}

Then,

\begin{equation*}
\tilde{u}_{i,j+1} = \tilde{u}_{i,j} + \tilde{h}f(x_{i}) + \sigma(x_{i})\delta W_{0}(\tilde{h}), \quad (i(j+1)) \text{ even}, j \geq 2.
\end{equation*}

Given the data $\{\tilde{u}_{i,j+1}, \tilde{u}_{i,j}, \tilde{h}\}_{i,j}$, the forcing terms $f$, $\sigma$ can be learned by minimizing of the loss (\ref{eq:total_loss}). 

The training data is generated according to the parameters in Table \ref{tab:SPDE_training_data}. The initial values of $u$ are evaluated on a uniform grid spanning the spatial domain $x_0$ and the temporal domain $t$. The step size $h$ dictates the grid resolution, thereby determining the total number of training data points.

\begin{table}[ht!]
\begin{threeparttable}
\begin{adjustbox}{center}
\begin{tabular}{|c|c|c|c|c|c|c|c|c|c|}
\hline
Exp. & Dim. & 
\begin{tabular}[c]{@{}c@{}}True Drift\\ $f(x, t, u)$\end{tabular} & 
\begin{tabular}[c]{@{}c@{}}True Diffusion\\ $\sigma(x, t, u)$\end{tabular} & 
$u_0(x)$ & 
$v_0(x)$ & 
\begin{tabular}[c]{@{}c@{}}Step\\ Size $h$\end{tabular} & 
$\mathcal{D}_{x_0}$ & 
$\mathcal{D}_t$ & 
\begin{tabular}[c]{@{}c@{}}\#\\ Points \end{tabular} \\ \hline

7 & 1 & 
\footnotesize{$5\sin\left(4\pi x \right)$} & 
\footnotesize{$\frac{1}{20} \left(1+\mathrm{e}^{ \left(-150\left(x - \frac{1}{2}\right)^2\right)} \right)$} & 
\footnotesize{$\frac{1}{20} \mathrm{e}^{\left(-150 \left(x - \frac{1}{2} \right)^2 \right)}$} & \footnotesize{$-2 \frac{du_0}{dx}$} & 
0.001 & {[}0,1{]} & {[}0,2{]} & \footnotesize{995004} \\ \hline
\end{tabular}
\end{adjustbox}
\end{threeparttable}
\caption{Summary of the training data for the stochastic wave equation experiment.}
\label{tab:SPDE_training_data}
\end{table}

This manuscript only considers the autonomous case, where $f$ and $\sigma$ do not depend on time. 

\vspace{1em}

For Experiments 1 to 5, the training datasets were generated using the EM scheme (\ref{eq:euler maruyama}) with an integration step size equal to one-thousandth of the sample step size $h$, ensuring a better approximation of the underlying It\^{o} diffusion process (\ref{eq:ito diffusion SDE}). Experiment 6 employs generation step size equal to one-tenth of the sample step size $h$ and Experiment 7 employs identical step sizes for integration and sampling. All datasets were split into 90\% for training and 10\% for validation.

\section{Results}\label{sec:results}

This work compares networks trained using Algorithm \ref{alg:ARFF_SDE} against networks trained using the Adam optimizer on the datasets specified in Tables \ref{tab:polynomial_training_data}, \ref{tab:Langevin_training_data}, \ref{tab:SIR_training_data}, and \ref{tab:SPDE_training_data}. Both methods trained networks in $\mathcal{N}_K$ for the same $K$. 

Hyperparameters were maintained across experiments unless the training struggled to converge, in which case modifications that lead to significant loss or timing improvements were implemented.

For Algorithm \ref{alg:ARFF_SDE}, most experiments employed a Tikhonov regularization parameter $\lambda = 0.002$, a proposal step length $\delta  = 0.1$, an exponent $\gamma = 1$, and a maximum number of iterations $M_{\text{max}}$ large enough to ensure early stopping via validation error stagnation. The exceptions include Experiment 4a, with Tikhonov parameter $\lambda = 2 \times 10^{-6}$, and Experiment 7, with proposal step lengths $\delta= 0.2$ and $\delta = 0.4$ for drift and diffusion training respectively. 

For Adam optimization, all experiments employed the Tensorflow default learning rate of 0.001. The batch sizes were set as follows: Experiments 1 and 7: 512; Experiments 2, 4b and 6: 32; and Experiments 3a, 3b, 4a, and 5: 8. Each experiment was run for 200 epochs.

\begin{figure}[ht!]
    \centering
    \begin{subfigure}[t]{0.43\textwidth}
        \centering
        \includegraphics[width=\textwidth]{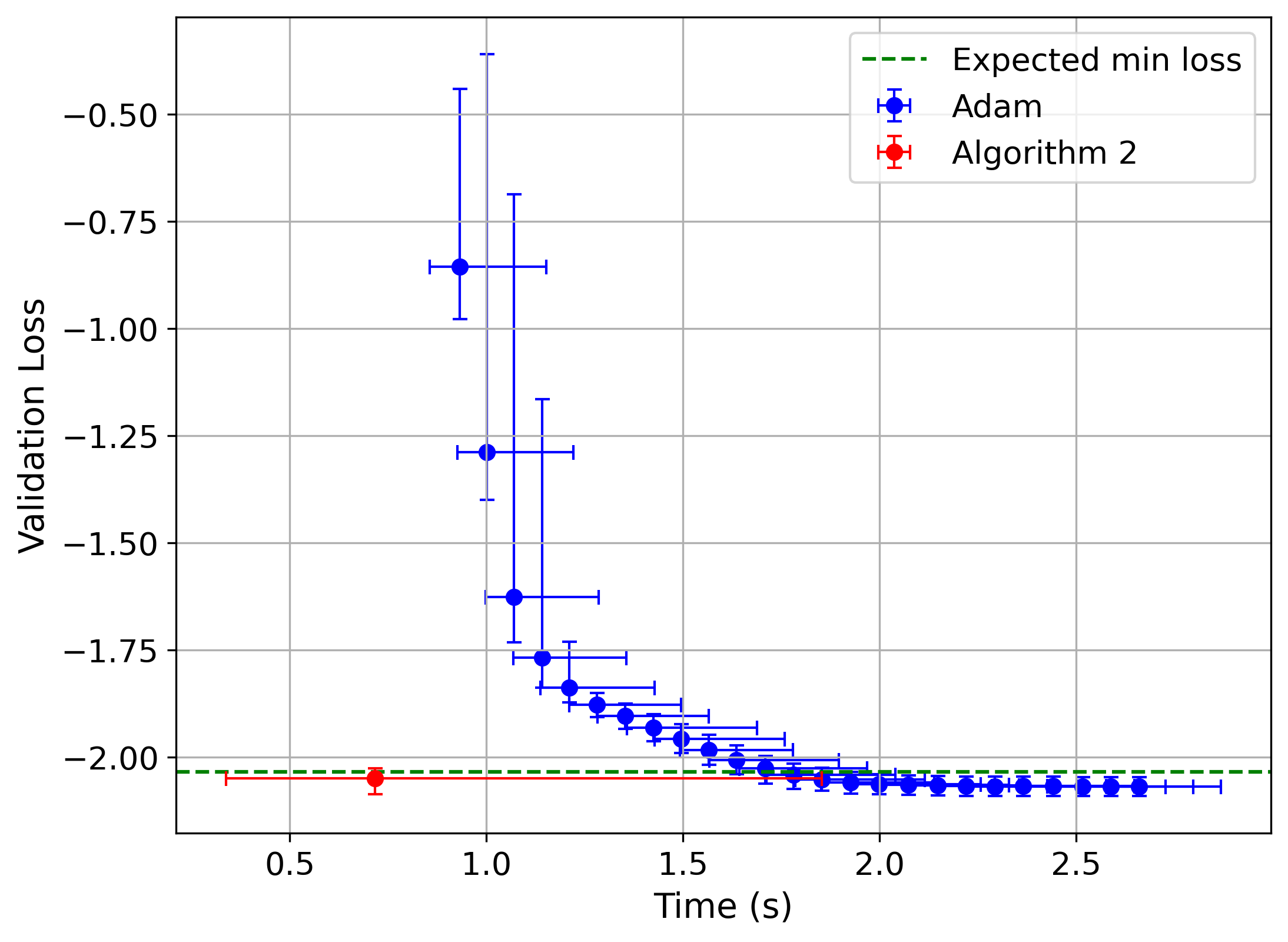}
        \caption{Experiment 1: one-dimensional, linear.}
        \label{fig:example1}
    \end{subfigure}
    \hfill
    \begin{subfigure}[t]{0.43\textwidth}
        \centering
        \includegraphics[width=\textwidth]{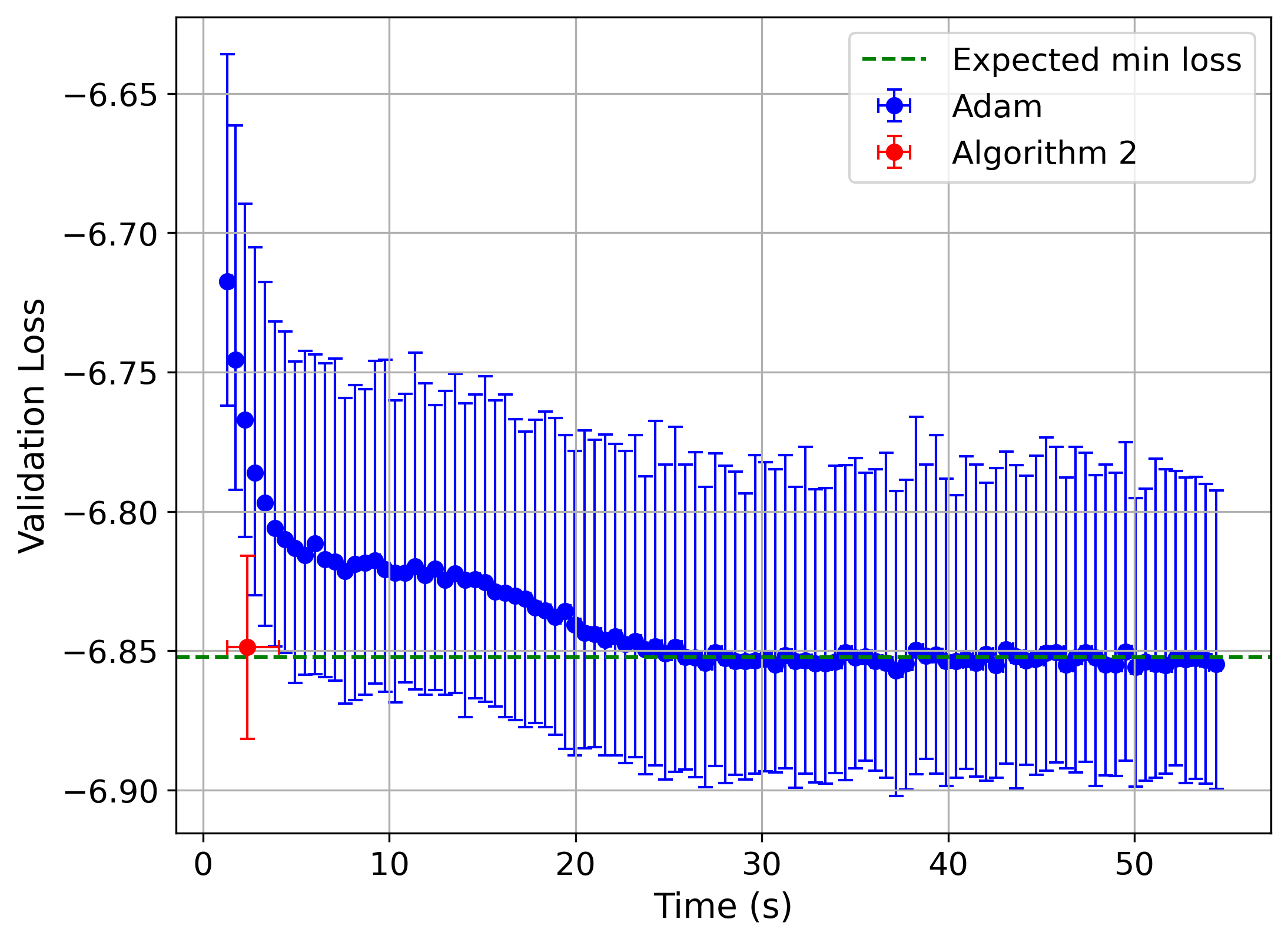}
        \caption{Experiment 2: two-dimensional, linear.}
        \label{fig:example2}
    \end{subfigure}

    \begin{subfigure}[t]{0.43\textwidth}
        \centering
        \includegraphics[width=\textwidth]{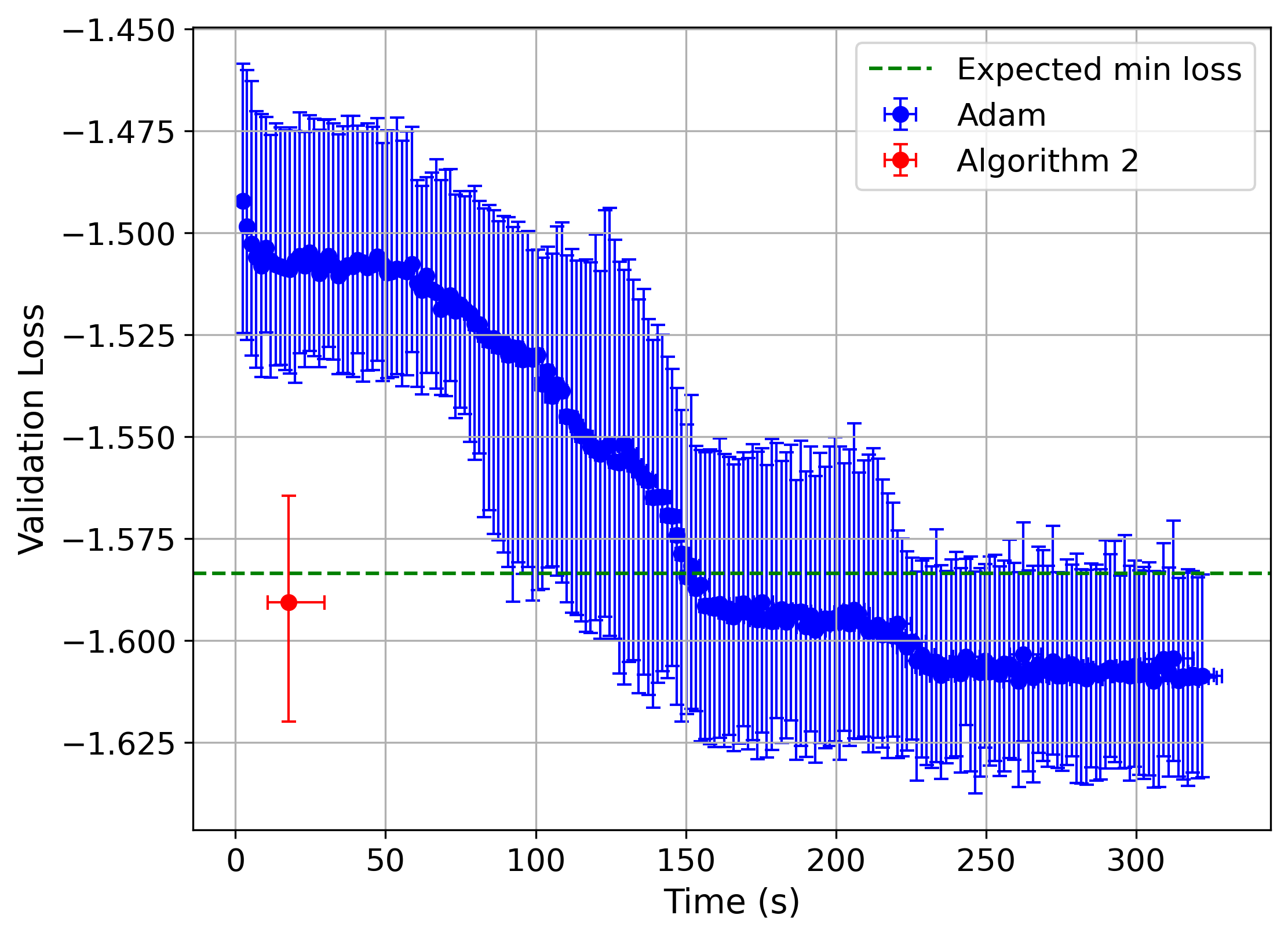}
        \caption{Experiment 3a: one-dimensional, cubic.}
        \label{fig:example3a}
    \end{subfigure}
    \hfill
    \begin{subfigure}[t]{0.43\textwidth}
        \centering
        \includegraphics[width=\textwidth]{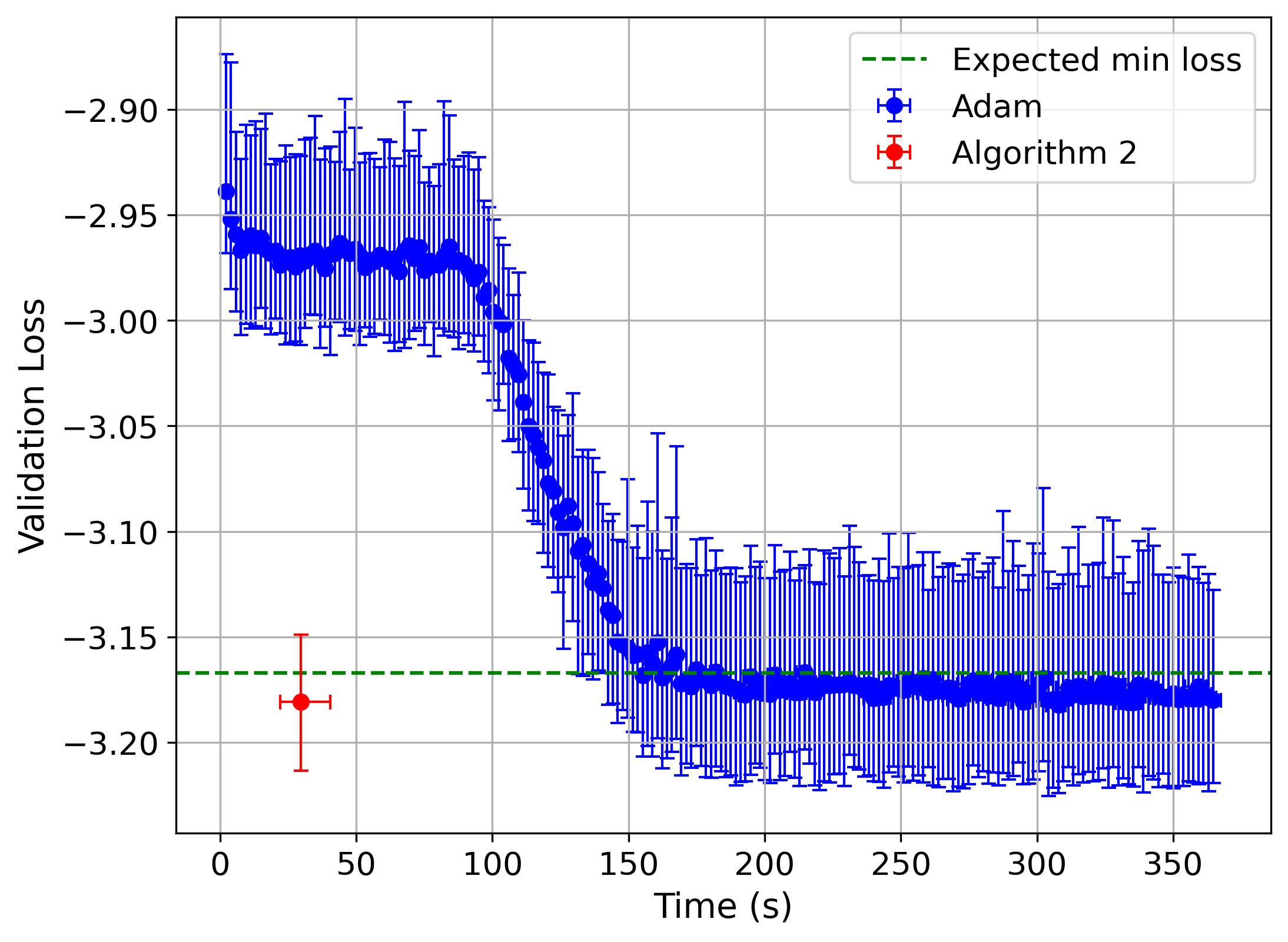}
        \caption{Experiment 3b: two-dimensional, cubic.}
        \label{fig:example3b}
    \end{subfigure}

    \begin{subfigure}[t]{0.43\textwidth}
        \centering
        \includegraphics[width=\textwidth]{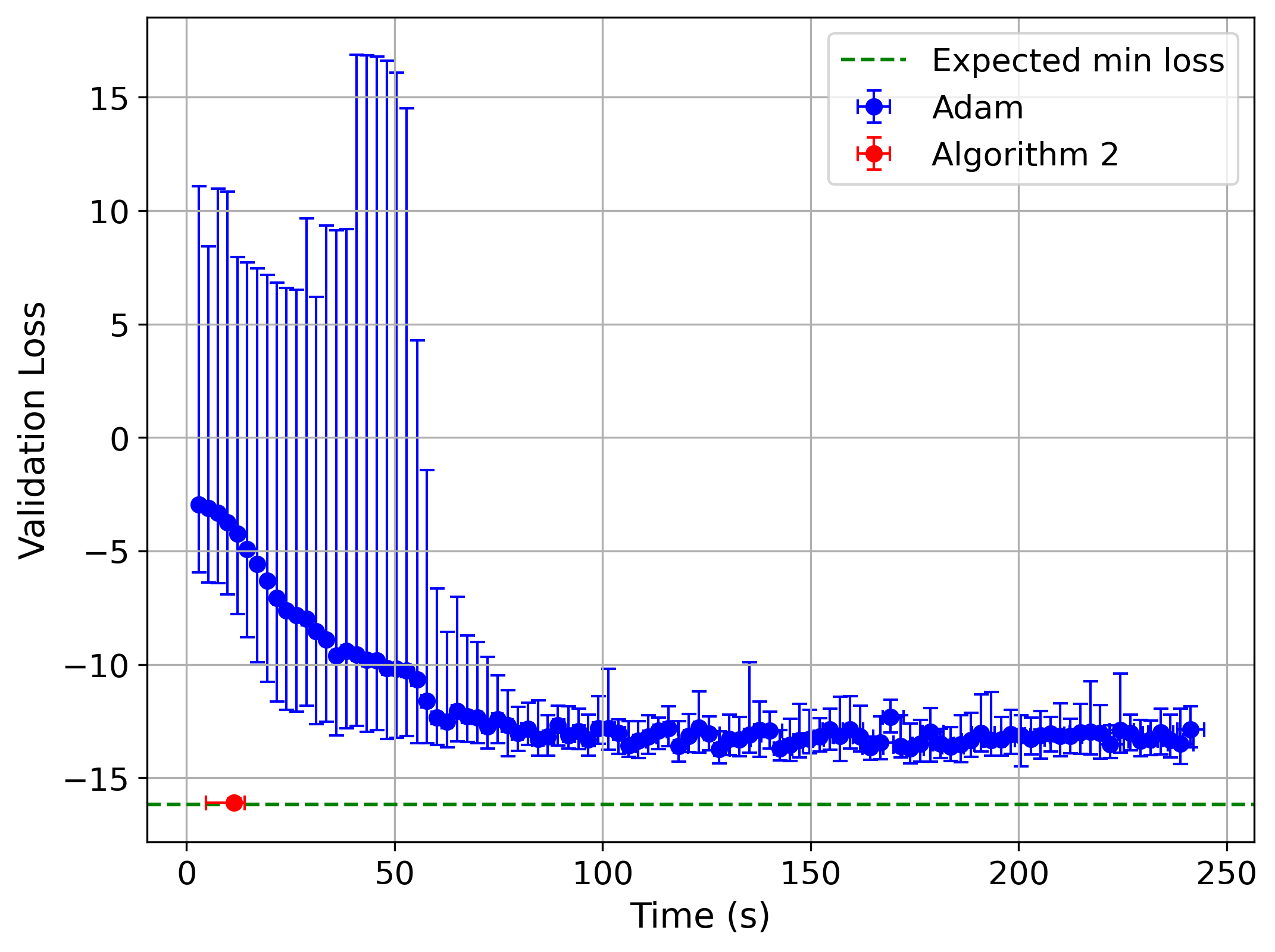}
        \caption{Experiment 4a: three-dimensional, symmetric diffusion.}
        \label{fig:example4a}
    \end{subfigure}
    \hfill
    \begin{subfigure}[t]{0.43\textwidth}
        \centering
        \includegraphics[width=\textwidth]{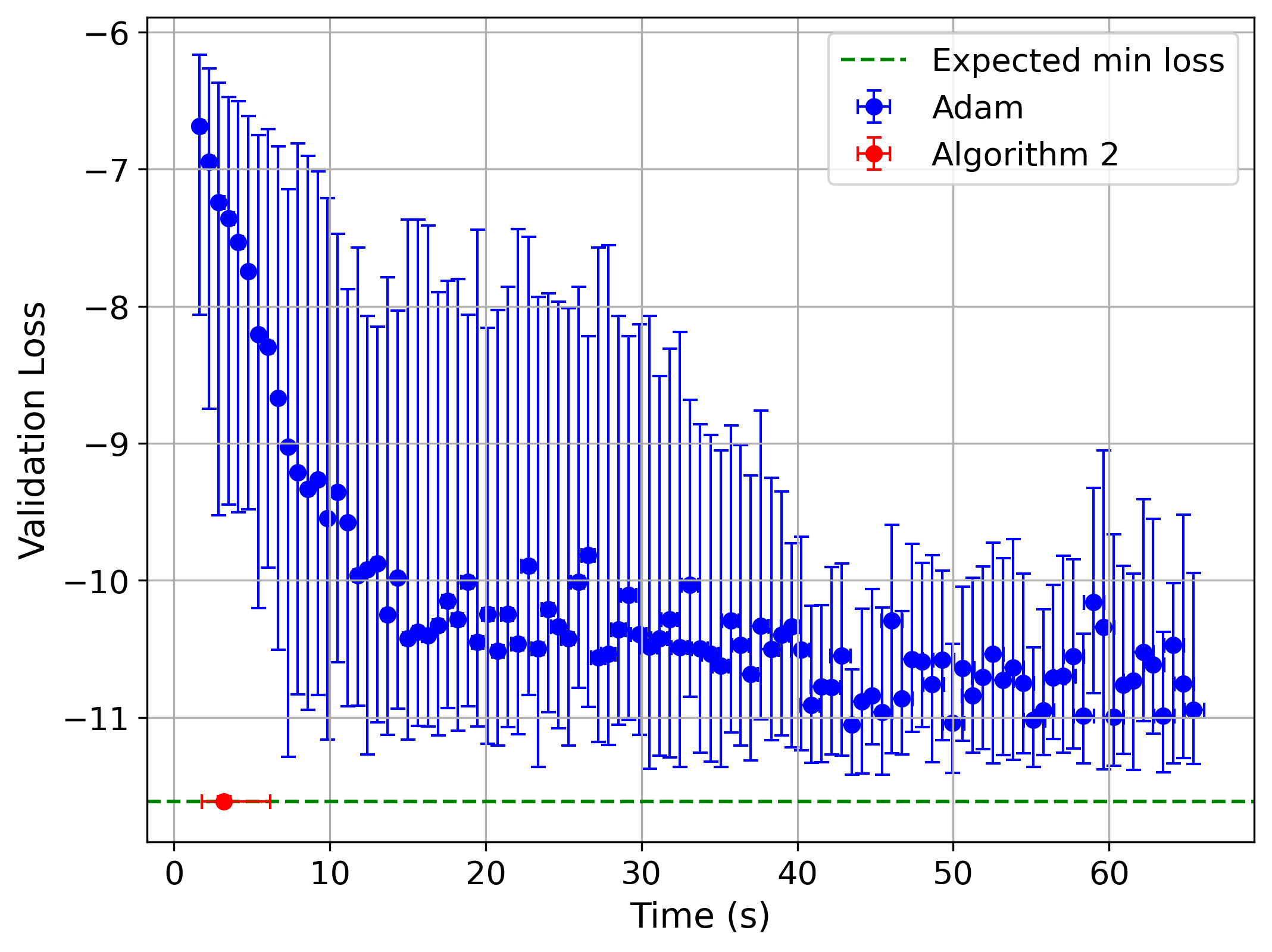}
        \caption{Experiment 4b: three-dimensional, lower-triangular diffusion.}
        \label{fig:example4b}
    \end{subfigure}

    \begin{subfigure}[t]{0.43\textwidth}
        \centering
        \includegraphics[width=\textwidth]{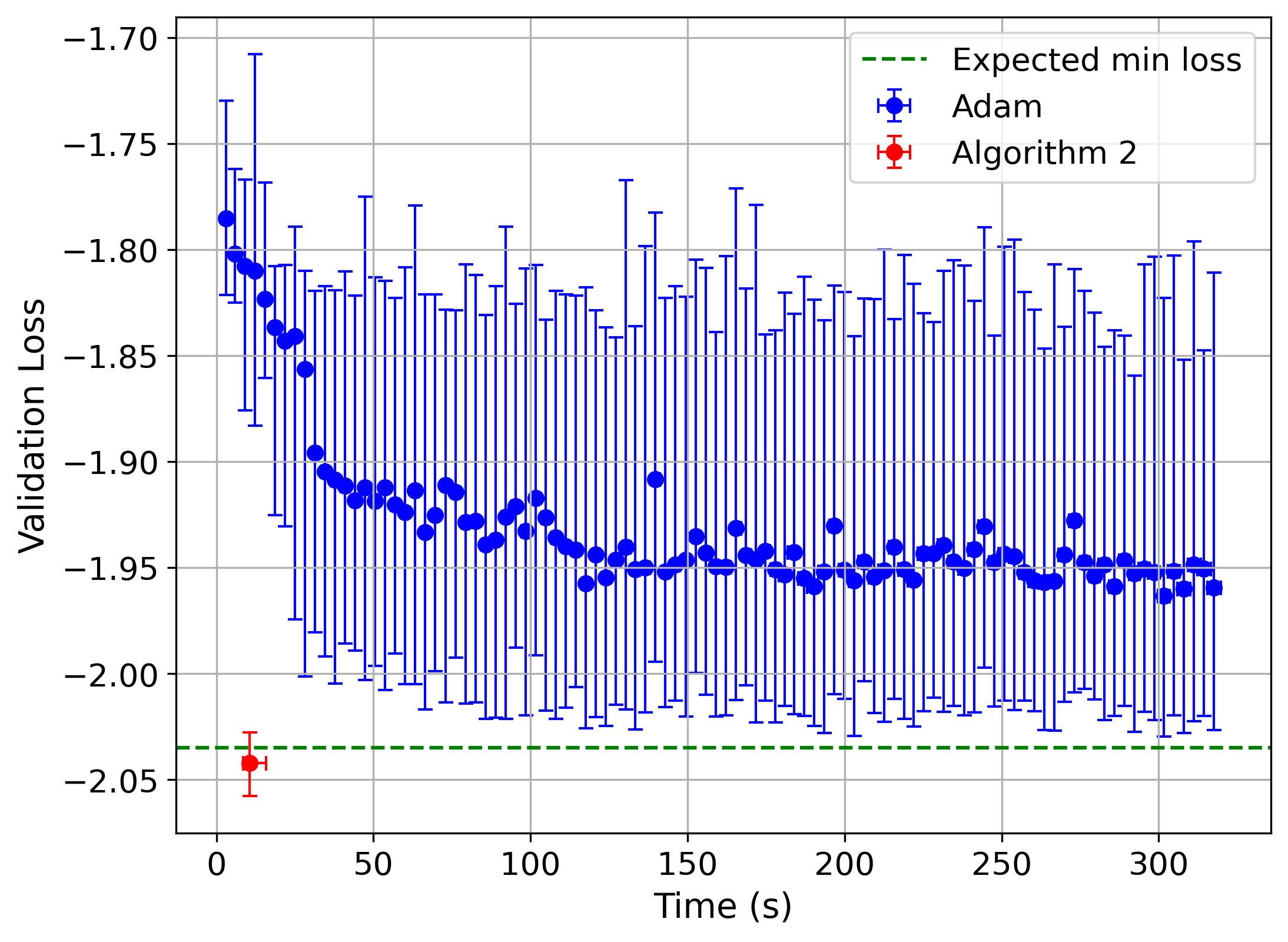}
        \caption{Experiment 5: underdamped Langevin.}
        \label{fig:example5}
    \end{subfigure}
    \hfill
    \begin{subfigure}[t]{0.43\textwidth}
        \centering
        \includegraphics[width=\textwidth]{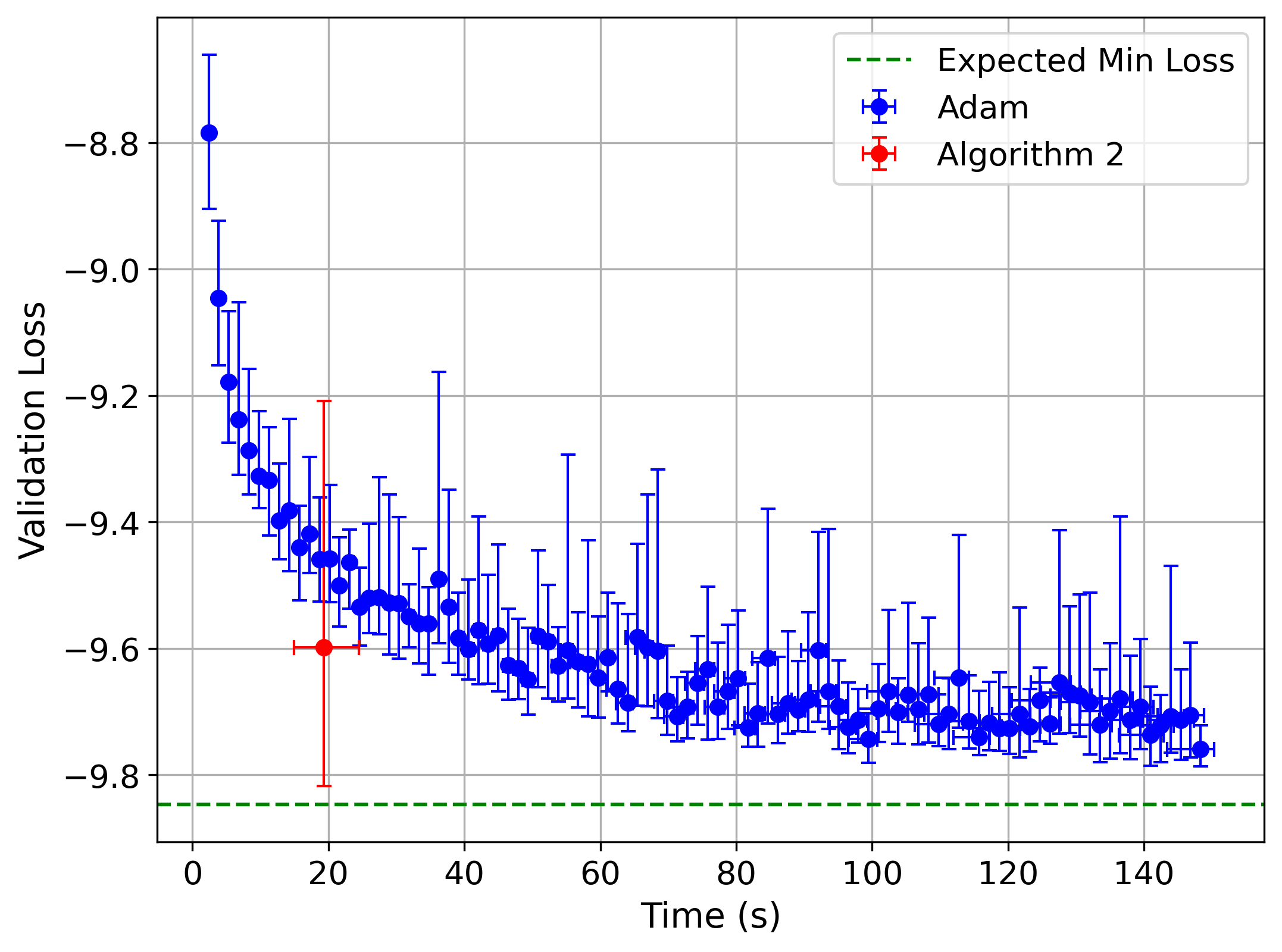}
        \caption{Experiment 6: susceptible, infected, recovered.}
        \label{fig:example6}
    \end{subfigure}

    \caption{Mean validation loss vs mean training time across 10 runs of Experiments 1 to 6 with error bars indicating $\pm$ one standard deviation. For the Adam optimizer, sufficient epochs are plotted to display the important features of the loss-time profile.}
    \label{fig:loss_v_time}
\end{figure}

Table \ref{tab:results} and Figure \ref{fig:loss_v_time} present the mean validation losses and mean training times across 10 runs of the experiments.

\begin{table}[ht!]
\begin{threeparttable}
\begin{adjustbox}{center}
\begin{minipage}{\textwidth}
\begin{tabular}{|c|c|c|cc|cc|}
\hline
\multirow{2}{*}{Exp.} & \multirow{2}{*}{\begin{tabular}[c]{@{}c@{}}Layer\\ size\end{tabular}} & \multirow{2}{*}{\begin{tabular}[c]{@{}c@{}}Expected\\ min loss\tnote{1}\end{tabular}} & \multicolumn{2}{c|}{Min validation loss}                                      & \multicolumn{2}{c|}{Training time (s)}                                       \\ \cline{4-7} 
                      &                                                                       &                                                                                                        & \multicolumn{1}{c|}{Algorithm \ref{alg:ARFF_SDE}} & Adam    & \multicolumn{1}{c|}{Algorithm \ref{alg:ARFF_SDE}} & Adam   \\ \hline
1                     & $2^5$                                                                 &                                                                                                        -2.0349& \multicolumn{1}{c|}{-2.0500}                                               &         -2.0693& \multicolumn{1}{c|}{0.71656}                                               &        2.3370\\
2                     & $2^6$                                                                 &                                                                                                        -6.8523& \multicolumn{1}{c|}{-6.8487}                                               &         -6.8547& \multicolumn{1}{c|}{2.3831}                                               &        26.758\\
3a                    & $2^7$                                                                 &                                                                                                        -1.5835& \multicolumn{1}{c|}{-1.5906}                                               &         -1.5121& \multicolumn{1}{c|}{17.683}                                               &        18.946\\
3b                    & $2^8$                                                                 &                                                                                                        -3.1671& \multicolumn{1}{c|}{-3.1807}                                               &         -2.9826& \multicolumn{1}{c|}{29.606}                                               &        19.734\\
4a                    & $2^7$                                                                 &                                                                                                        -16.170& \multicolumn{1}{c|}{-16.102}                                               &         -14.031& \multicolumn{1}{c|}{11.380}                                               &        58.231\\
4b                    & $2^7$                                                                 &                                                                                                        -11.613& \multicolumn{1}{c|}{-11.612}                                               &         -11.406& \multicolumn{1}{c|}{3.2147}                                               &        17.755\\
5                     & $2^7$                                                                 &                                                                                                        -2.0349& \multicolumn{1}{c|}{-2.0421}                                               &         -1.9331& \multicolumn{1}{c|}{10.528}                                               &        51.303\\
6                     & $2^6$                                                                 &                                                                                                        -9.8469& \multicolumn{1}{c|}{-9.5986}                                               &         -9.646& \multicolumn{1}{c|}{19.325}                                               &        41.557\\
7                     & $2^7$                                                                 & -9.4135& \multicolumn{1}{c|}{-9.4084}                                        & -9.3849& \multicolumn{1}{c|}{2203.2}                                         & 1036.4 \\ \hline
\end{tabular}
\begin{tablenotes}
\item[1] Expected minimum loss is the loss (\ref{eq:total_loss}) calculated on the validation data using the true drift and diffusion functions.
\end{tablenotes}
\end{minipage}
\end{adjustbox}
\end{threeparttable}
\caption{Comparison of the mean minimum validation loss and mean training time across 10 runs of Experiments 1 to 7. For the Adam optimizer, the minimum validation loss is defined as the lowest value before stagnation, characterized as five consecutive epochs with no decrease in the moving average. Training time corresponds to the time taken to reach the minimum loss before stagnation.}
\label{tab:results}
\end{table}

Table \ref{tab:results} and Figure \ref{fig:loss_v_time} reveal that in no experiment does the Adam optimizer achieve a smaller loss in a shorter time than Algorithm \ref{alg:ARFF_SDE}. For all experiments in Figure \ref{fig:loss_v_time}, the loss-time mean for Algorithm \ref{alg:ARFF_SDE} appears below the loss-time means for the Adam optimizer, showing that Algorithm \ref{alg:ARFF_SDE} consistently reaches a given loss threshold faster than Adam.

In Experiments 1, 2, 3a and 6 (Figures \ref{fig:example1}, \ref{fig:example2}, \ref{fig:example3a}, and \ref{fig:example6}), the Adam optimizer surpasses the loss obtained by Algorithm \ref{alg:ARFF_SDE}. In contexts without timing concerns, the Adam optimizer can still outperform Algorithm \ref{alg:ARFF_SDE}.

For most Experiments in Figure \ref{fig:loss_v_time}, the validation loss standard deviation is smaller for Algorithm \ref{alg:ARFF_SDE}, perhaps indicating a higher tendency for the Adam optimizer to become stuck in a range of local minima.  

\vspace{1em}

\begin{figure}[ht!]
    \centering
    \noindent\makebox[\textwidth][c]{%
        \begin{subfigure}[t]{0.45\textwidth}
            \centering
            \includegraphics[width=\textwidth]{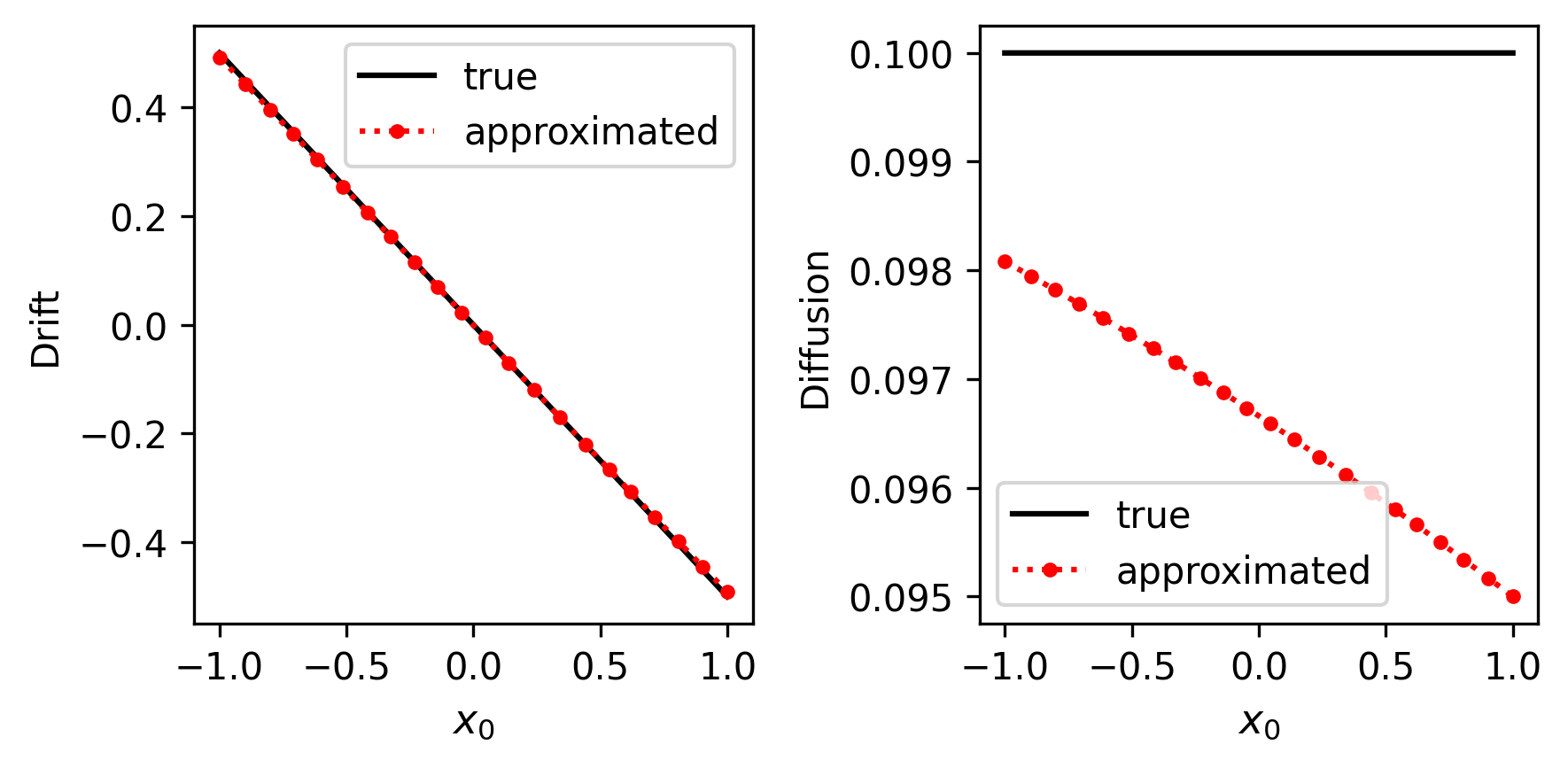}
            \subcaption{Algorithm \ref{alg:ARFF_SDE}.}
        \end{subfigure}
        \hspace{0.5cm}
        \begin{subfigure}[t]{0.45\textwidth}
            \centering
            \includegraphics[width=\textwidth]{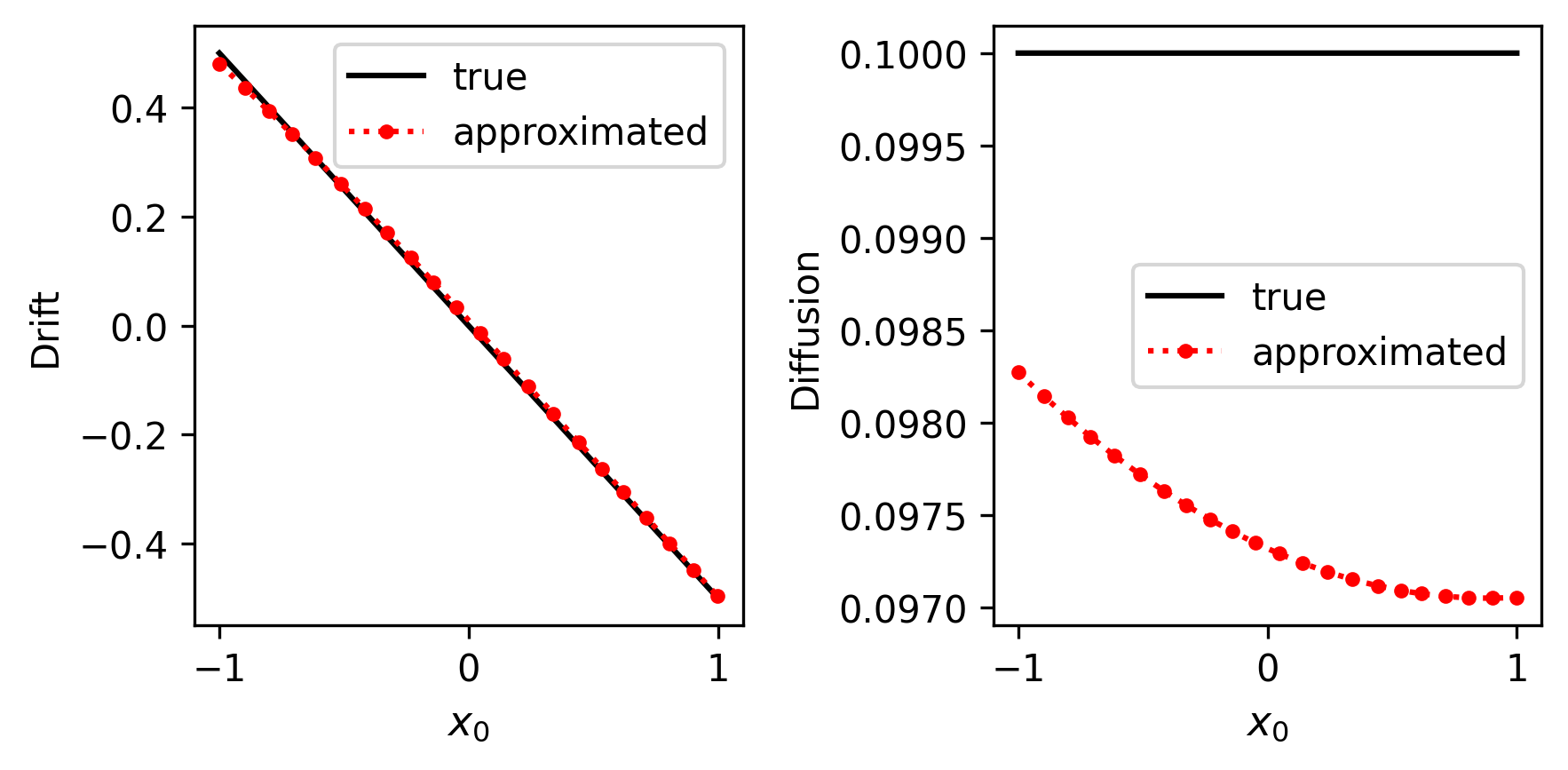}
            \subcaption{Adam optimizer.}
        \end{subfigure}
    }
    \caption{Comparison of the true and trained drift and diffusion functions for Experiment 1: one-dimensional, linear.}
    \label{fig:Exp_1_true_trained}
\end{figure}

\begin{figure}[ht!]
    \centering
    \noindent\makebox[\textwidth][c]{%
        \begin{subfigure}[t]{0.45\textwidth}
            \centering
            \includegraphics[width=\textwidth]{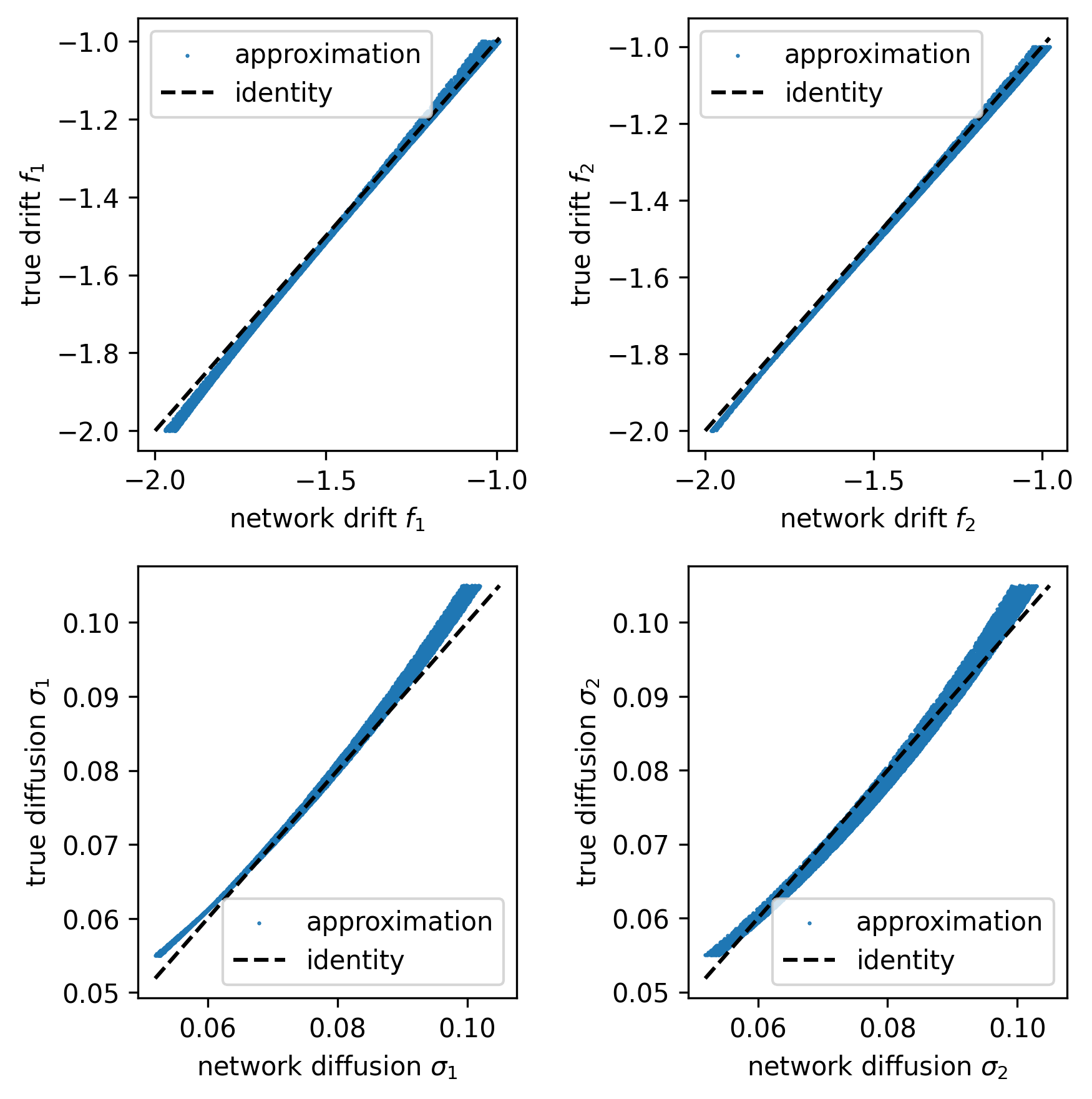}
            \subcaption{Algorithm \ref{alg:ARFF_SDE}.}
        \end{subfigure}
        \hspace{0.5cm}
        \begin{subfigure}[t]{0.45\textwidth}
            \centering
            \includegraphics[width=\textwidth]{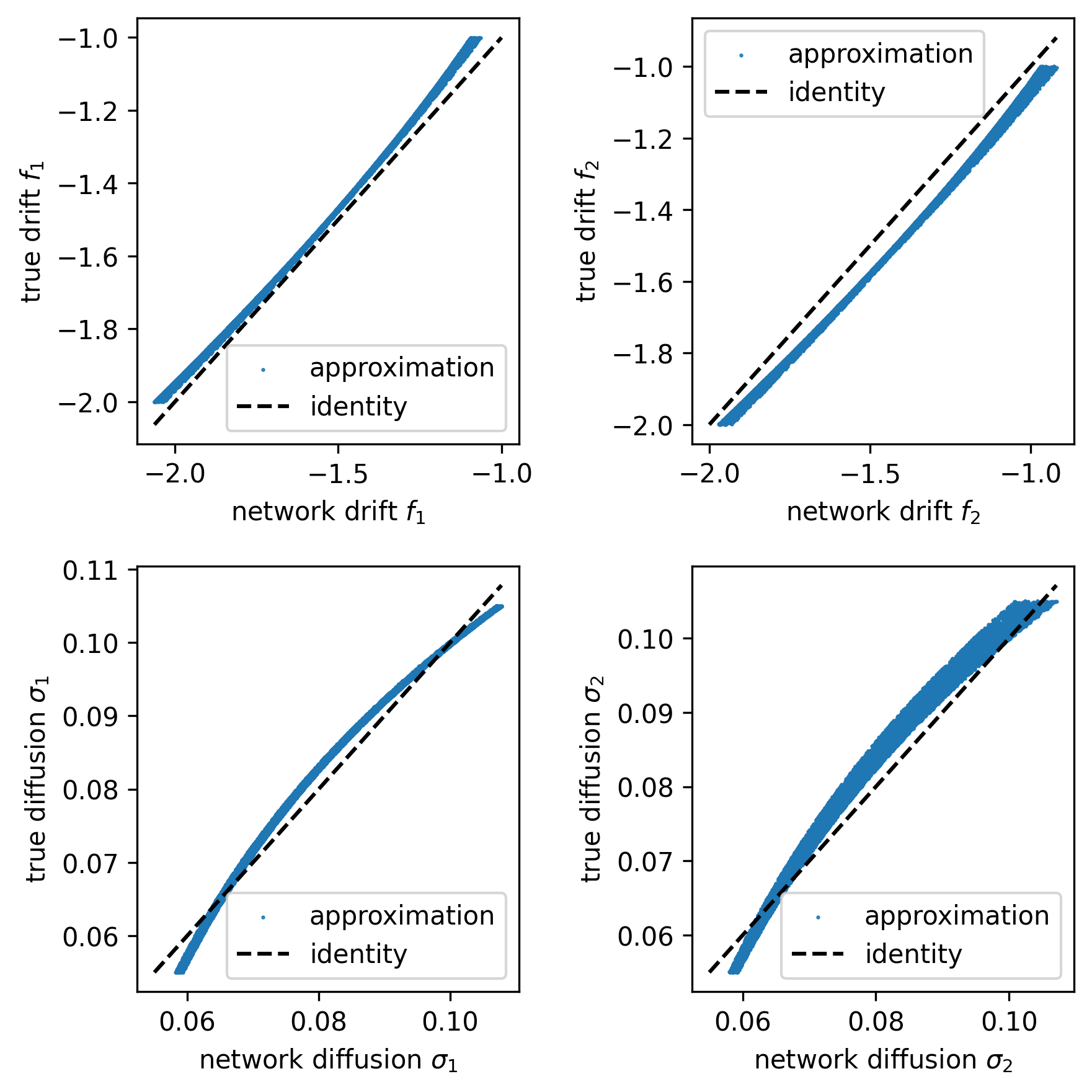}
            \subcaption{Adam optimizer.}
        \end{subfigure}
    }
    \caption{Comparison of the true and trained drift and diffusion functions for Experiment 2: two-dimensional, linear.}
    \label{fig:Exp_2_true_trained}
\end{figure}

\begin{figure}[ht!]
    \centering
    \noindent\makebox[\textwidth][c]{%
        \begin{subfigure}[t]{0.45\textwidth}
            \centering
            \includegraphics[width=\textwidth]{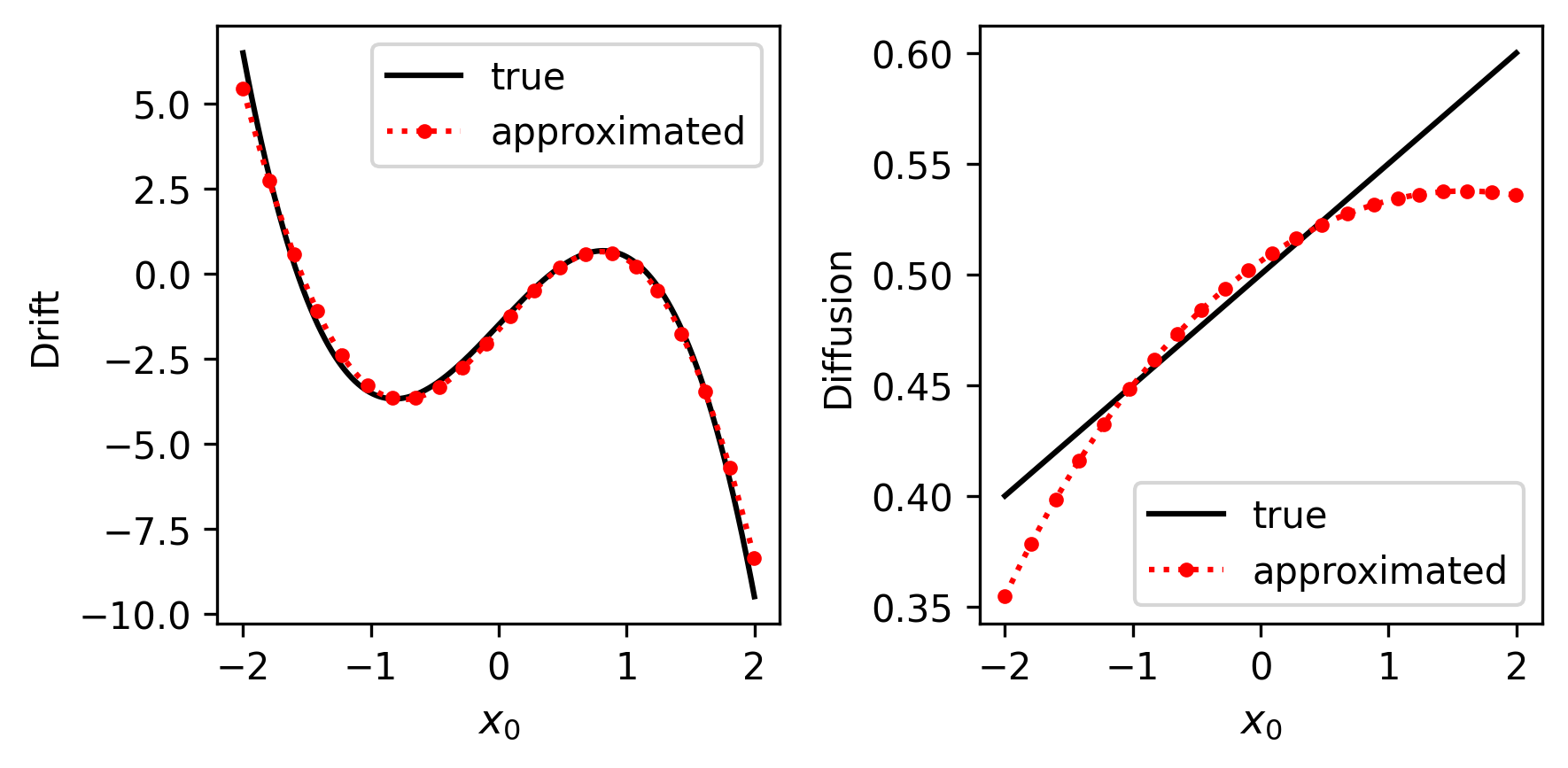}
            \subcaption{Algorithm \ref{alg:ARFF_SDE}.}
        \end{subfigure}
        \hspace{0.5cm}
        \begin{subfigure}[t]{0.45\textwidth}
            \centering
            \includegraphics[width=\textwidth]{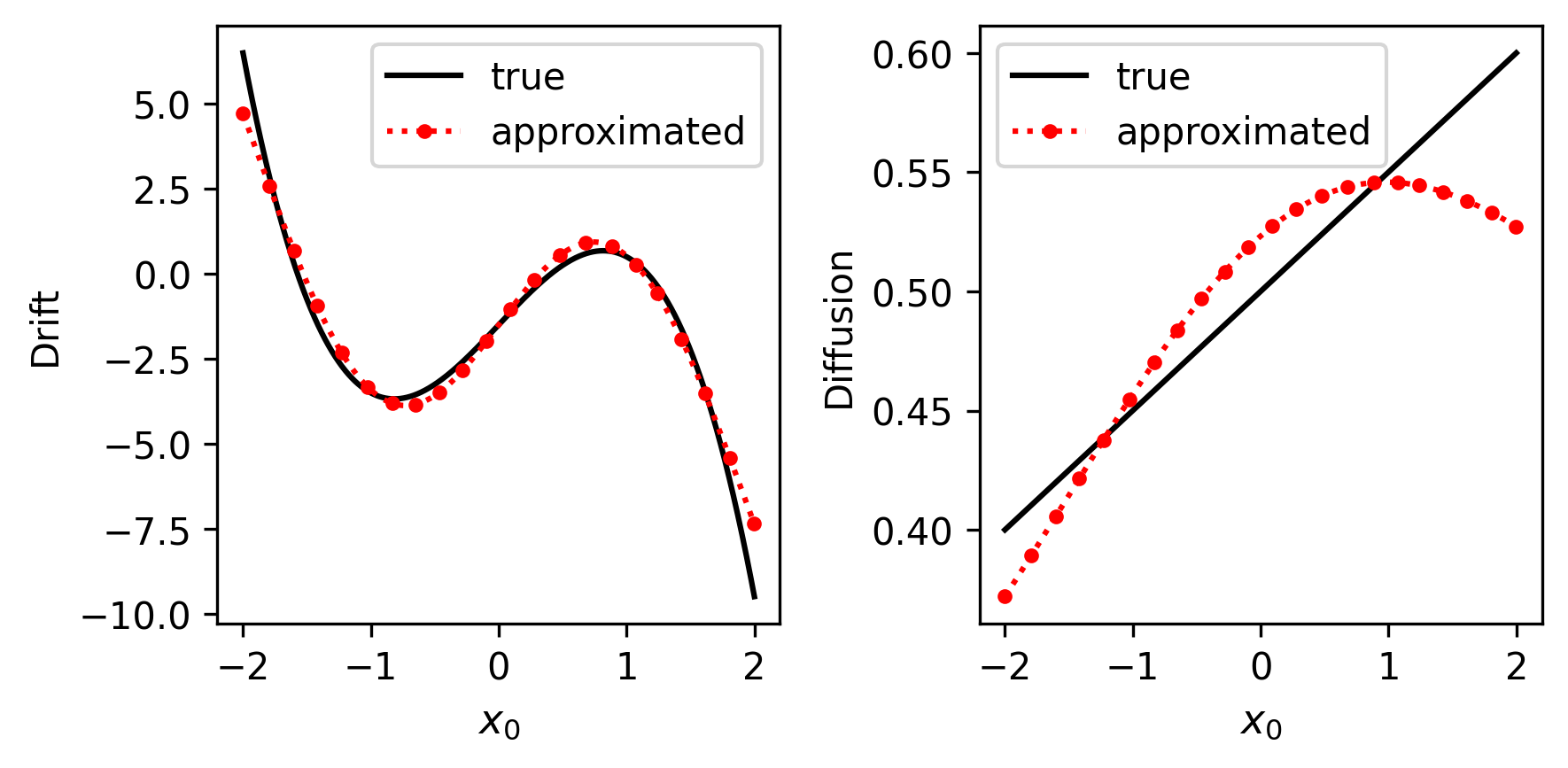}
            \subcaption{Adam optimizer.}
        \end{subfigure}
    }
    \caption{Comparison of the true and trained drift and diffusion functions for Experiment 3a: one-dimensional, cubic.}
    \label{fig:Exp_3a_true_trained}
\end{figure}

\begin{figure}[ht!]
    \centering
    \noindent\makebox[\textwidth][c]{%
        \begin{subfigure}[t]{0.45\textwidth}
            \centering
            \includegraphics[width=\textwidth]{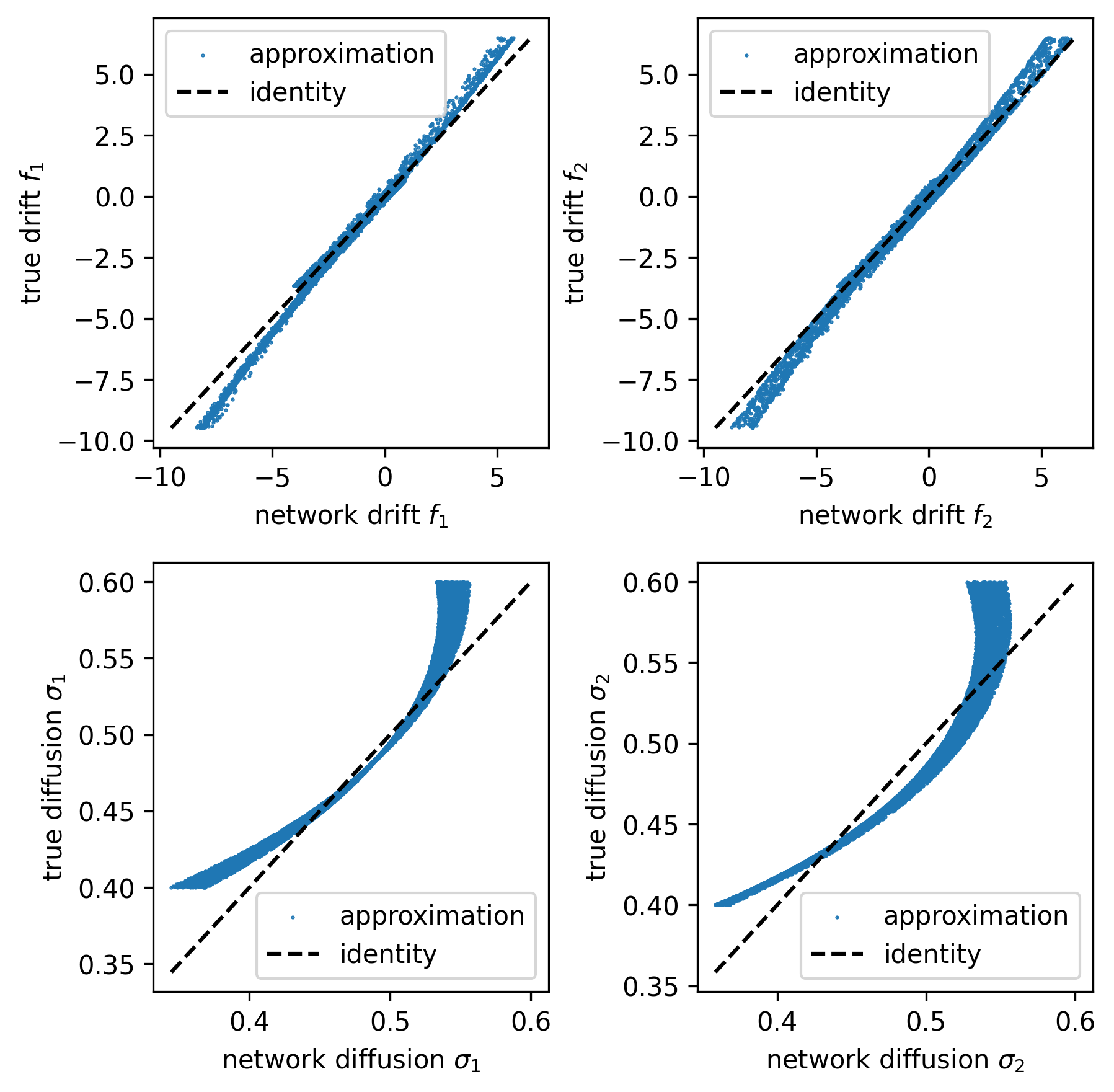}
            \subcaption{Algorithm \ref{alg:ARFF_SDE}.}
        \end{subfigure}
        \hspace{0.5cm}
        \begin{subfigure}[t]{0.45\textwidth}
            \centering
            \includegraphics[width=\textwidth]{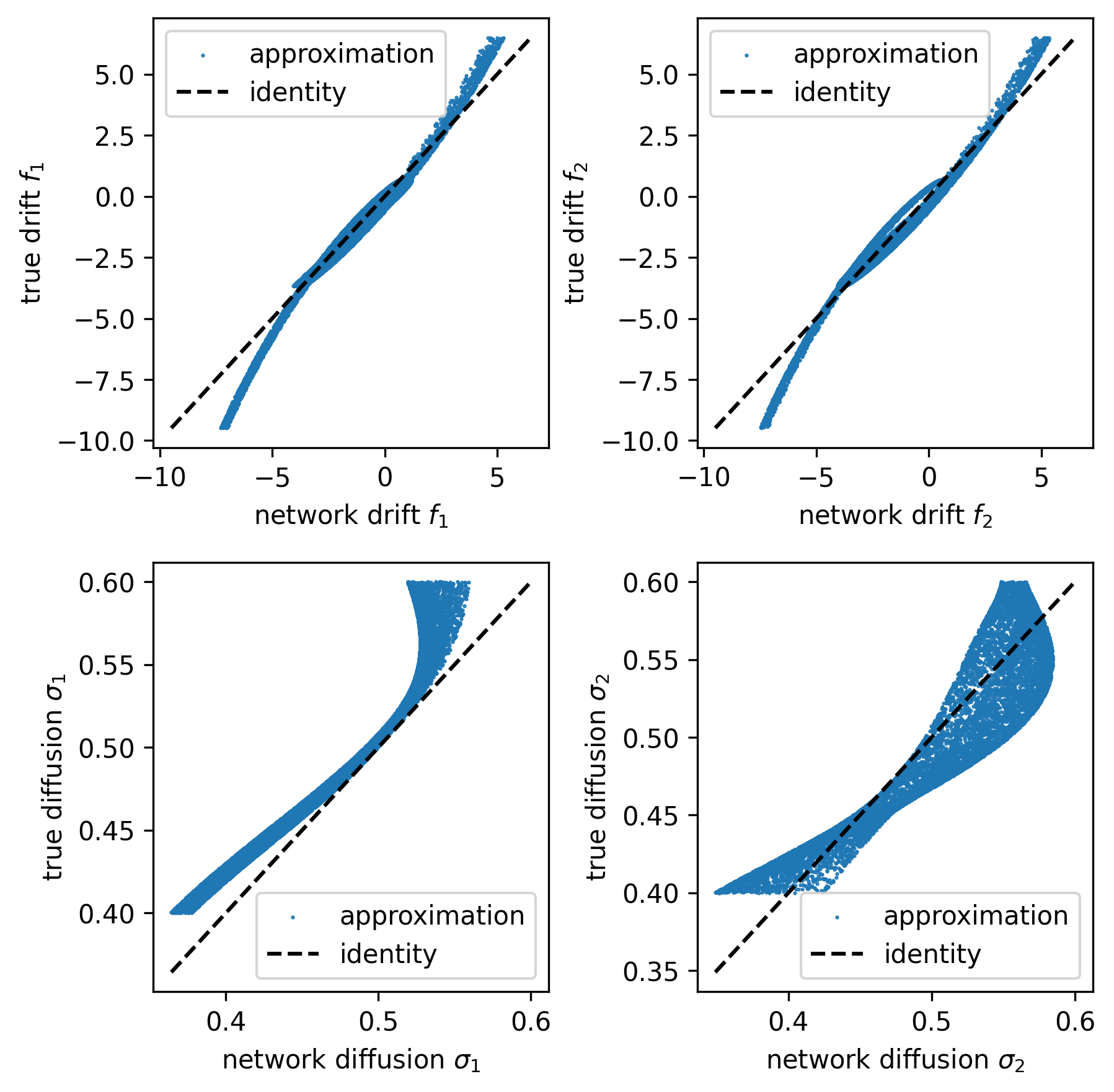}
            \subcaption{Adam optimizer.}
        \end{subfigure}
    }
    \caption{Comparison of the true and trained drift and diffusion functions for Experiment 3b: two-dimensional, cubic.}
    \label{fig:Exp_3b_true_trained}
\end{figure}

\begin{figure}[ht!]
    \centering
    \noindent\makebox[\textwidth][c]{%
        \begin{subfigure}[t]{0.45\textwidth}
            \centering
            \includegraphics[width=\textwidth]{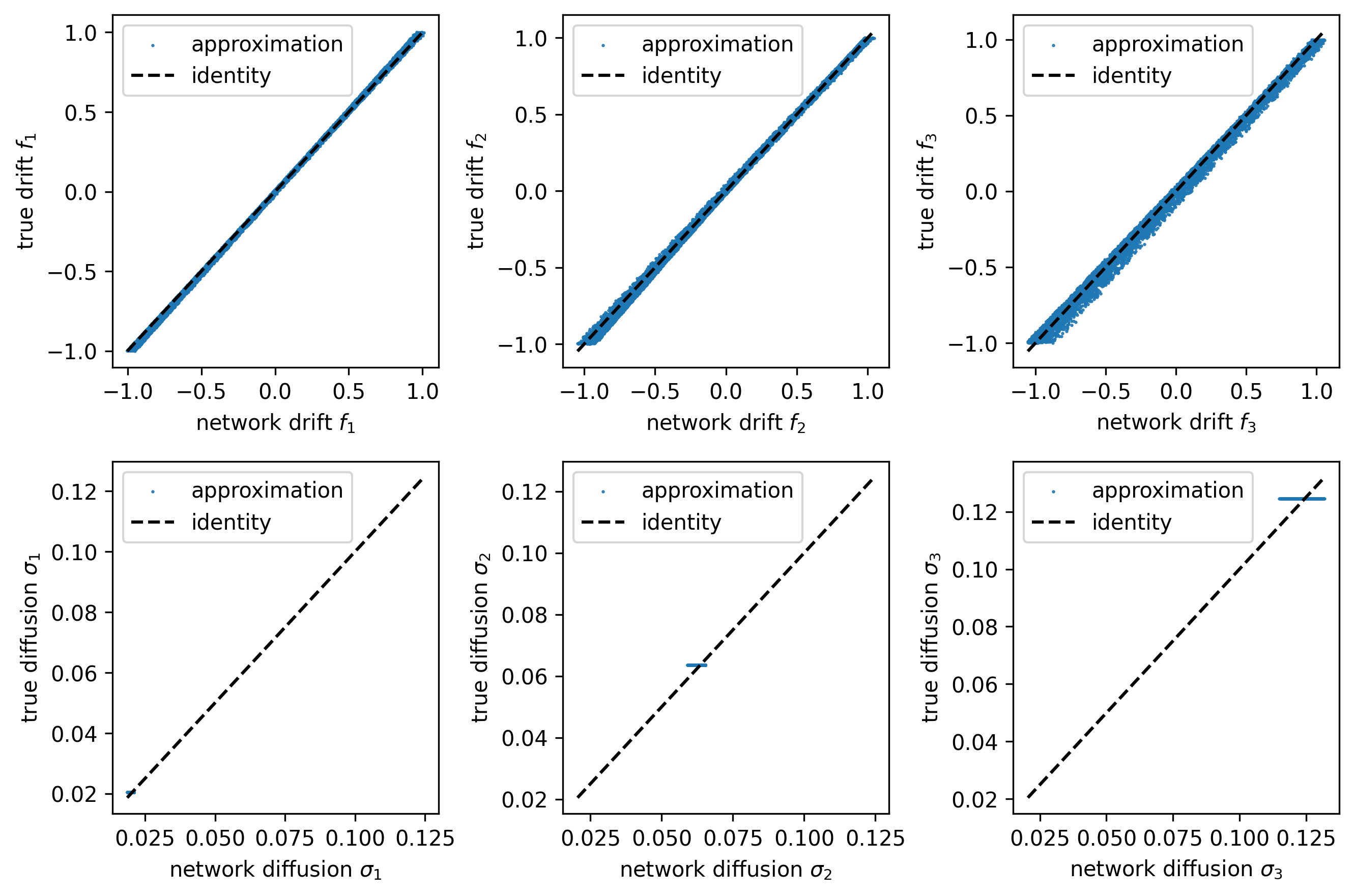}
            \subcaption{Algorithm \ref{alg:ARFF_SDE}.}
        \end{subfigure}
        \hspace{0.5cm}
        \begin{subfigure}[t]{0.45\textwidth}
            \centering
            \includegraphics[width=\textwidth]{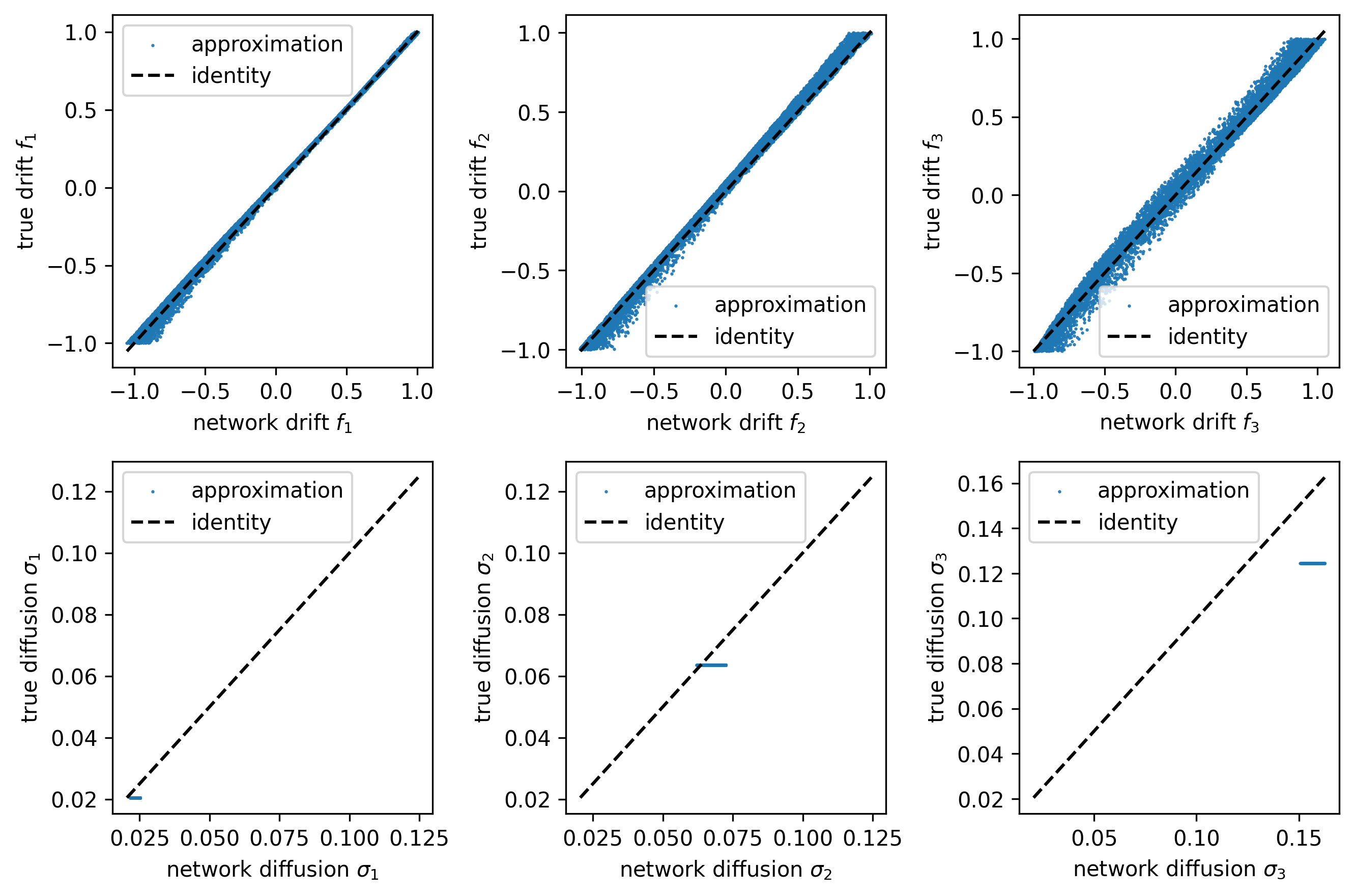}
            \subcaption{Adam optimizer.}
        \end{subfigure}
    }
    \caption{Comparison of the true and trained drift and diffusion functions for Experiment 4a: three-dimensional, symmetric diffusion.}
    \label{fig:Exp_4a_true_trained}
\end{figure}

\begin{figure}[ht!]
    \centering
    \noindent\makebox[\textwidth][c]{%
        \begin{subfigure}[t]{0.45\textwidth}
            \centering
            \includegraphics[width=\textwidth]{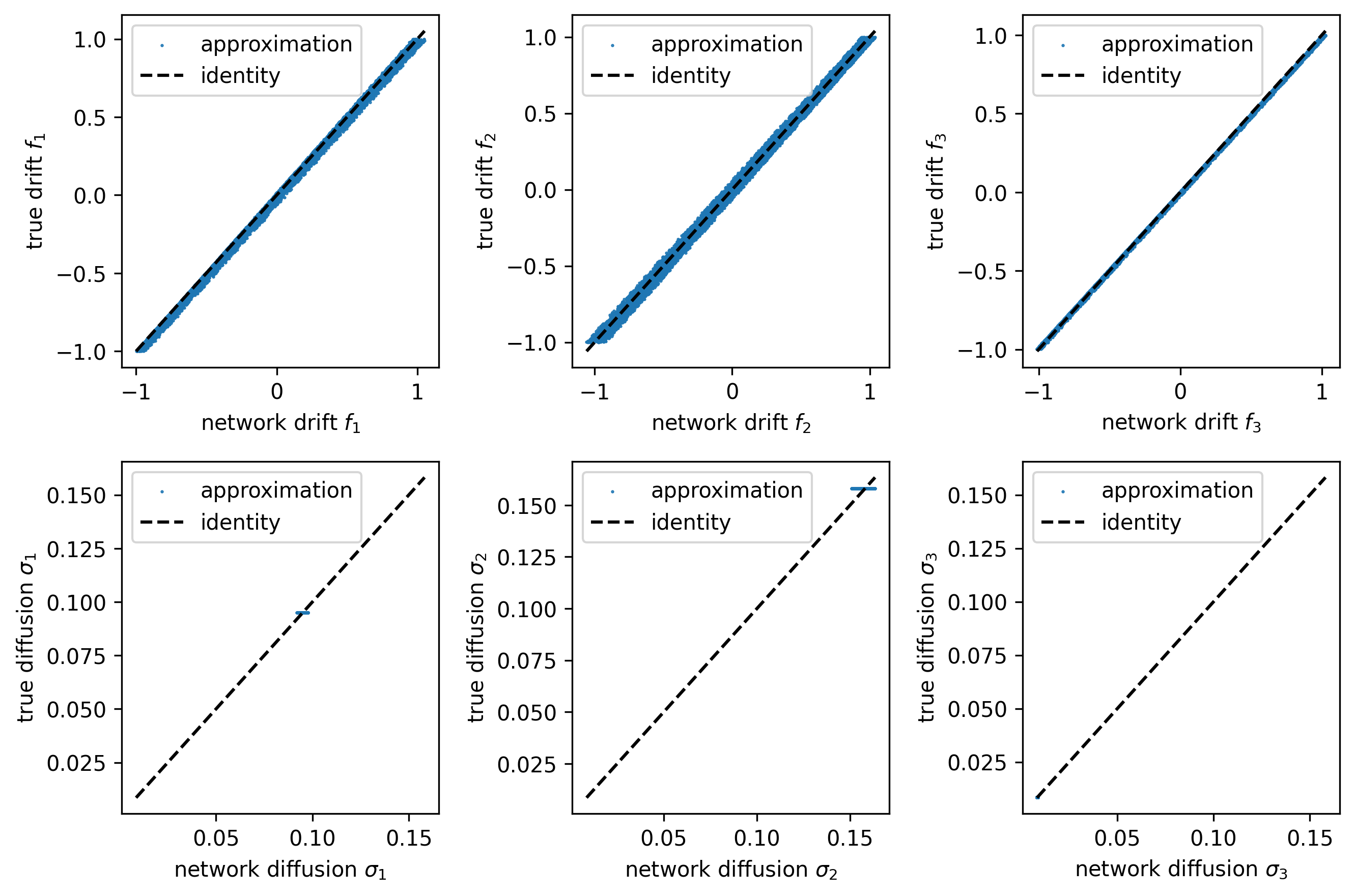}
            \subcaption{Algorithm \ref{alg:ARFF_SDE}.}
        \end{subfigure}
        \hspace{0.5cm}
        \begin{subfigure}[t]{0.45\textwidth}
            \centering
            \includegraphics[width=\textwidth]{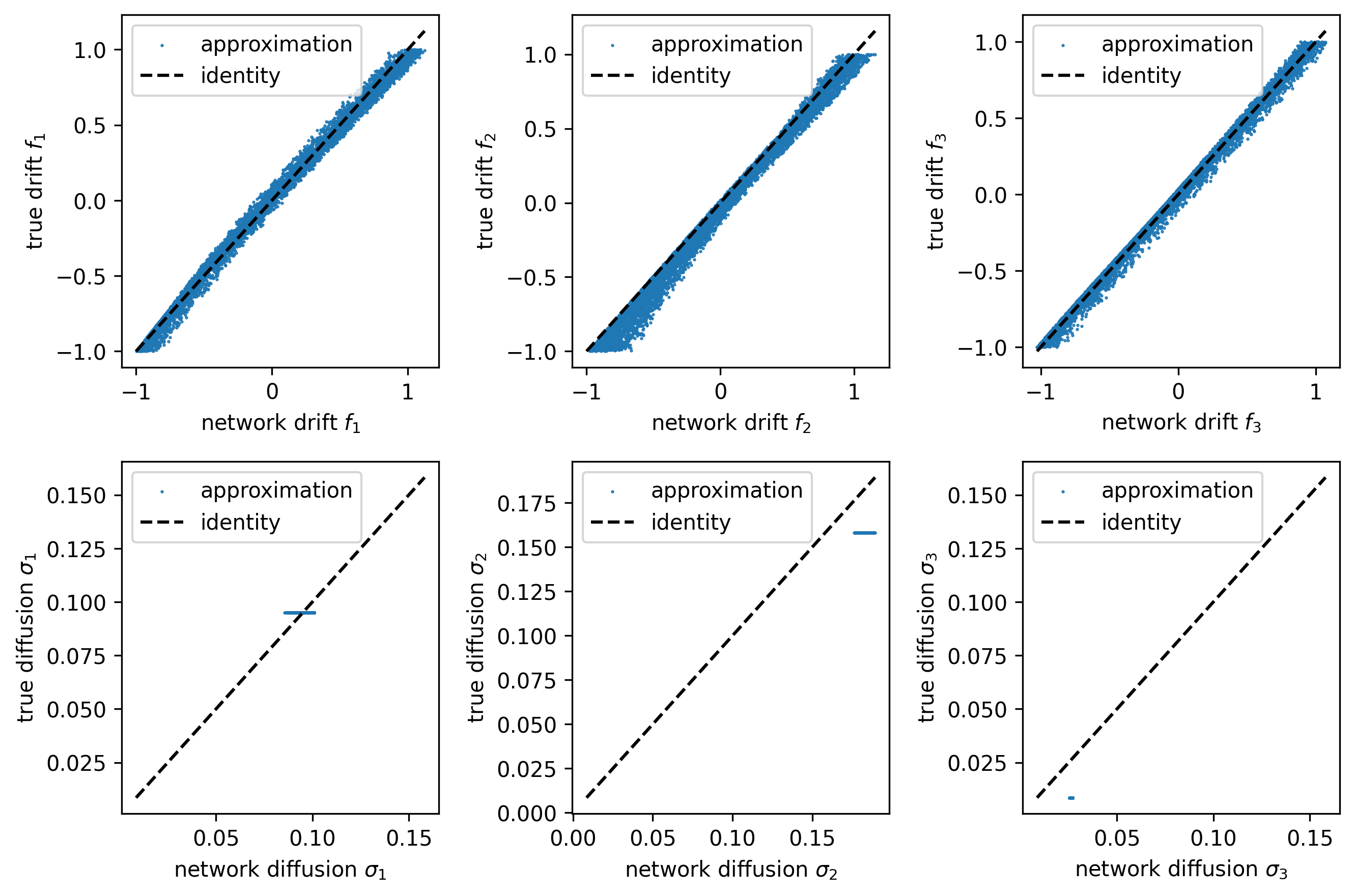}
            \subcaption{Adam optimizer.}
        \end{subfigure}
    }
    \caption{Comparison of the true and trained drift and diffusion functions for Experiment 4b: three-dimensional, lower-triangular diffusion.}
    \label{fig:Exp_4b_true_trained}
\end{figure}

\begin{figure}[ht!]
    \centering
    \noindent\makebox[\textwidth][c]{%
        \begin{subfigure}[t]{0.45\textwidth}
            \centering
            \includegraphics[width=\textwidth]{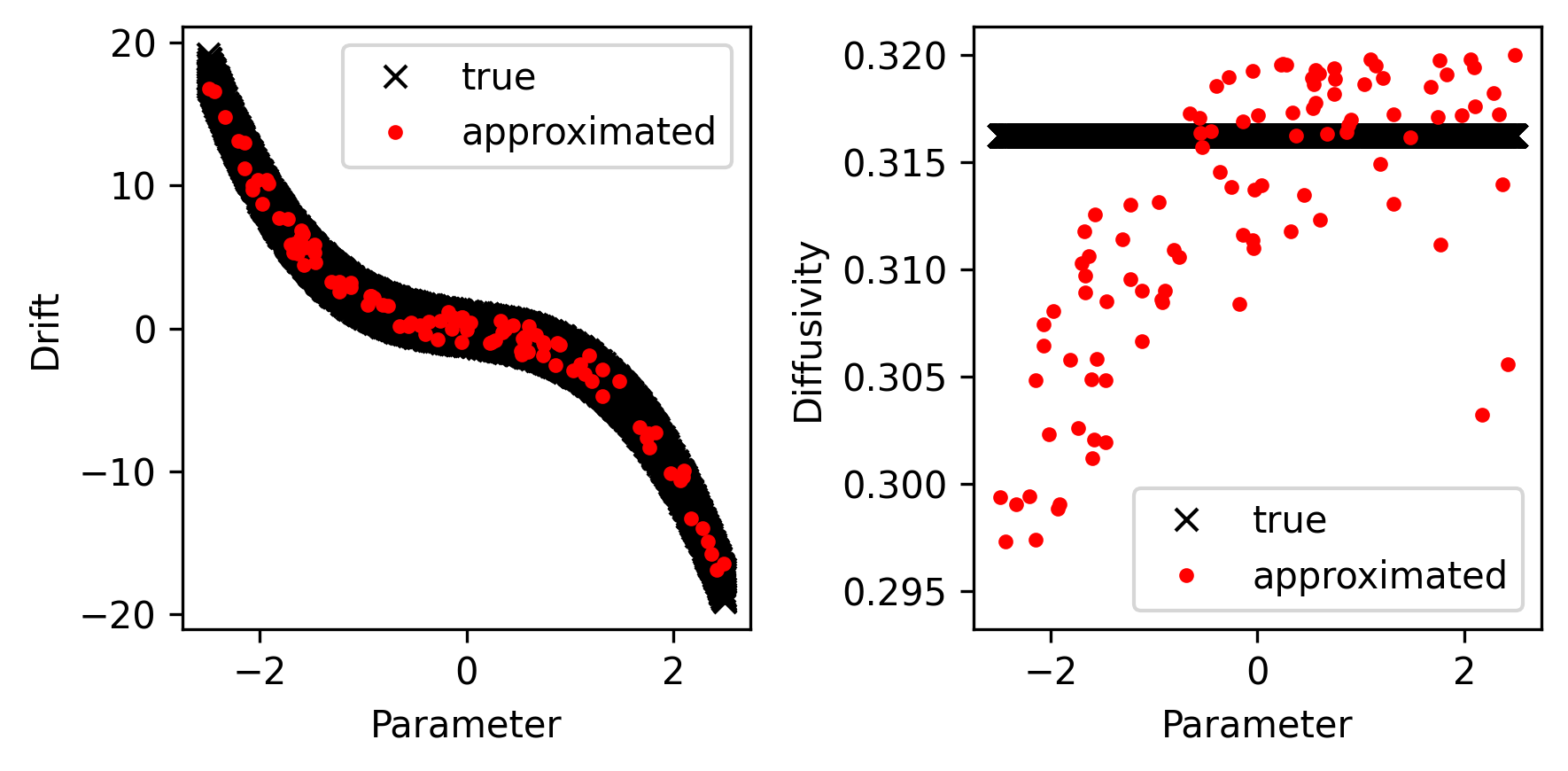}
            \subcaption{Algorithm \ref{alg:ARFF_SDE}.}
        \end{subfigure}
        \hspace{0.5cm}
        \begin{subfigure}[t]{0.45\textwidth}
            \centering
            \includegraphics[width=\textwidth]{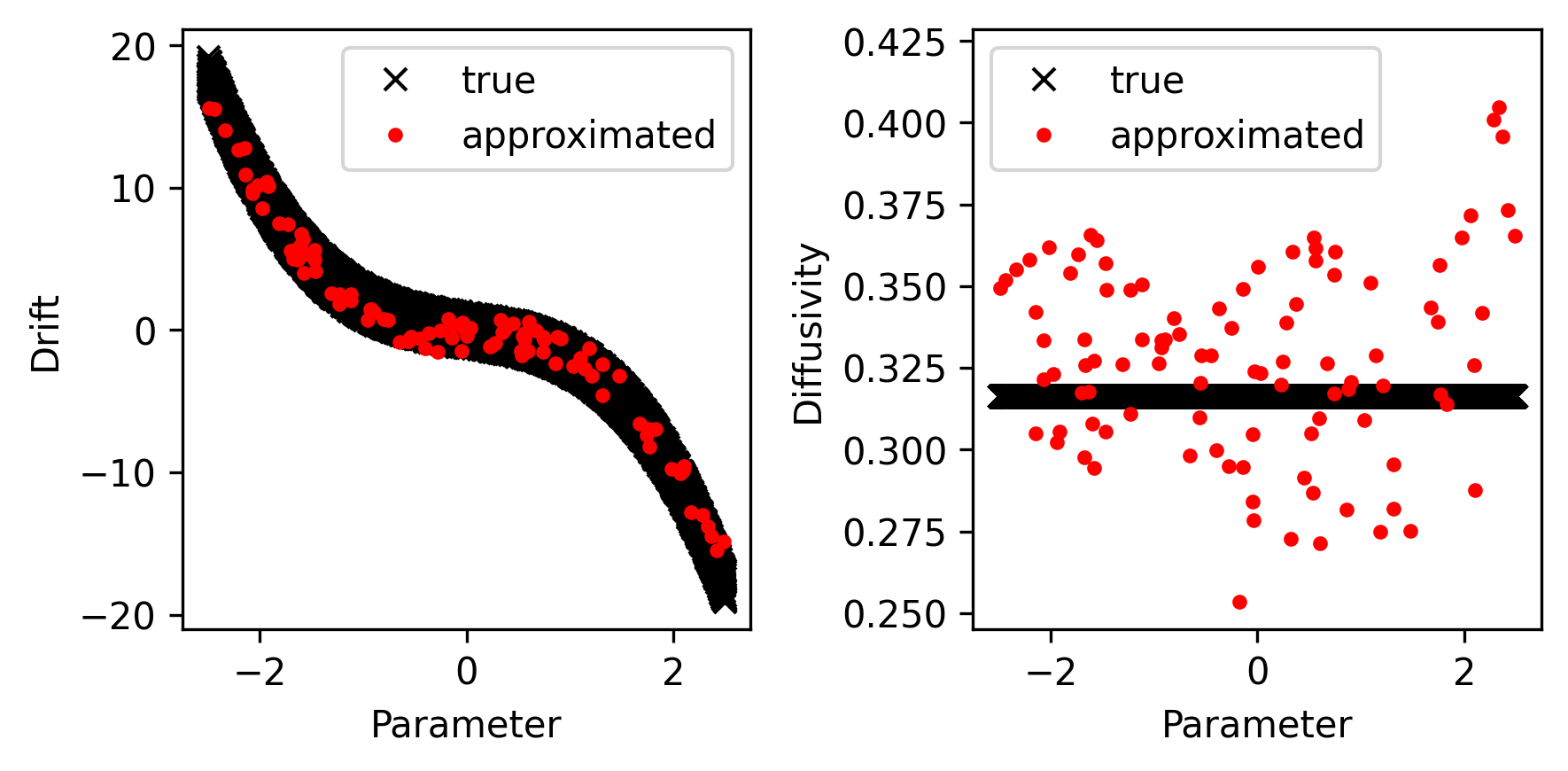}
            \subcaption{Adam optimizer.}
        \end{subfigure}
    }
    \caption{Comparison of the true and trained drift and diffusion functions for Experiment 5: underdamped Langevin.}
    \label{fig:Exp_5_true_trained}
\end{figure}

\begin{figure}[ht!]
    \centering
    \noindent\makebox[\textwidth][c]{%
        \begin{subfigure}[t]{0.45\textwidth}
            \centering
            \includegraphics[width=\textwidth]{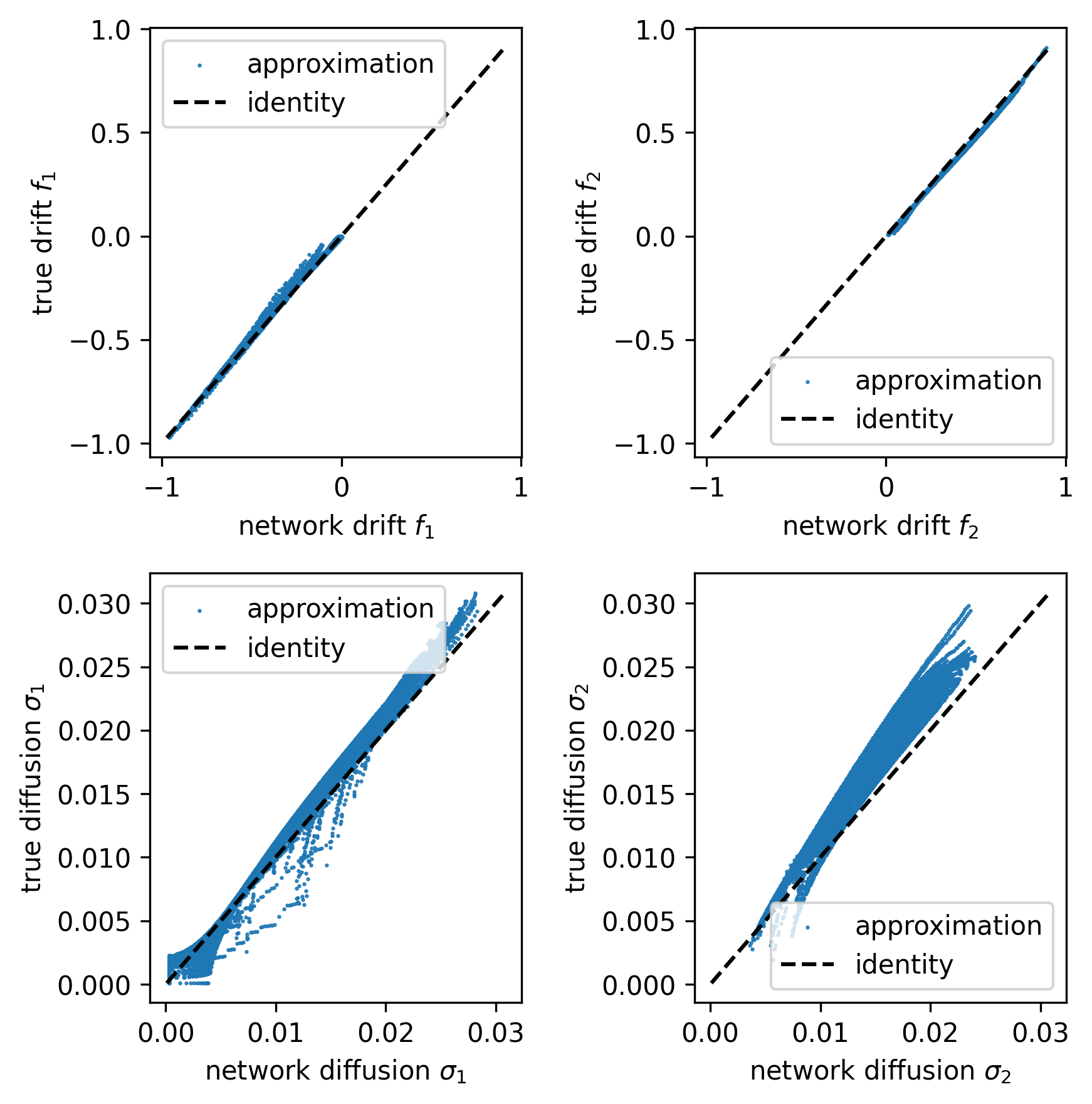}
            \subcaption{Algorithm \ref{alg:ARFF_SDE}.}
        \end{subfigure}
        \hspace{0.5cm}
        \begin{subfigure}[t]{0.45\textwidth}
            \centering
            \includegraphics[width=\textwidth]{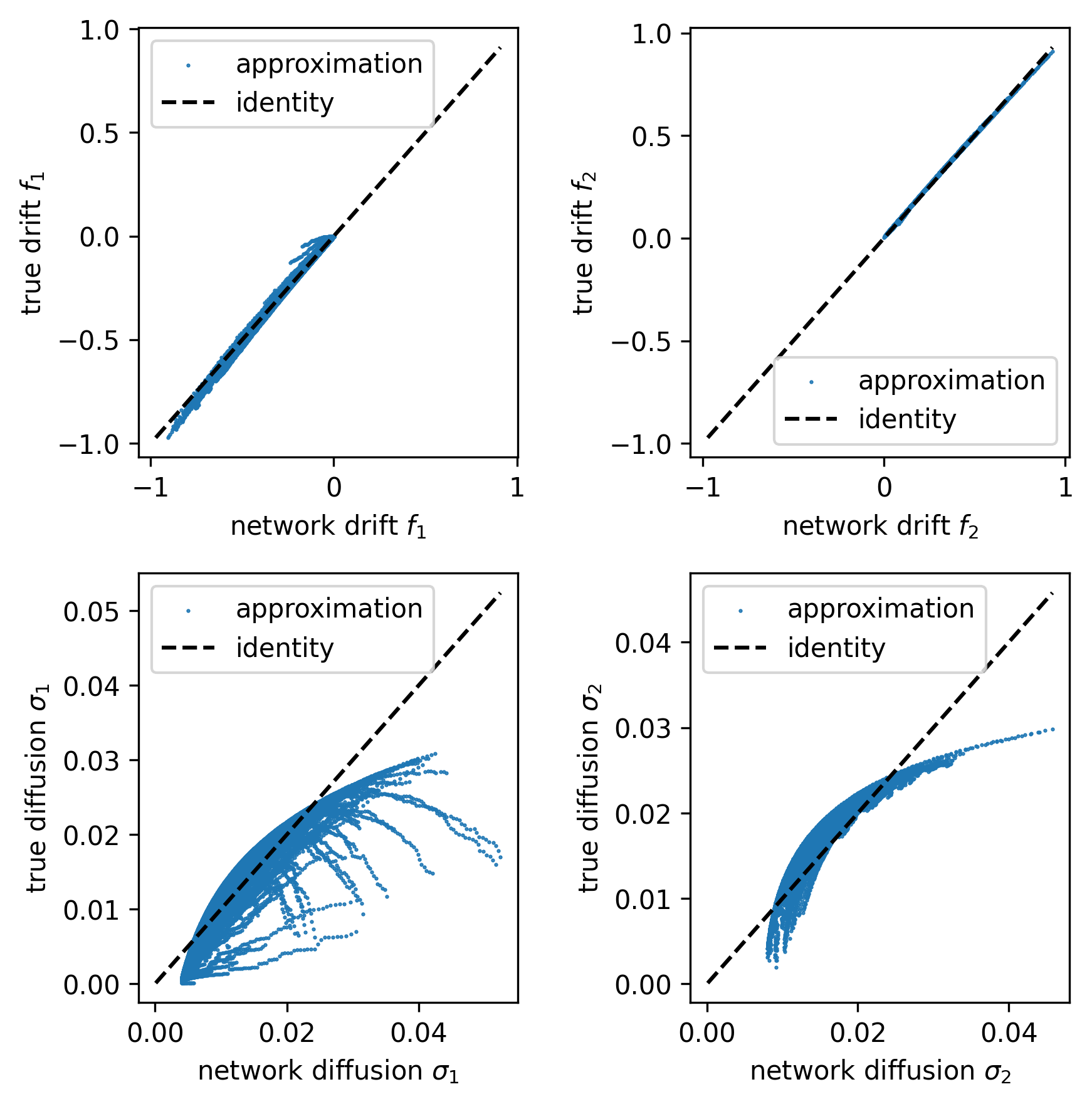}
            \subcaption{Adam optimizer.}
        \end{subfigure}
    }
    \caption{Comparison of the true and trained drift and diffusion functions for Experiment 6: susceptible, infected, recovered.}
    \label{fig:Exp_6_true_trained}
\end{figure}

\begin{figure}[ht!]
    \centering
    \noindent\makebox[\textwidth][c]{%
        \begin{subfigure}[t]{0.45\textwidth}
            \centering
            \includegraphics[width=\textwidth]{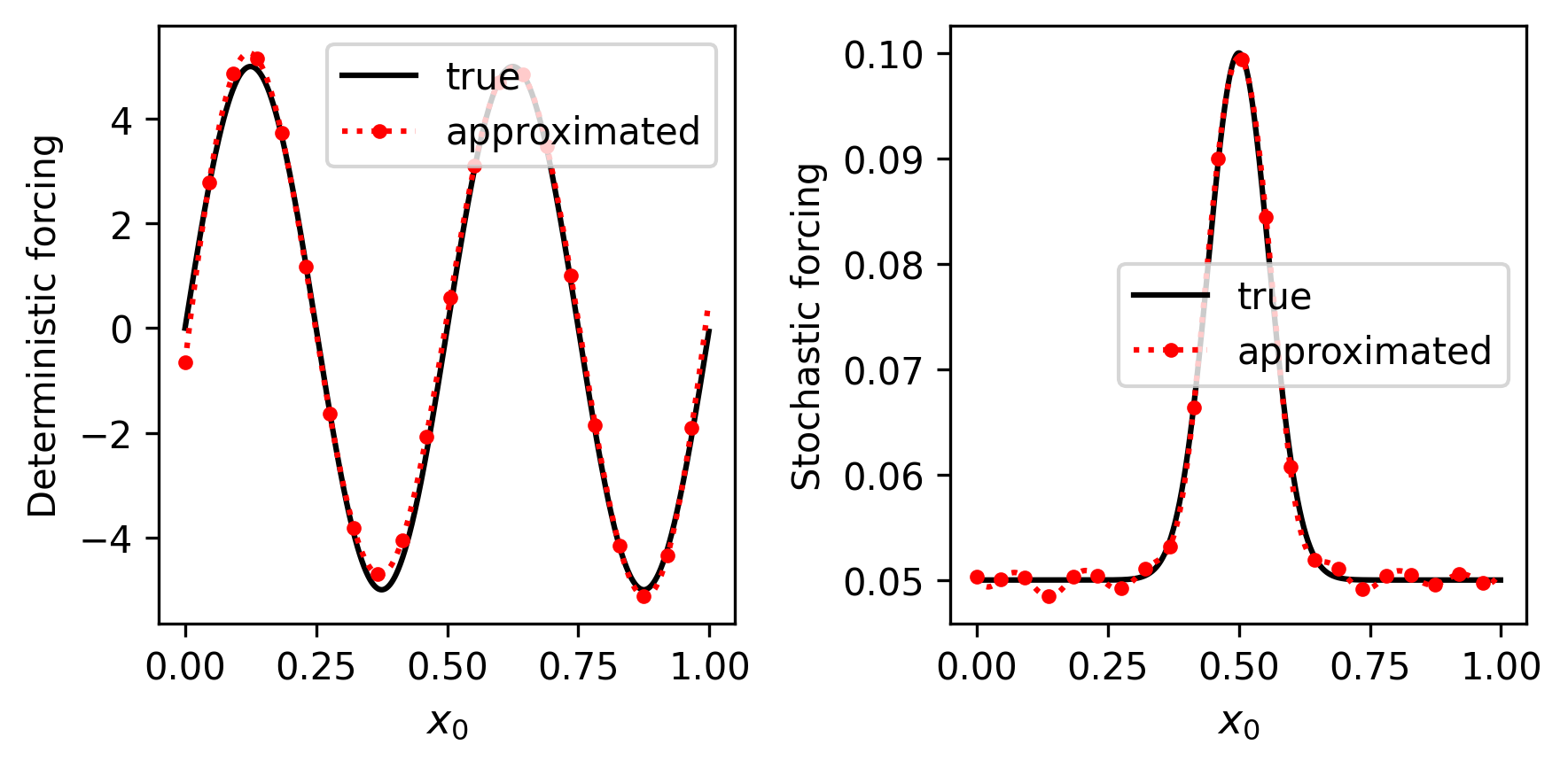}
            \subcaption{Algorithm \ref{alg:ARFF_SDE}.}
        \end{subfigure}
        \hspace{0.5cm}
        \begin{subfigure}[t]{0.45\textwidth}
            \centering
            \includegraphics[width=\textwidth]{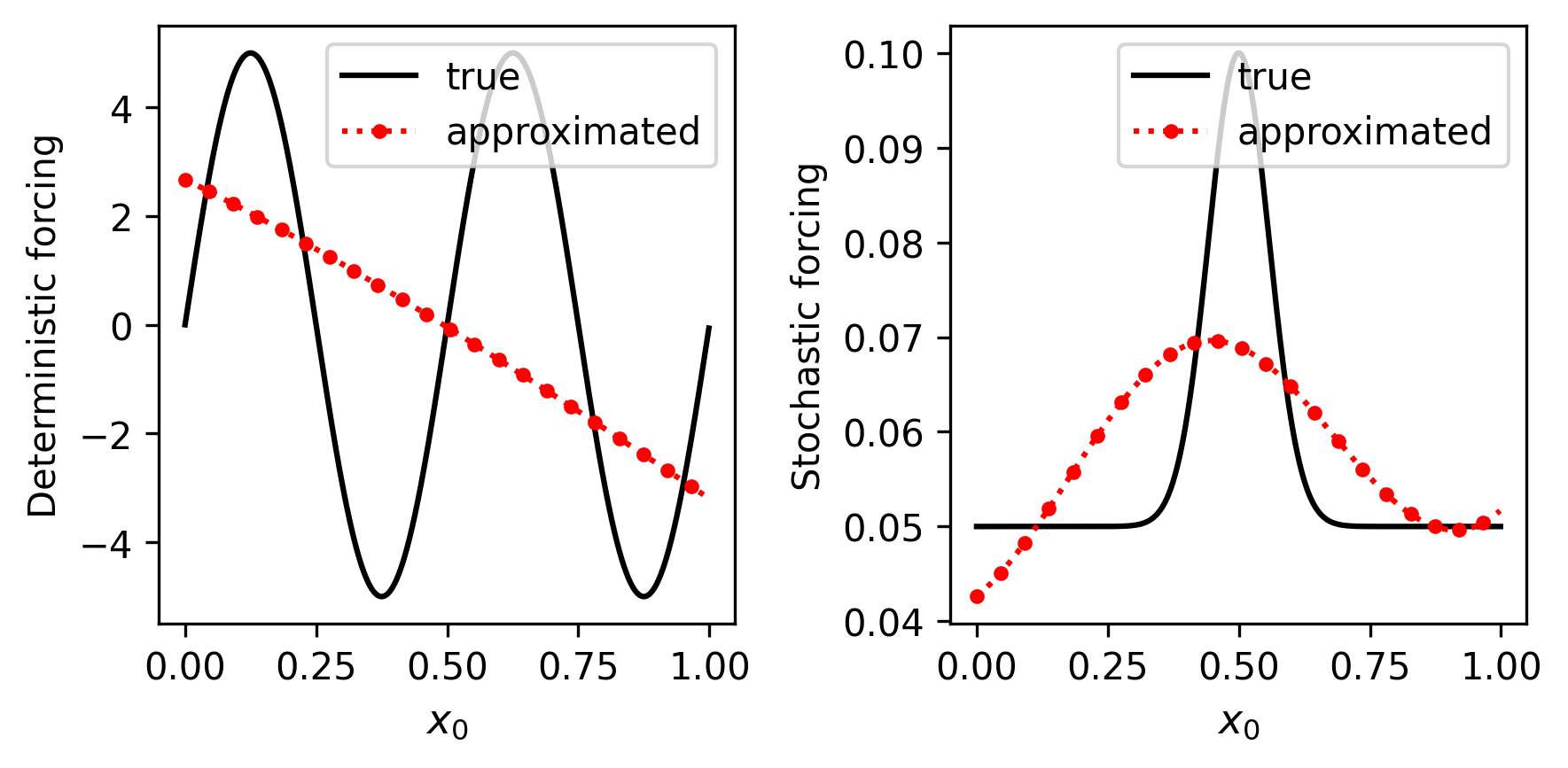}
            \subcaption{Adam optimizer.}
        \end{subfigure}
    }
    \caption{Comparison of the true and trained drift and diffusion functions for Experiment 7: stochastic wave equation.}
    \label{fig:Exp_7_true_trained}
\end{figure}

Figures \ref{fig:Exp_1_true_trained} to \ref{fig:Exp_7_true_trained} compare the true and trained drift and diffusion functions for Experiments 1 to 7. The trained functions are the network configurations that resulted in the minimum validation loss in the training procedure. Access to the structural information required to calculate the diffusion $\sigma_{\theta'}(x_0)$ from the diffusion covariance $\Sigma_{\theta'}$ is assumed.

Generally, both Algorithm \ref{alg:ARFF_SDE} and the Adam optimizer converge toward the true underlying dynamics, except in Experiment 7, where Adam-optimization represents a complete failure case. In all experiments (Figures \ref{fig:Exp_1_true_trained} to \ref{fig:Exp_7_true_trained}), Algorithm \ref{alg:ARFF_SDE} provides a better visual recreation of the true drift. The same is true for the diffusion across Experiments 2, 4a, 4b, 5, 6 and 7 (Figures \ref{fig:Exp_2_true_trained}, \ref{fig:Exp_4a_true_trained}, \ref{fig:Exp_4b_true_trained}, \ref{fig:Exp_5_true_trained}, \ref{fig:Exp_6_true_trained} and \ref{fig:Exp_7_true_trained}). The Adam optimizer only provides a more accurate recreation of the diffusion in Experiment 1 (Figure \ref{fig:Exp_1_true_trained}). Experiments 3a and 4b (Figures \ref{fig:Exp_3a_true_trained} and \ref{fig:Exp_3b_true_trained}) exhibit comparable diffusion performance for both training methods.  

\vspace{1em}

\begin{figure}[ht!]
    \centering
    \noindent\makebox[\textwidth][c]{%
        \begin{subfigure}[t]{0.35\textwidth}
            \centering
            \includegraphics[width=\textwidth]{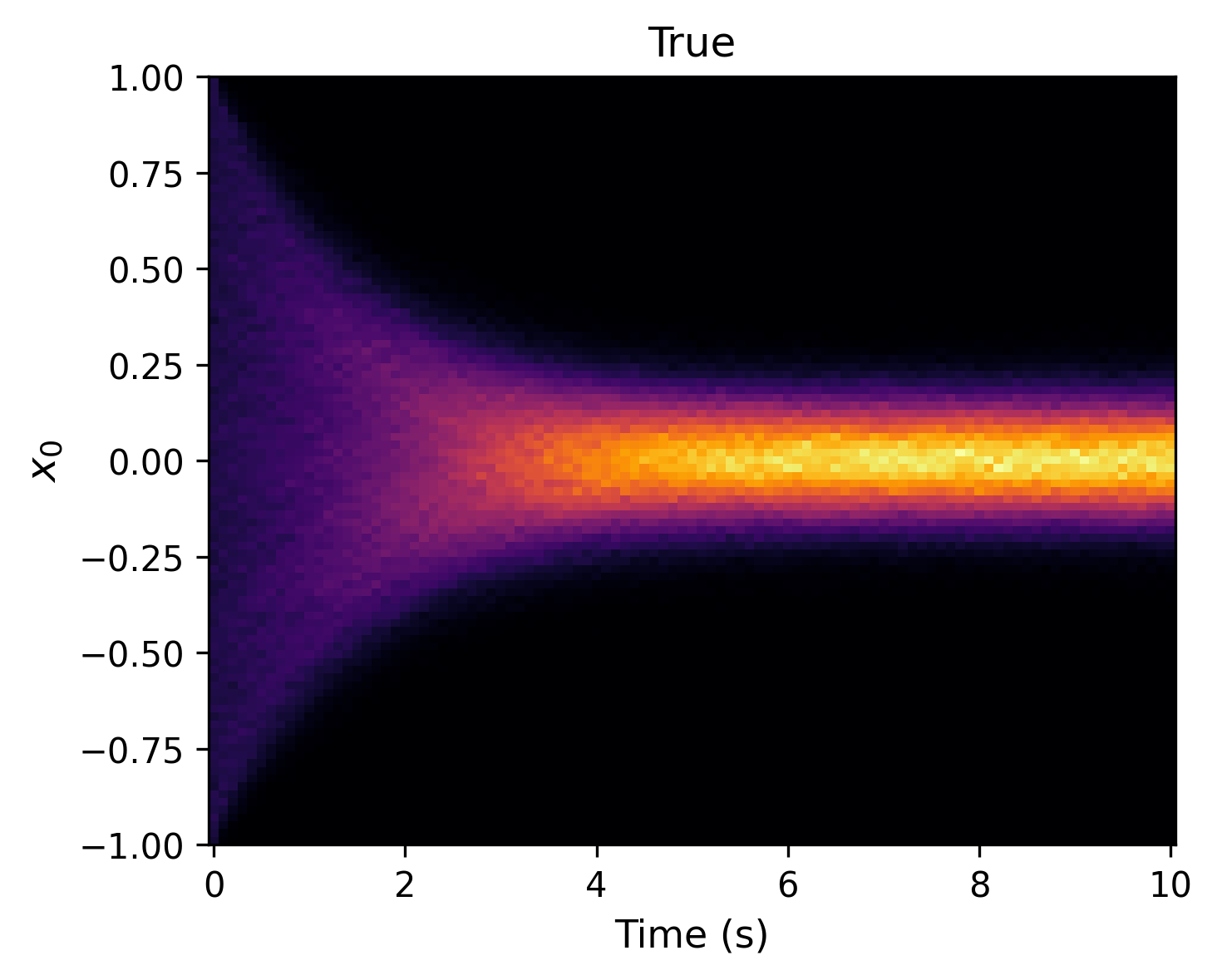}
        \end{subfigure}
        \begin{subfigure}[t]{0.35\textwidth}
            \centering
            \includegraphics[width=\textwidth]{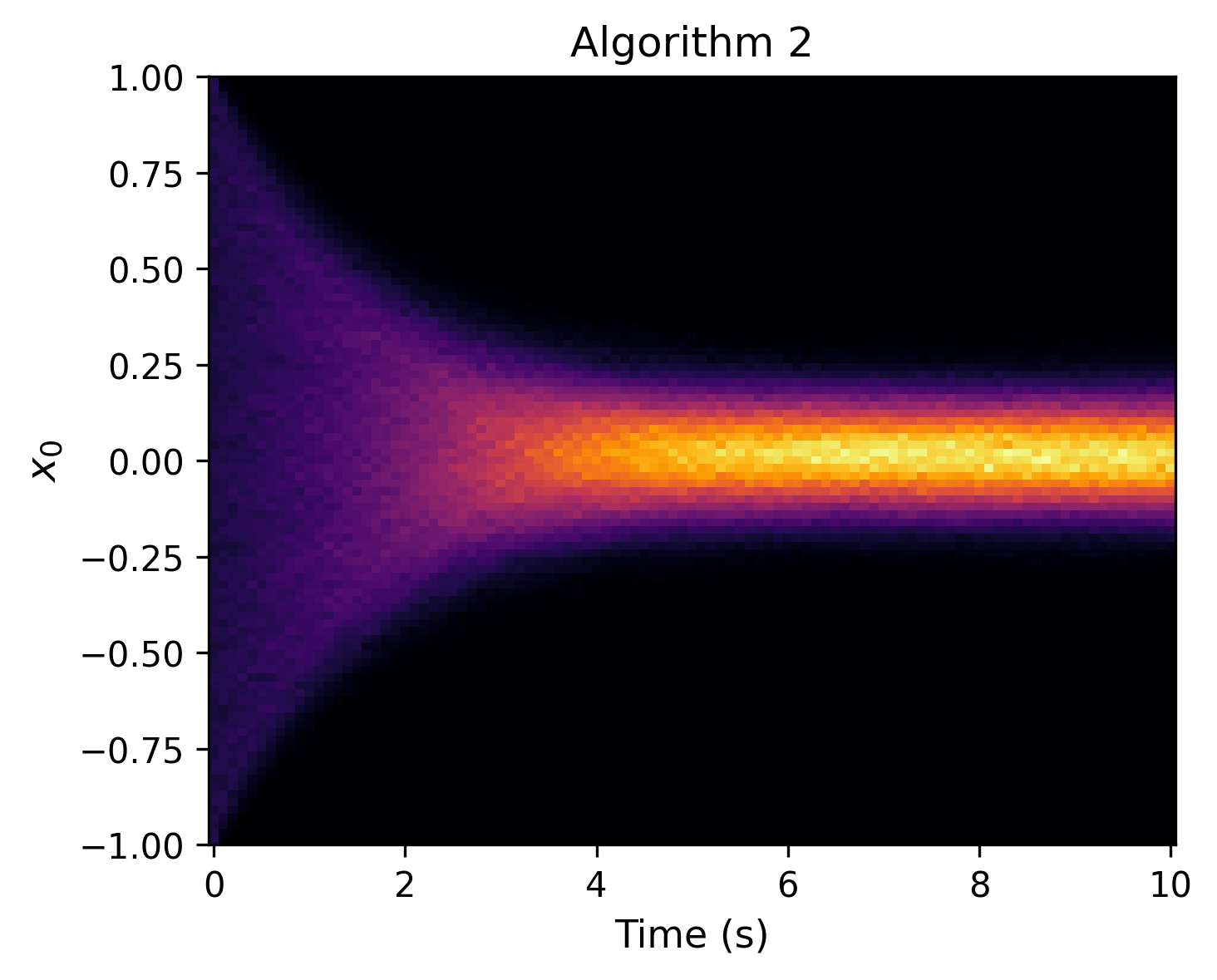}
        \end{subfigure}
        \begin{subfigure}[t]{0.35\textwidth}
            \centering
            \includegraphics[width=\textwidth]{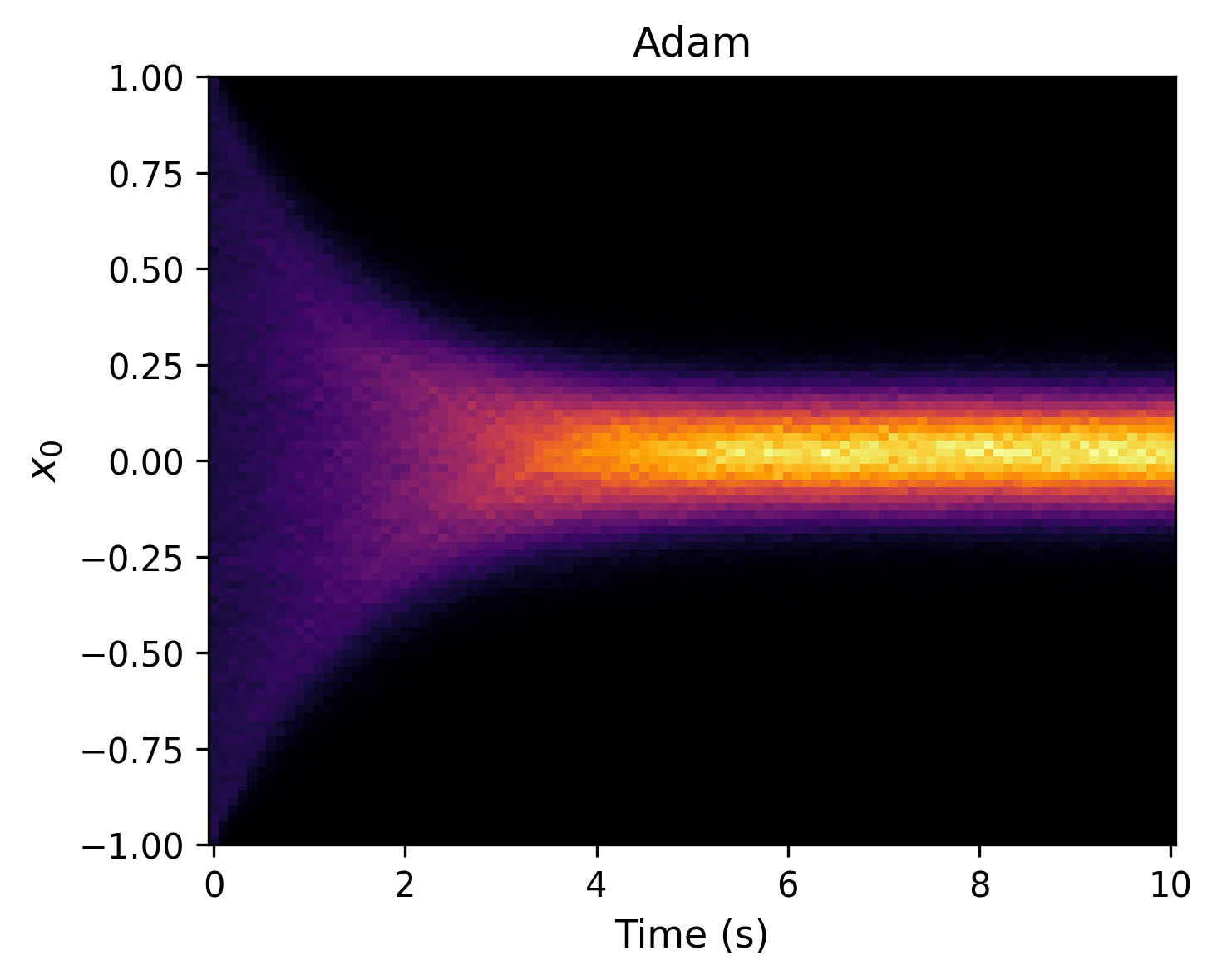}
        \end{subfigure}
    }
    \caption{Comparison of the trajectories generated using the true and trained functions for Experiment 1: one-dimensional, linear.}
    \label{fig:Exp_1_histogram}
\end{figure}

\begin{figure}[ht!]
    \centering
    \noindent\makebox[\textwidth][c]{%
        \begin{subfigure}[t]{0.35\textwidth}
            \centering
            \includegraphics[width=\textwidth]{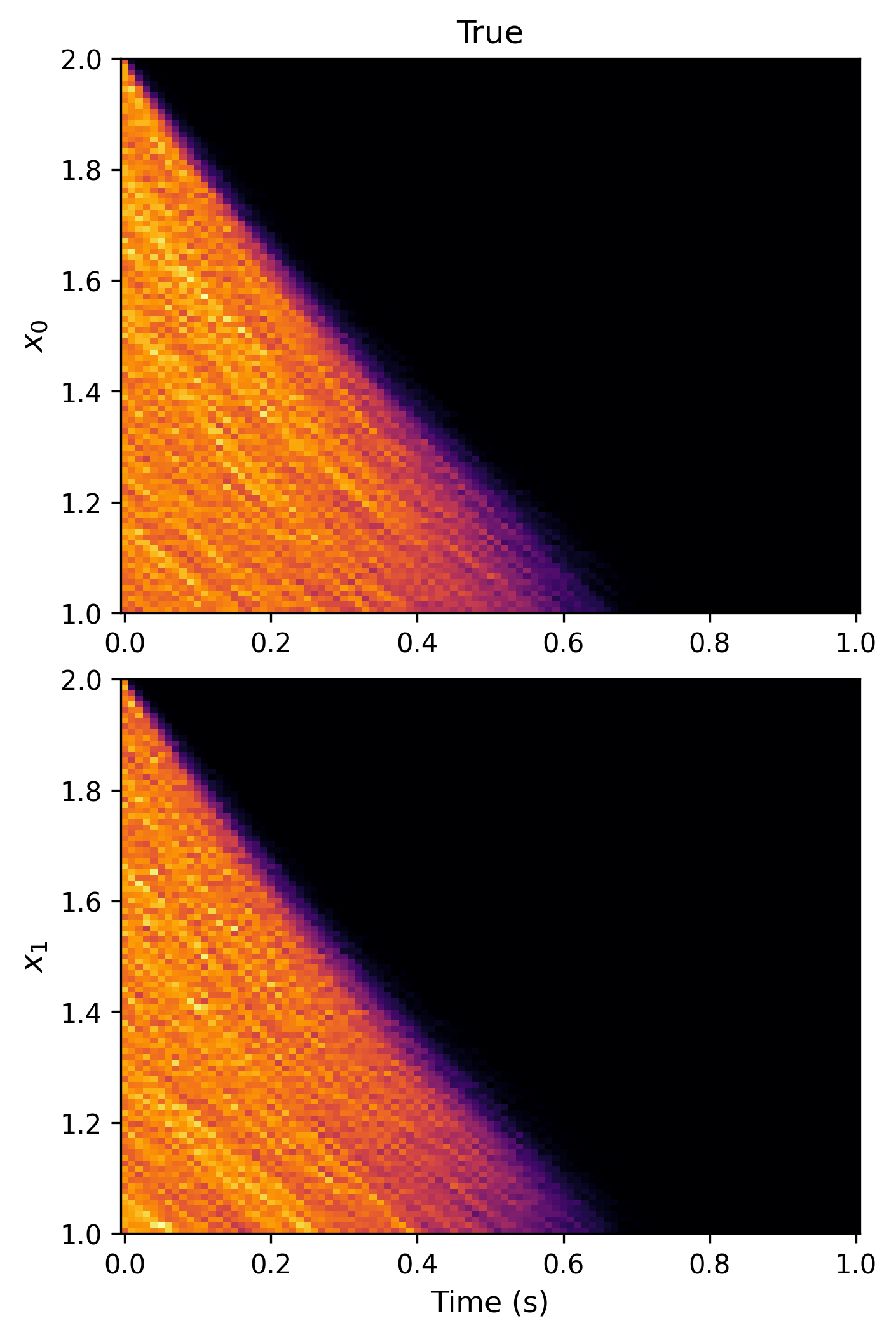}
        \end{subfigure}
        \begin{subfigure}[t]{0.35\textwidth}
            \centering
            \includegraphics[width=\textwidth]{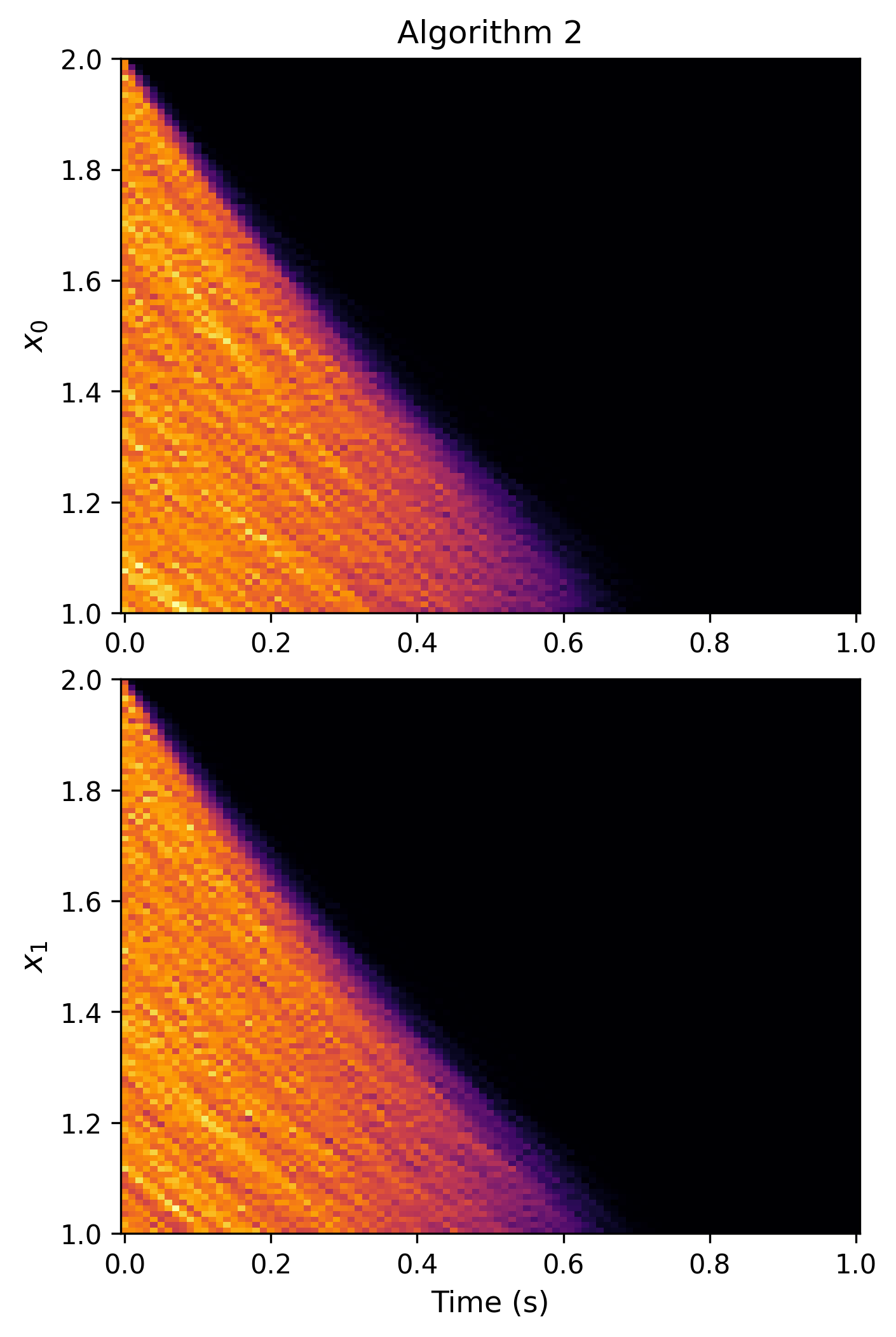}
        \end{subfigure}
        \begin{subfigure}[t]{0.35\textwidth}
            \centering
            \includegraphics[width=\textwidth]{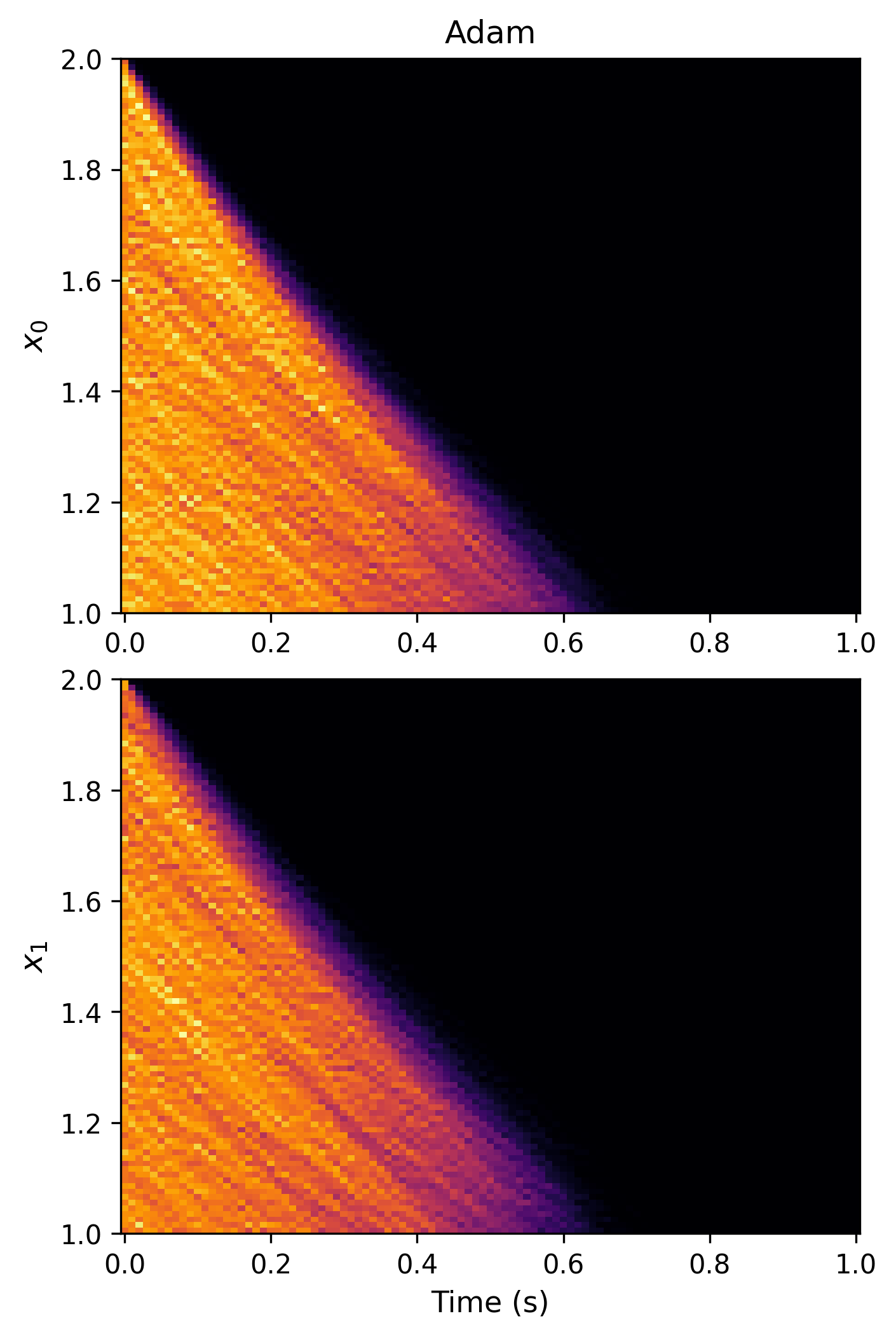}
        \end{subfigure}
    }
    \caption{Comparison of the trajectories generated using the true and trained functions for Experiment 2: two-dimensional, linear.}
    \label{fig:Exp_2_histogram}
\end{figure}

\begin{figure}[ht!]
    \centering
    \noindent\makebox[\textwidth][c]{%
        \begin{subfigure}[t]{0.35\textwidth}
            \centering
            \includegraphics[width=\textwidth]{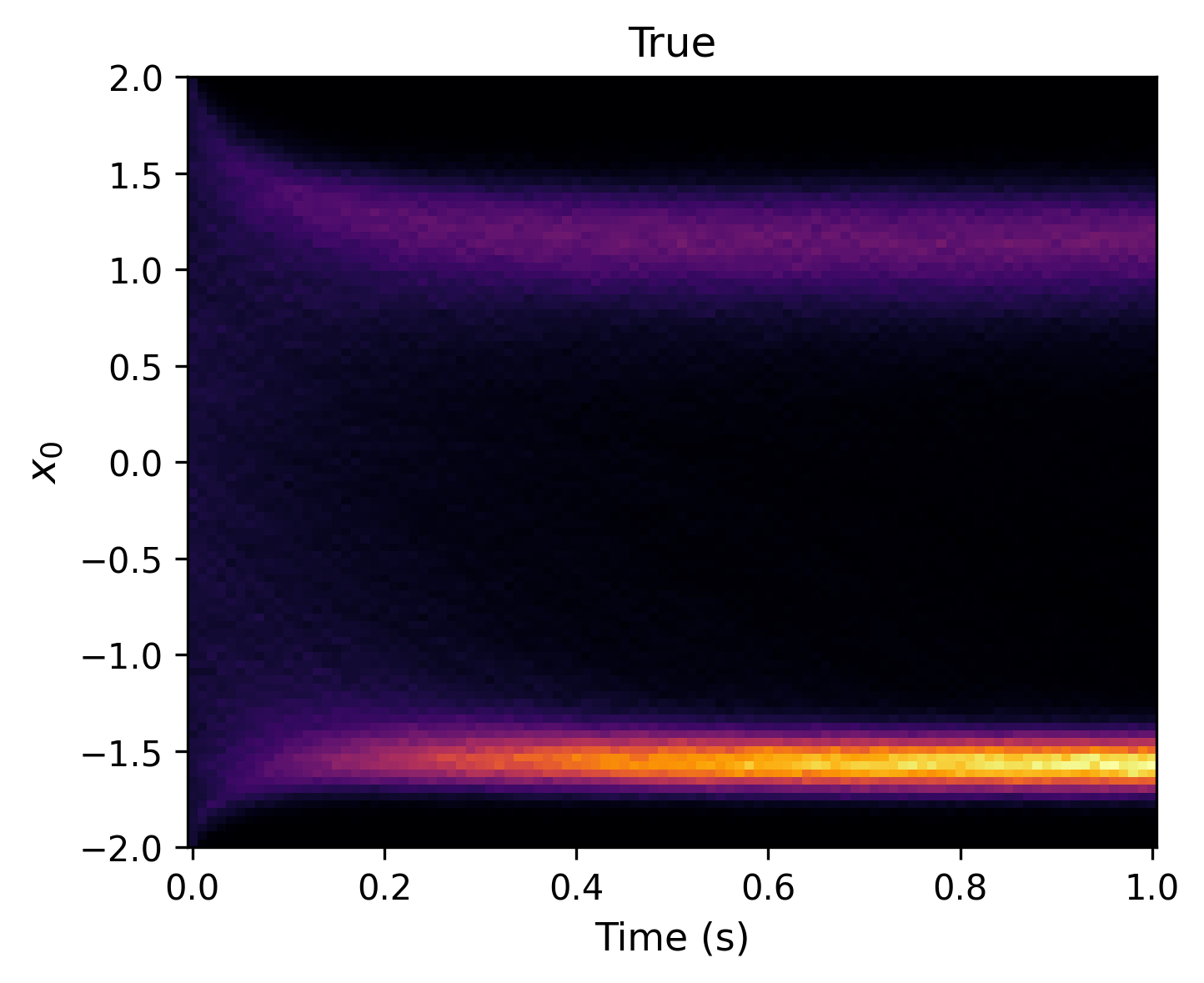}
        \end{subfigure}
        \begin{subfigure}[t]{0.35\textwidth}
            \centering
            \includegraphics[width=\textwidth]{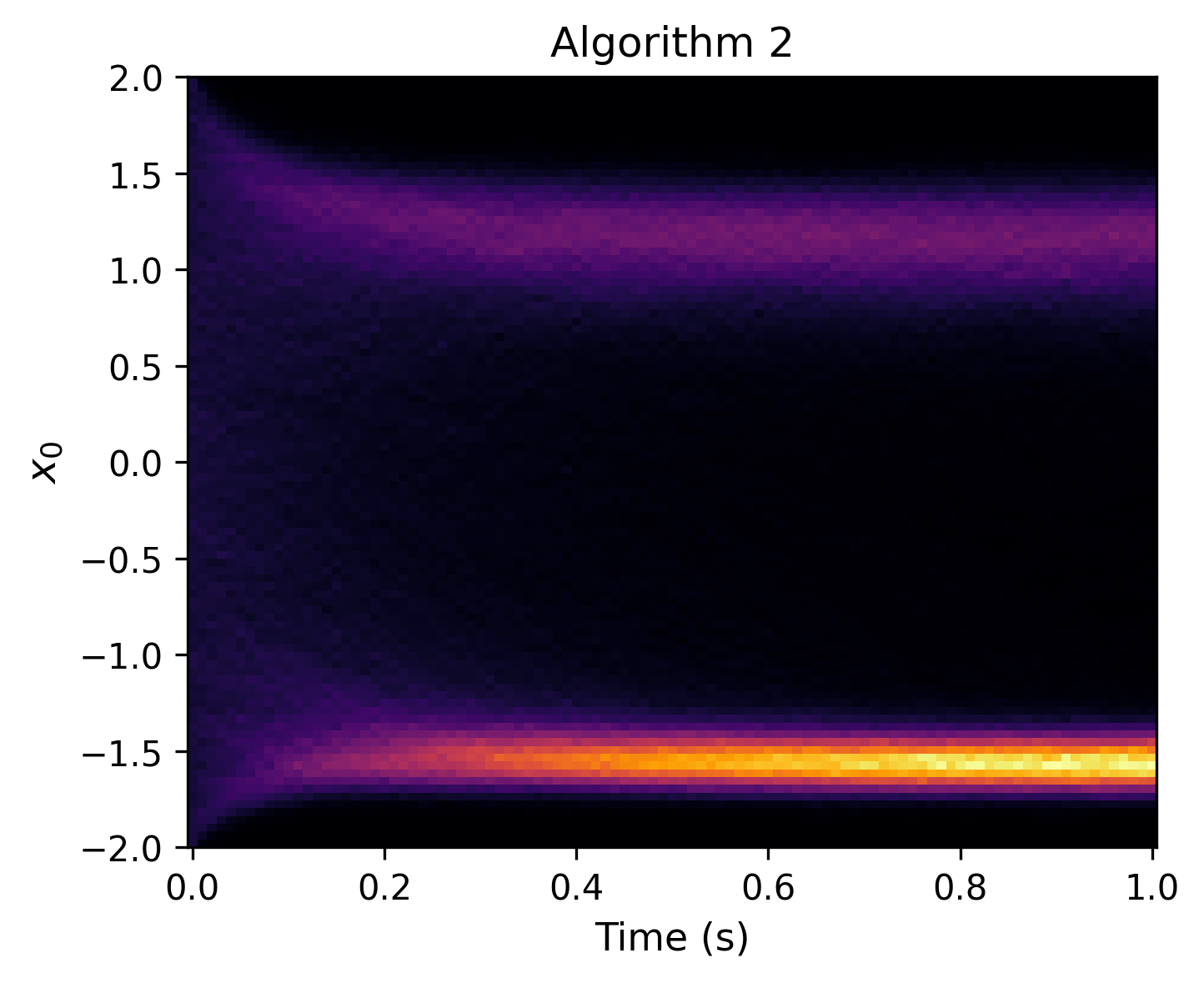}
        \end{subfigure}
        \begin{subfigure}[t]{0.35\textwidth}
            \centering
            \includegraphics[width=\textwidth]{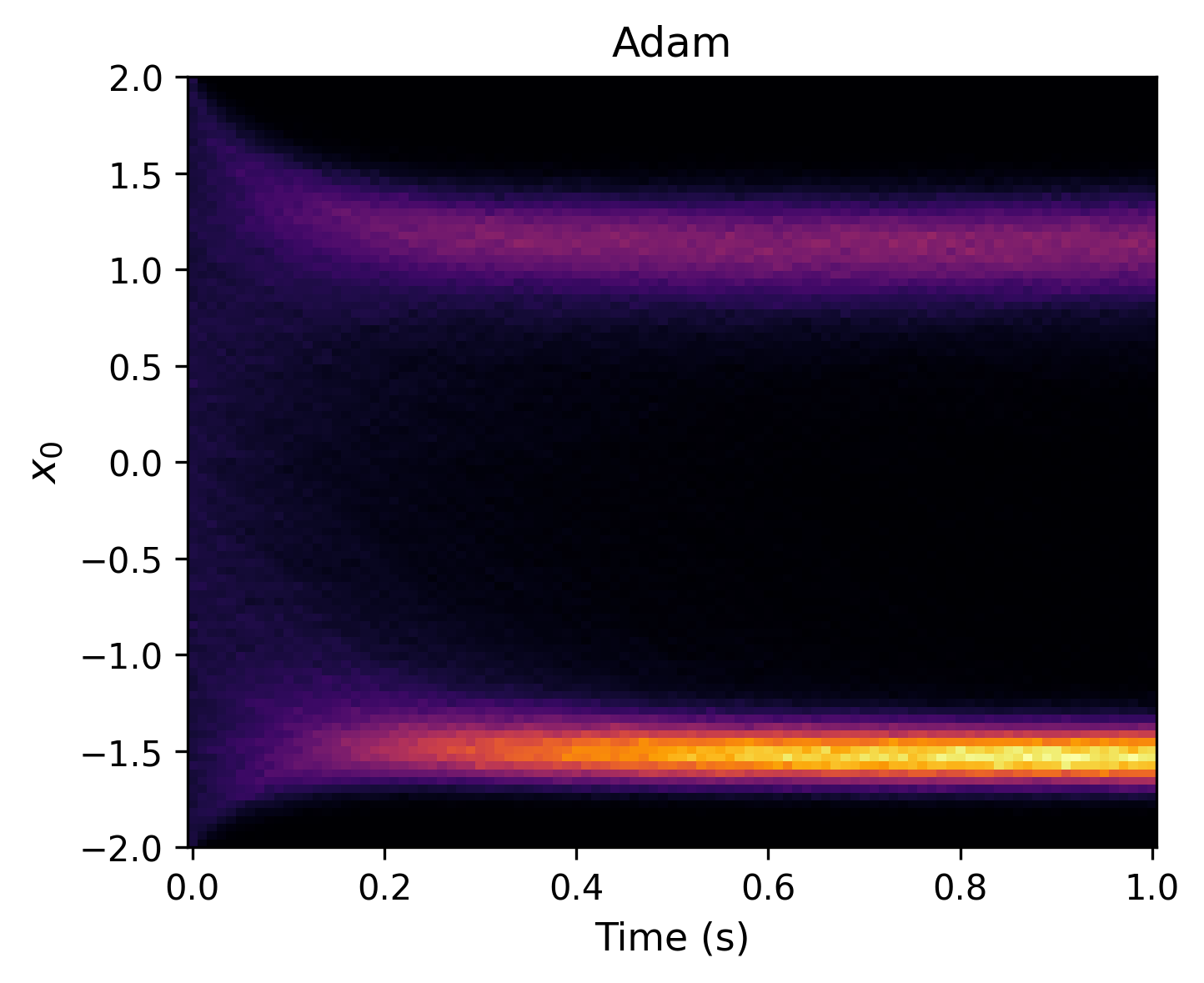}
        \end{subfigure}
    }
    \caption{Comparison of the trajectories generated using the true and trained functions for Experiment 3a: one-dimensional, cubic.}
    \label{fig:Exp_3a_histogram}
\end{figure}

\begin{figure}[ht!]
    \centering
    \noindent\makebox[\textwidth][c]{%
        \begin{subfigure}[t]{0.35\textwidth}
            \centering
            \includegraphics[width=\textwidth]{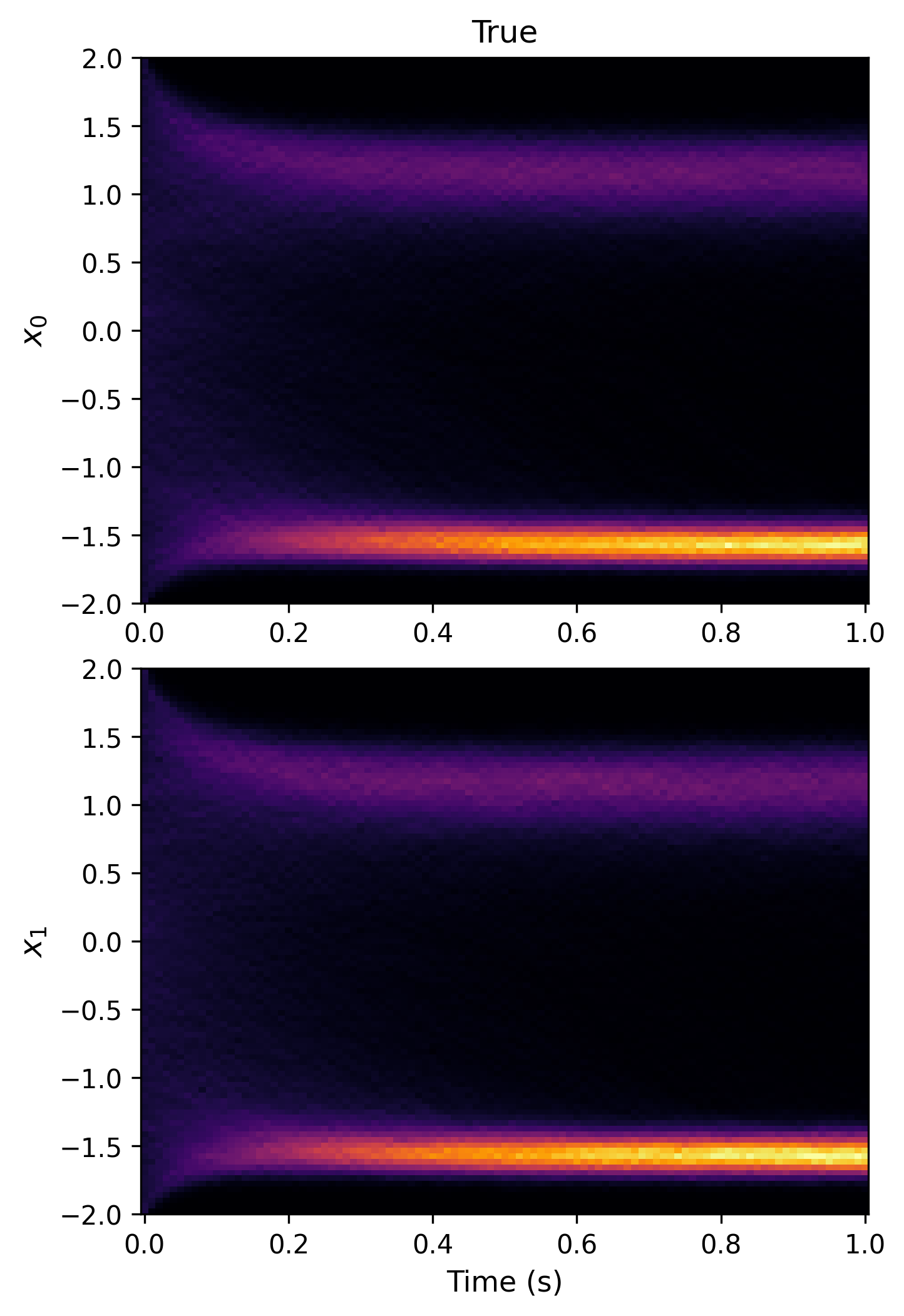}
        \end{subfigure}
        \begin{subfigure}[t]{0.35\textwidth}
            \centering
            \includegraphics[width=\textwidth]{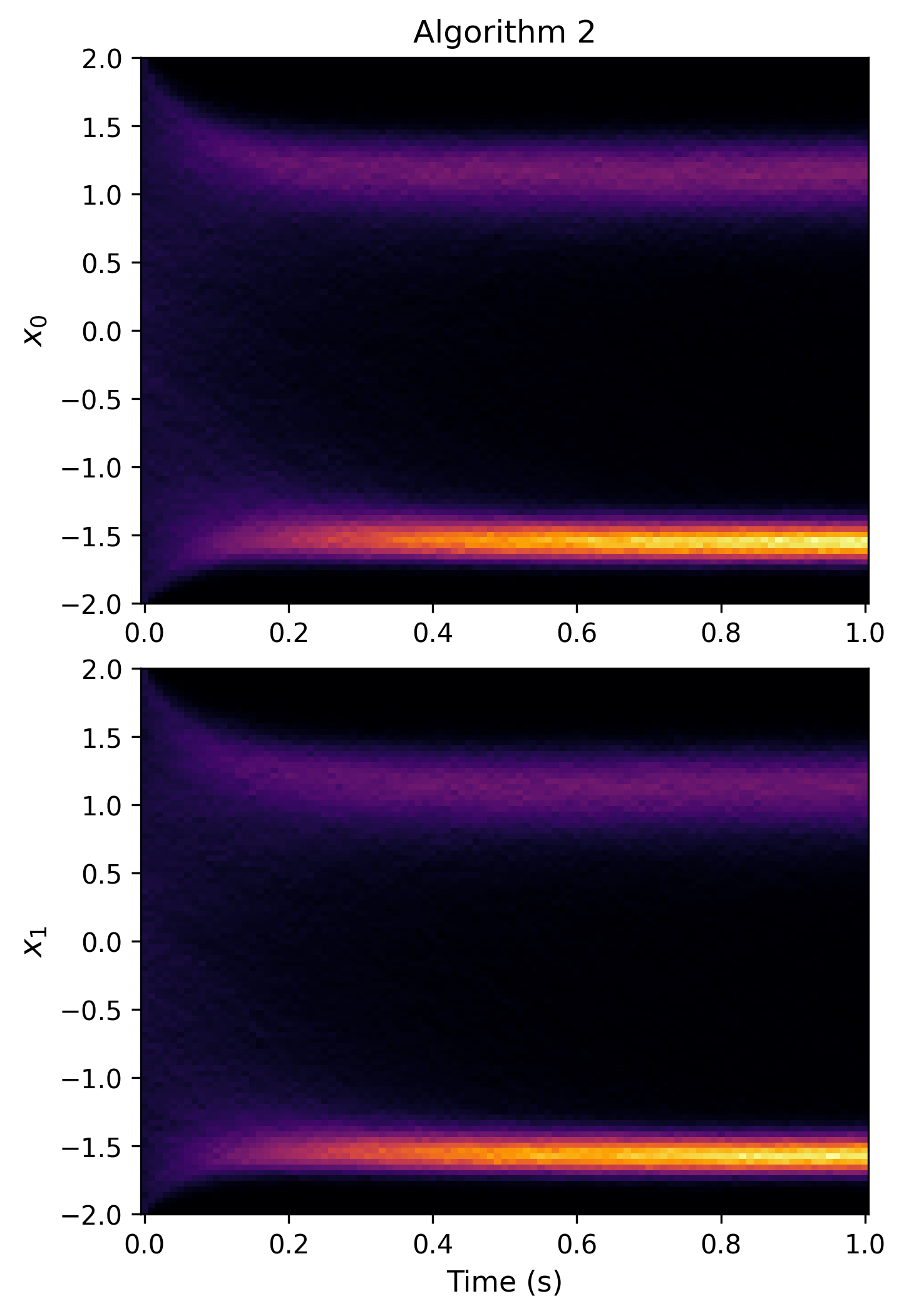}
        \end{subfigure}
        \begin{subfigure}[t]{0.35\textwidth}
            \centering
            \includegraphics[width=\textwidth]{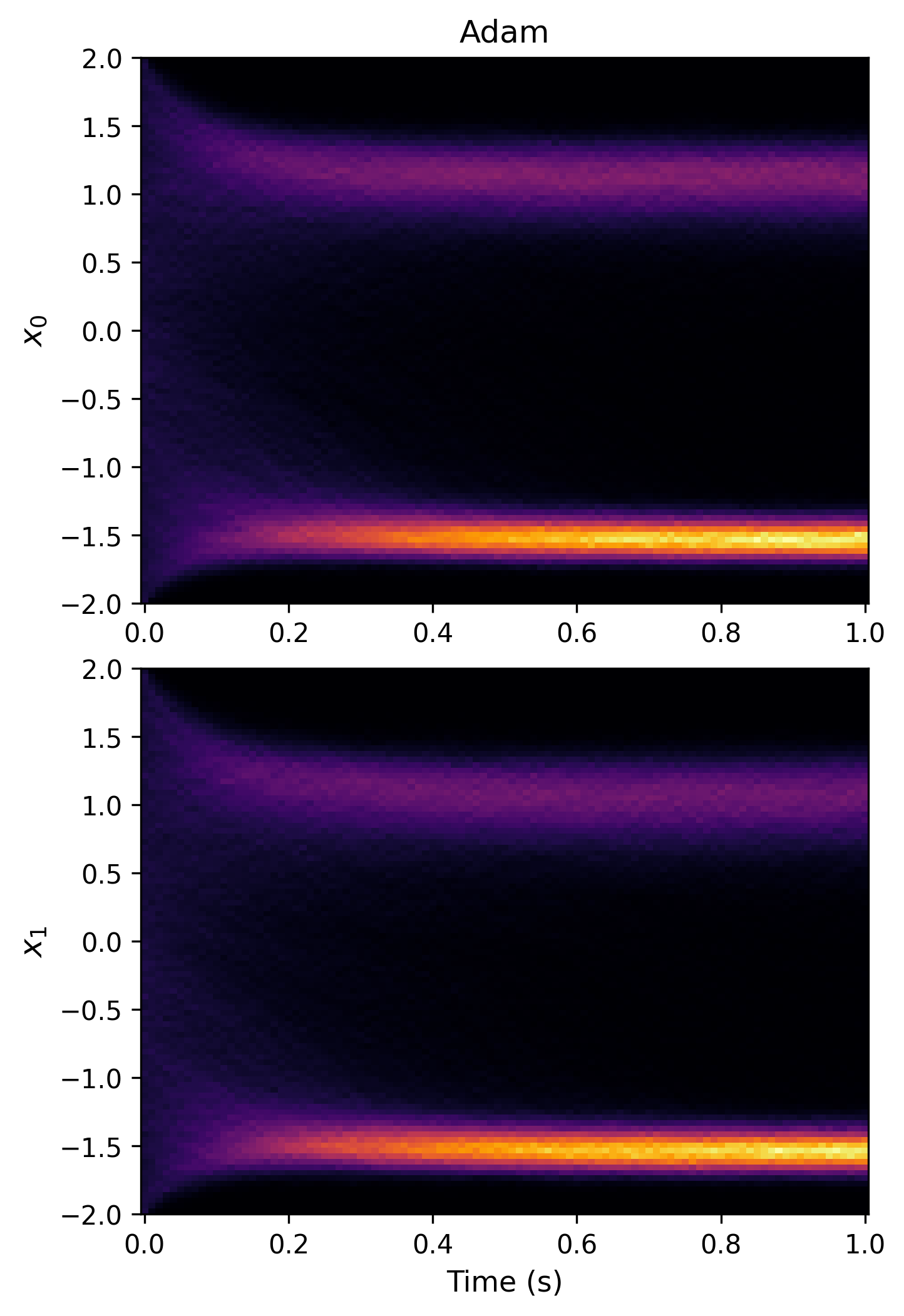}
        \end{subfigure}
    }
    \caption{Comparison of the trajectories generated using the true and trained functions for Experiment 3b: two-dimensional, cubic.}
    \label{fig:Exp_3b_histogram}
\end{figure}

\begin{figure}[ht!]
    \centering
    \noindent\makebox[\textwidth][c]{%
        \begin{subfigure}[t]{0.35\textwidth}
            \centering
            \includegraphics[width=\textwidth]{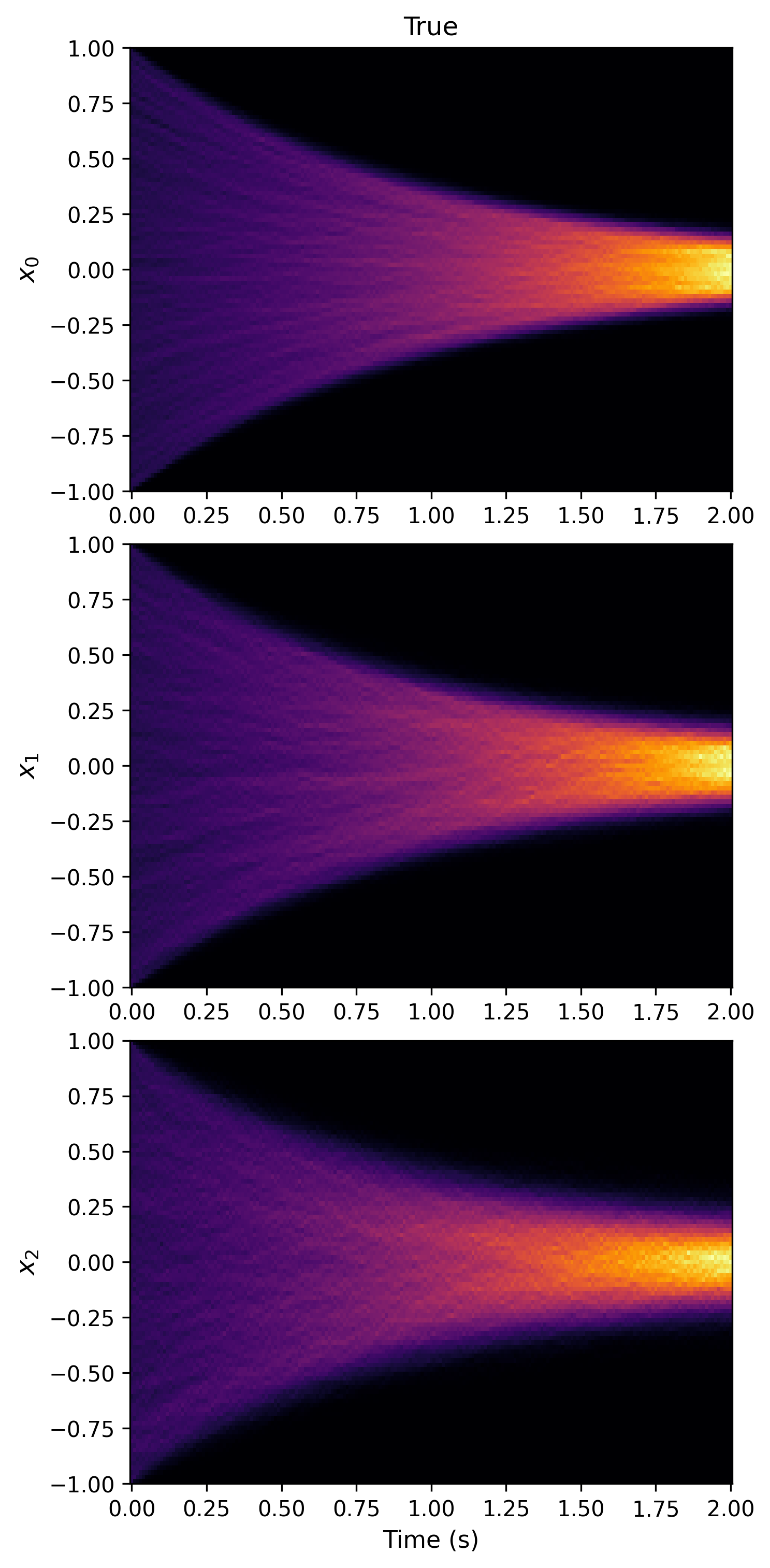}
        \end{subfigure}
        \begin{subfigure}[t]{0.35\textwidth}
            \centering
            \includegraphics[width=\textwidth]{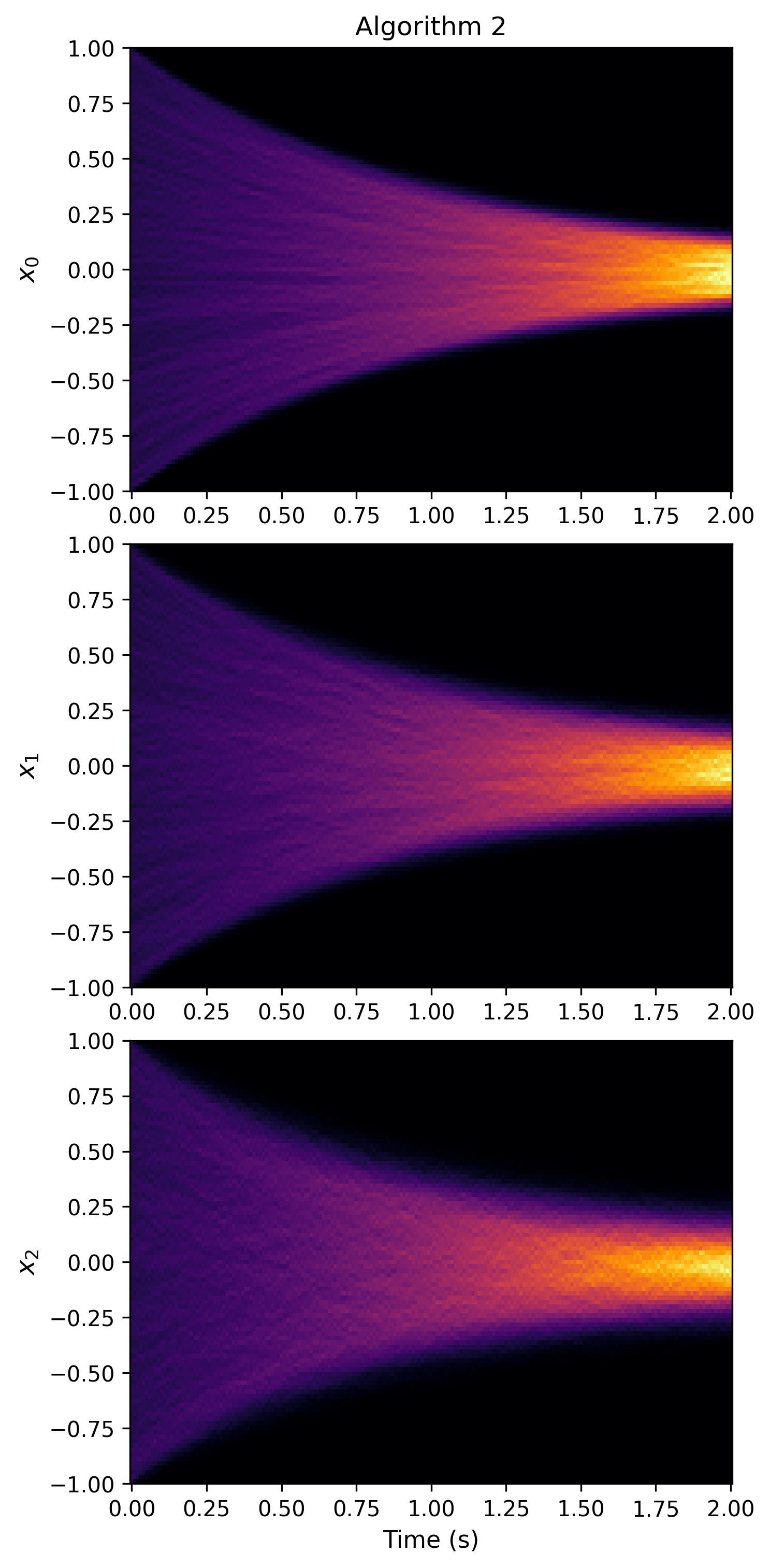}
        \end{subfigure}
        \begin{subfigure}[t]{0.35\textwidth}
            \centering
            \includegraphics[width=\textwidth]{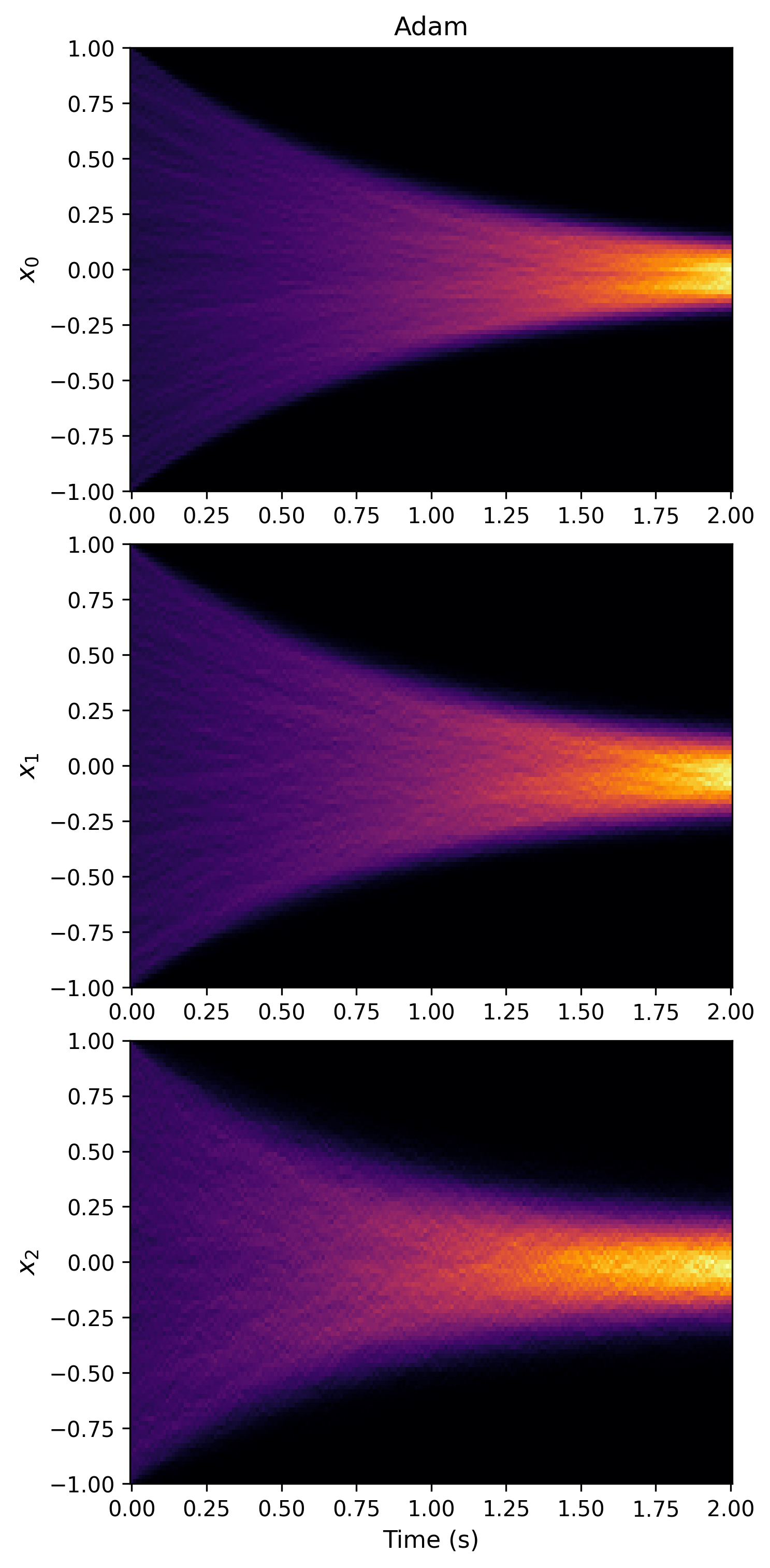}
        \end{subfigure}
    }
    \caption{Comparison of the trajectories generated using the true and trained functions for Experiment 4a: three-dimensional, symmetric diffusion.}
    \label{fig:Exp_4a_histogram}
\end{figure}

\begin{figure}[ht!]
    \centering
    \noindent\makebox[\textwidth][c]{%
        \begin{subfigure}[t]{0.35\textwidth}
            \centering
            \includegraphics[width=\textwidth]{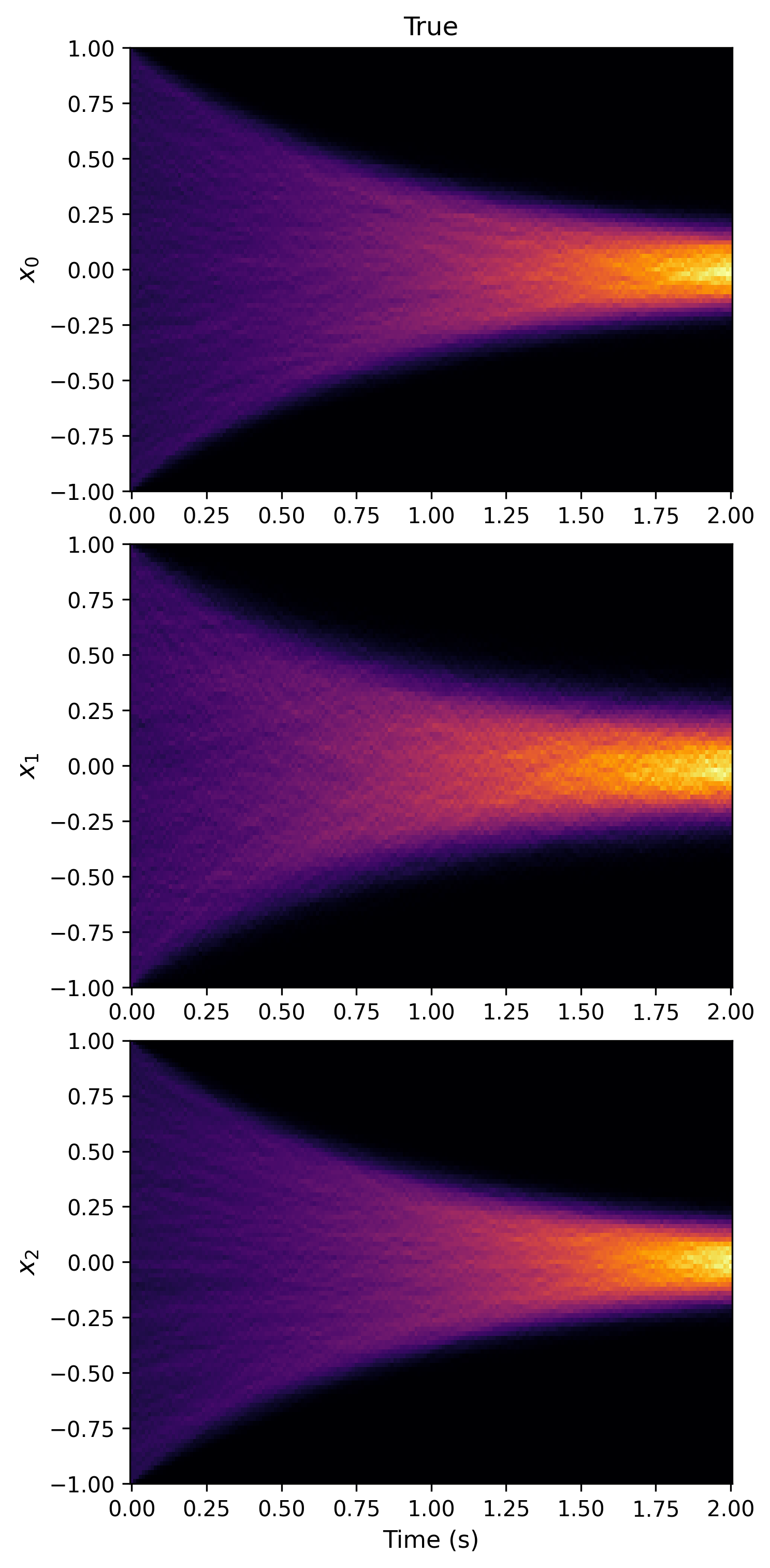}
        \end{subfigure}
        \begin{subfigure}[t]{0.35\textwidth}
            \centering
            \includegraphics[width=\textwidth]{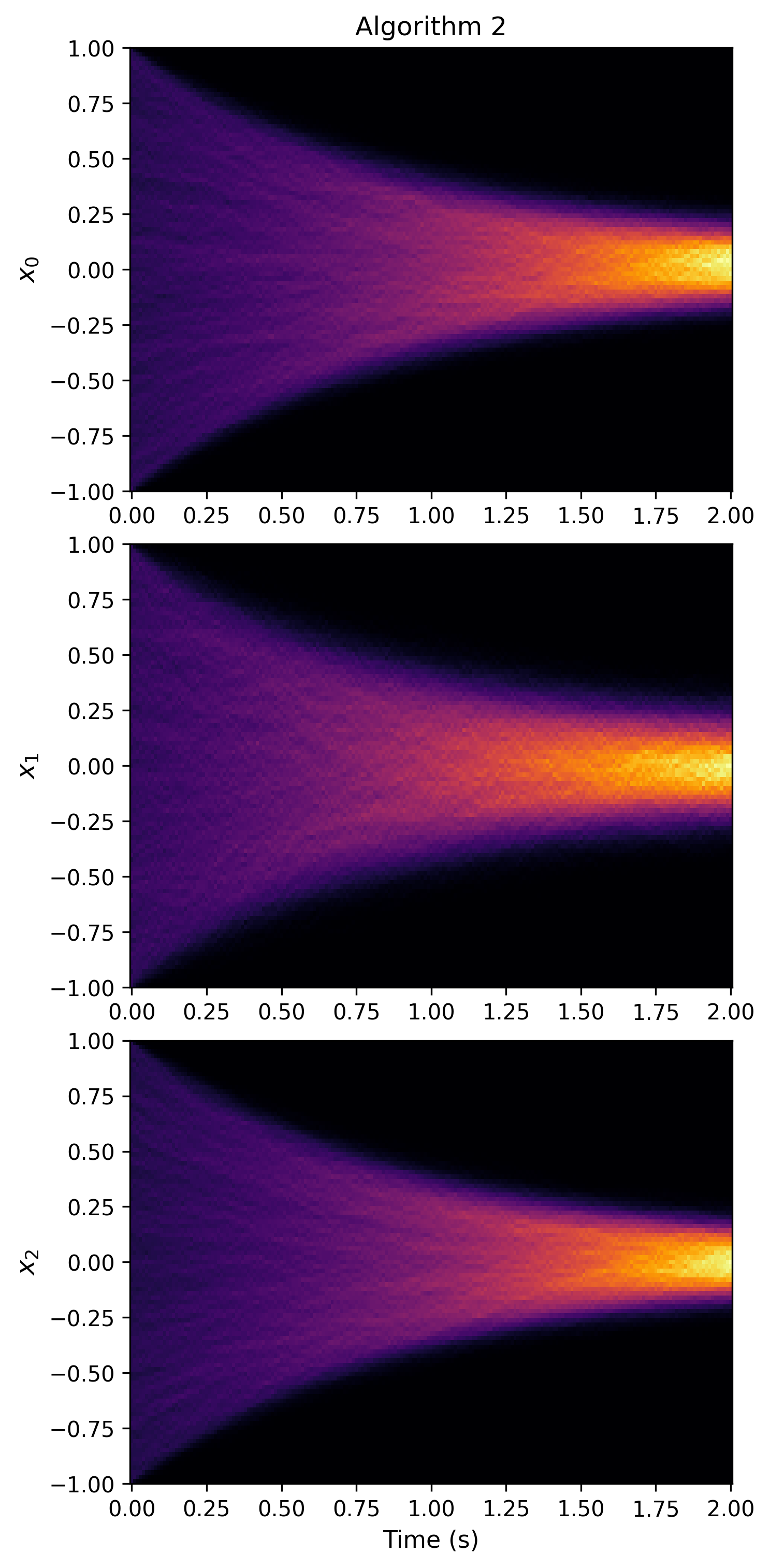}
        \end{subfigure}
        \begin{subfigure}[t]{0.35\textwidth}
            \centering
            \includegraphics[width=\textwidth]{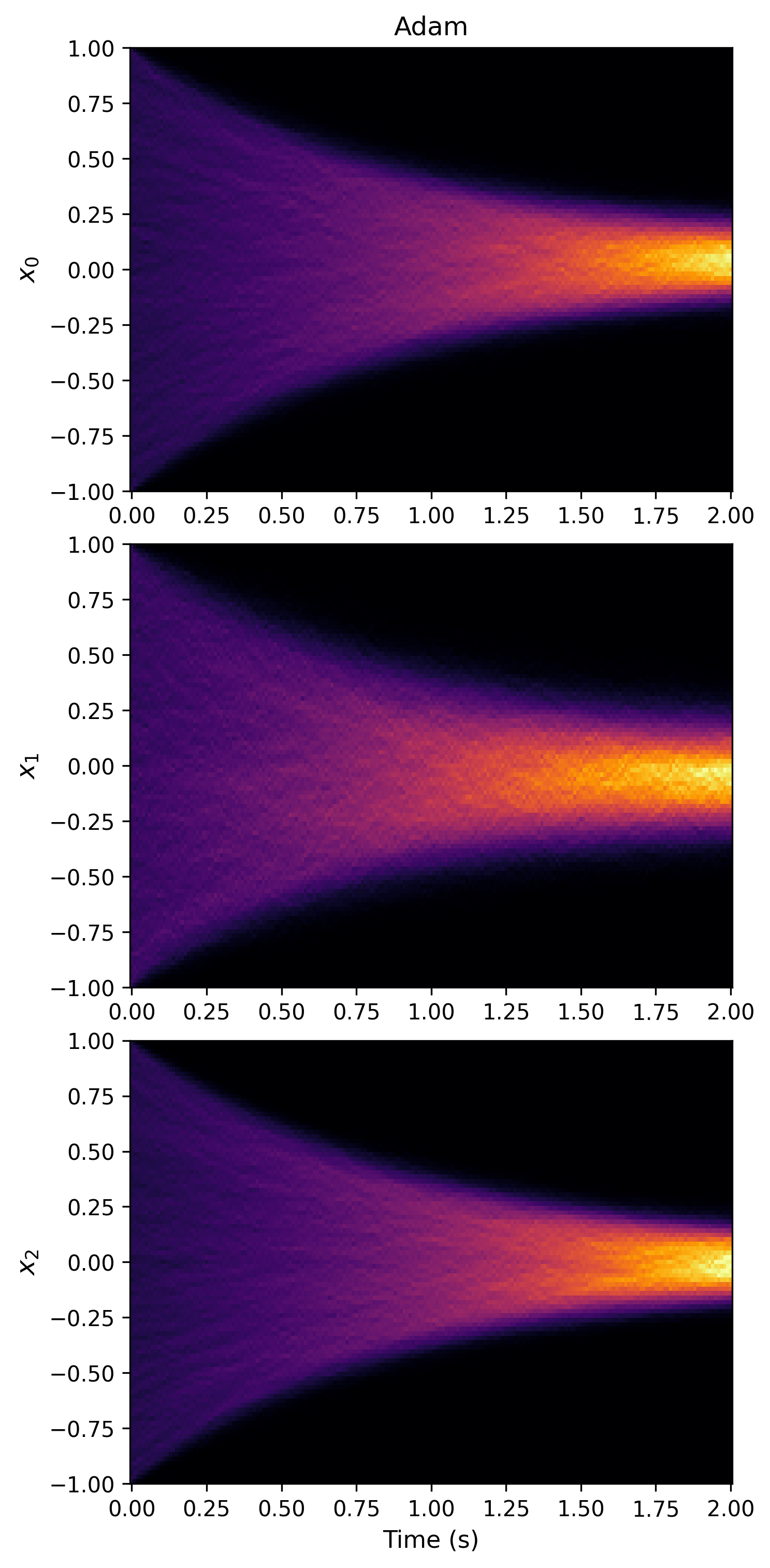}
        \end{subfigure}
    }
    \caption{Comparison of the trajectories generated using the true and trained functions for Experiment 4b: three-dimensional, lower-triangular diffusion.}
    \label{fig:Exp_4b_histogram}
\end{figure}

\begin{figure}[ht!]
    \centering
    \noindent\makebox[\textwidth][c]{%
        \begin{subfigure}[t]{0.35\textwidth}
            \centering
            \includegraphics[width=\textwidth]{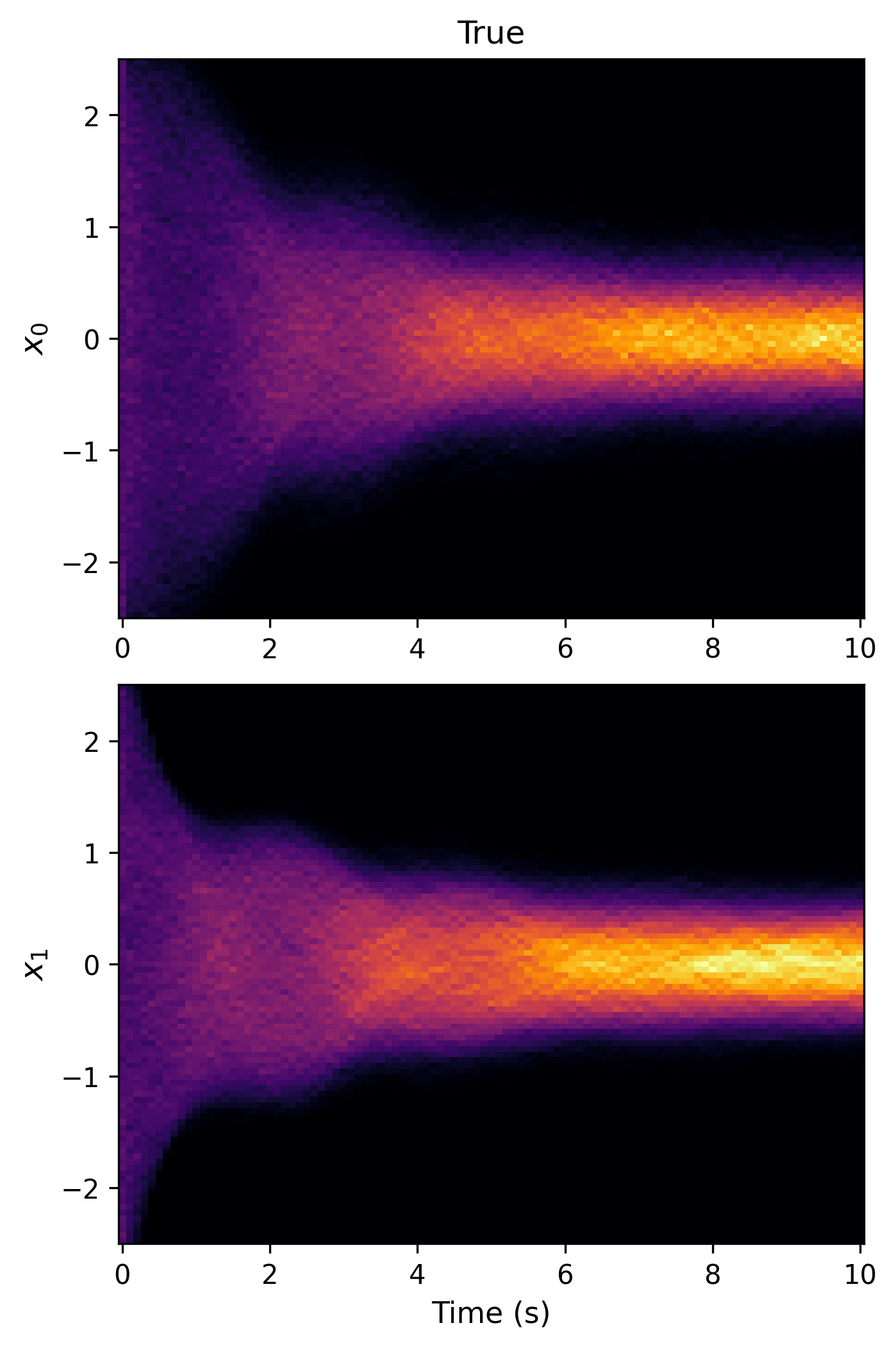}
        \end{subfigure}
        \begin{subfigure}[t]{0.35\textwidth}
            \centering
            \includegraphics[width=\textwidth]{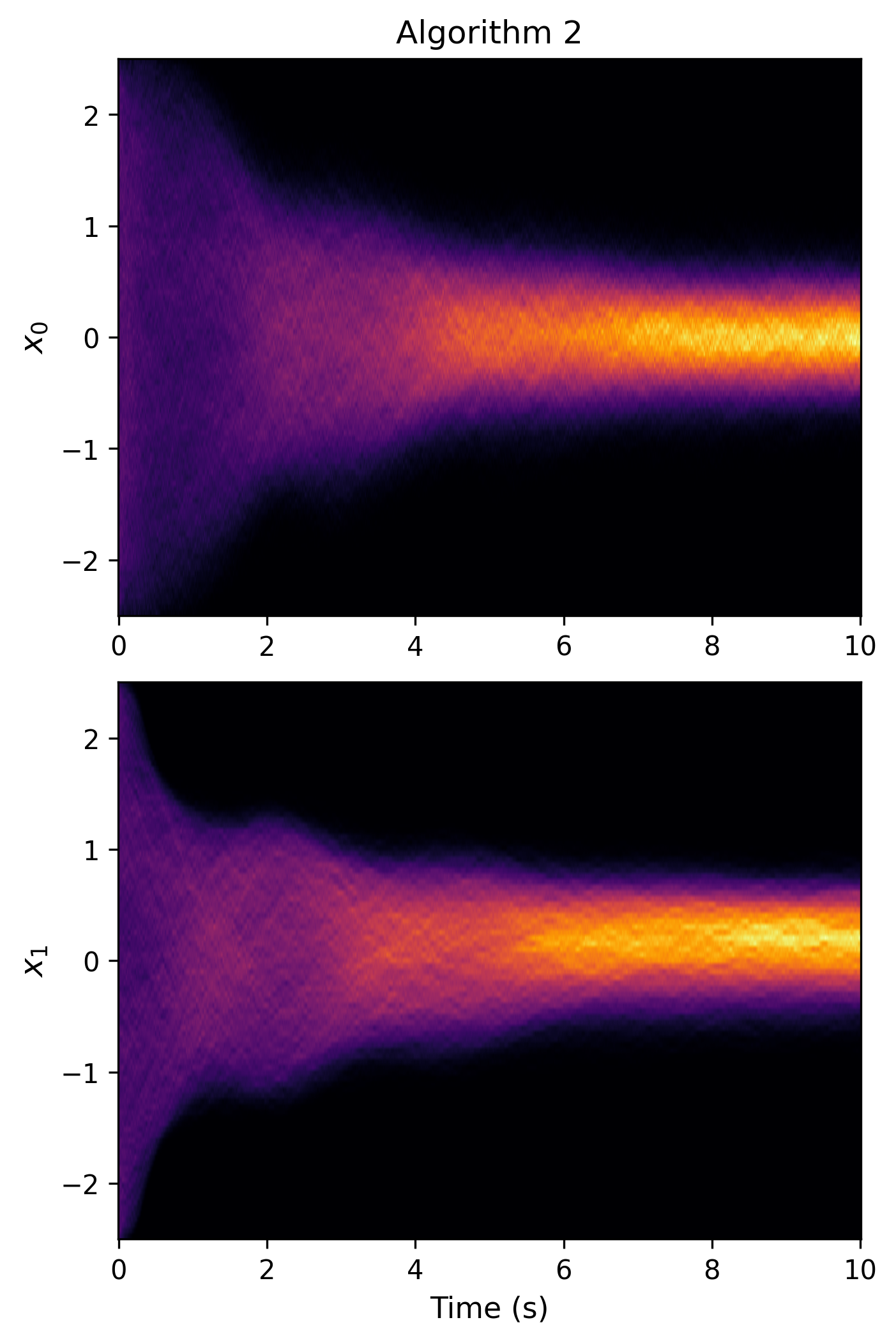}
        \end{subfigure}
        \begin{subfigure}[t]{0.35\textwidth}
            \centering
            \includegraphics[width=\textwidth]{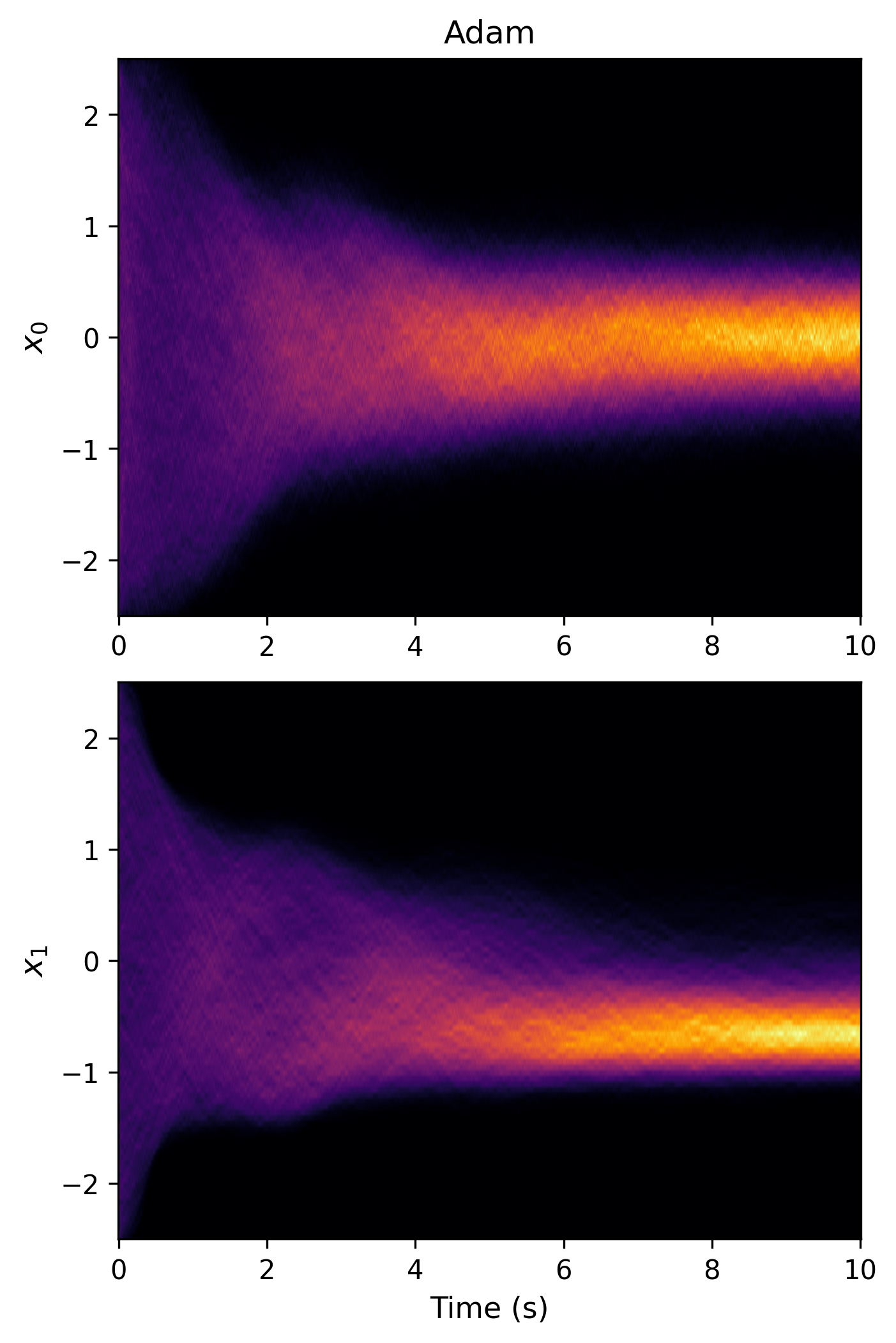}
        \end{subfigure}
    }
    \caption{Comparison of the trajectories generated using the true and trained functions for Experiment 5: underdamped Langevin.}
    \label{fig:Exp_5_histogram}
\end{figure}

\begin{figure}[ht!]
    \centering
    \noindent\makebox[\textwidth][c]{%
        \begin{subfigure}[t]{0.35\textwidth}
            \centering
            \includegraphics[width=\textwidth]{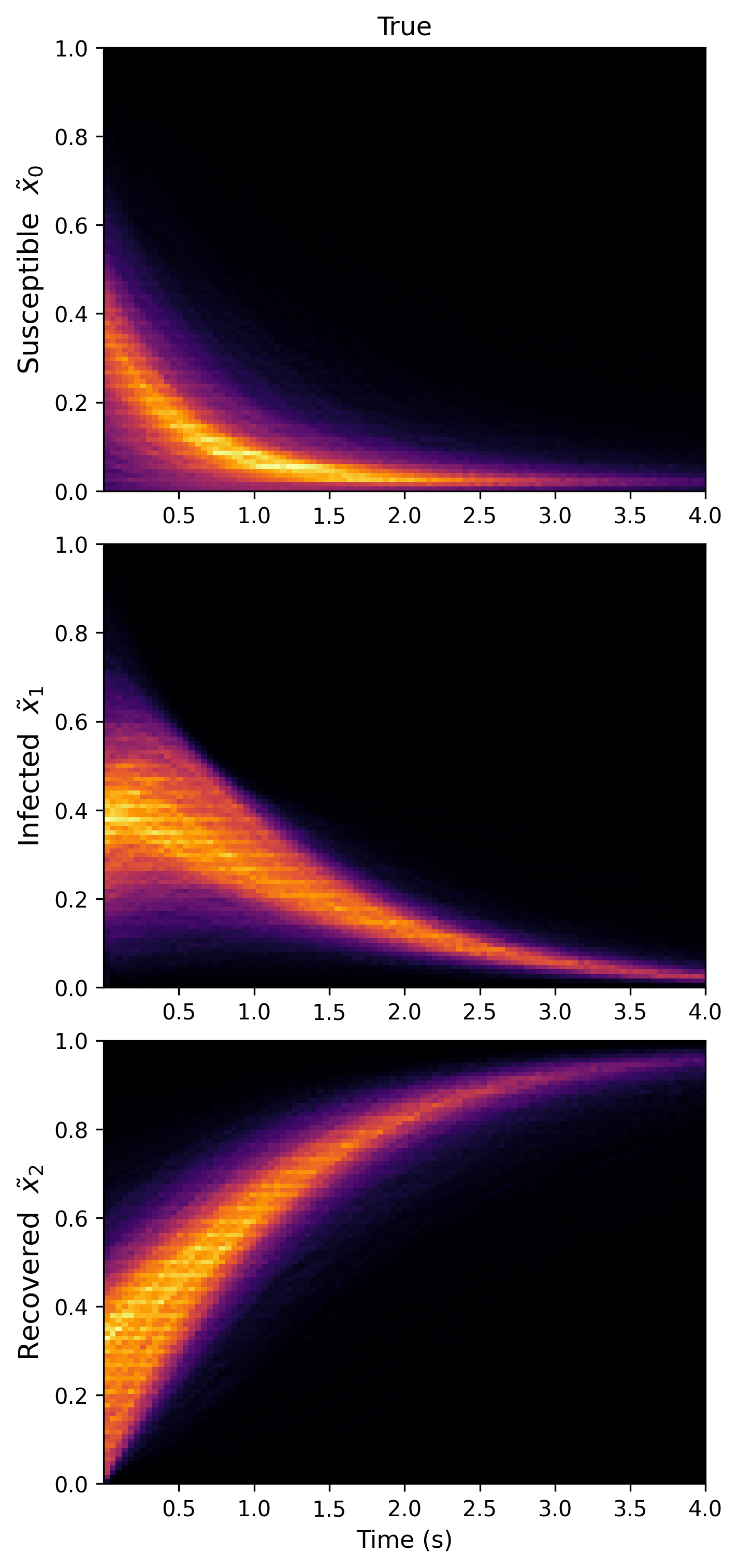}
        \end{subfigure}
        \begin{subfigure}[t]{0.35\textwidth}
            \centering
            \includegraphics[width=\textwidth]{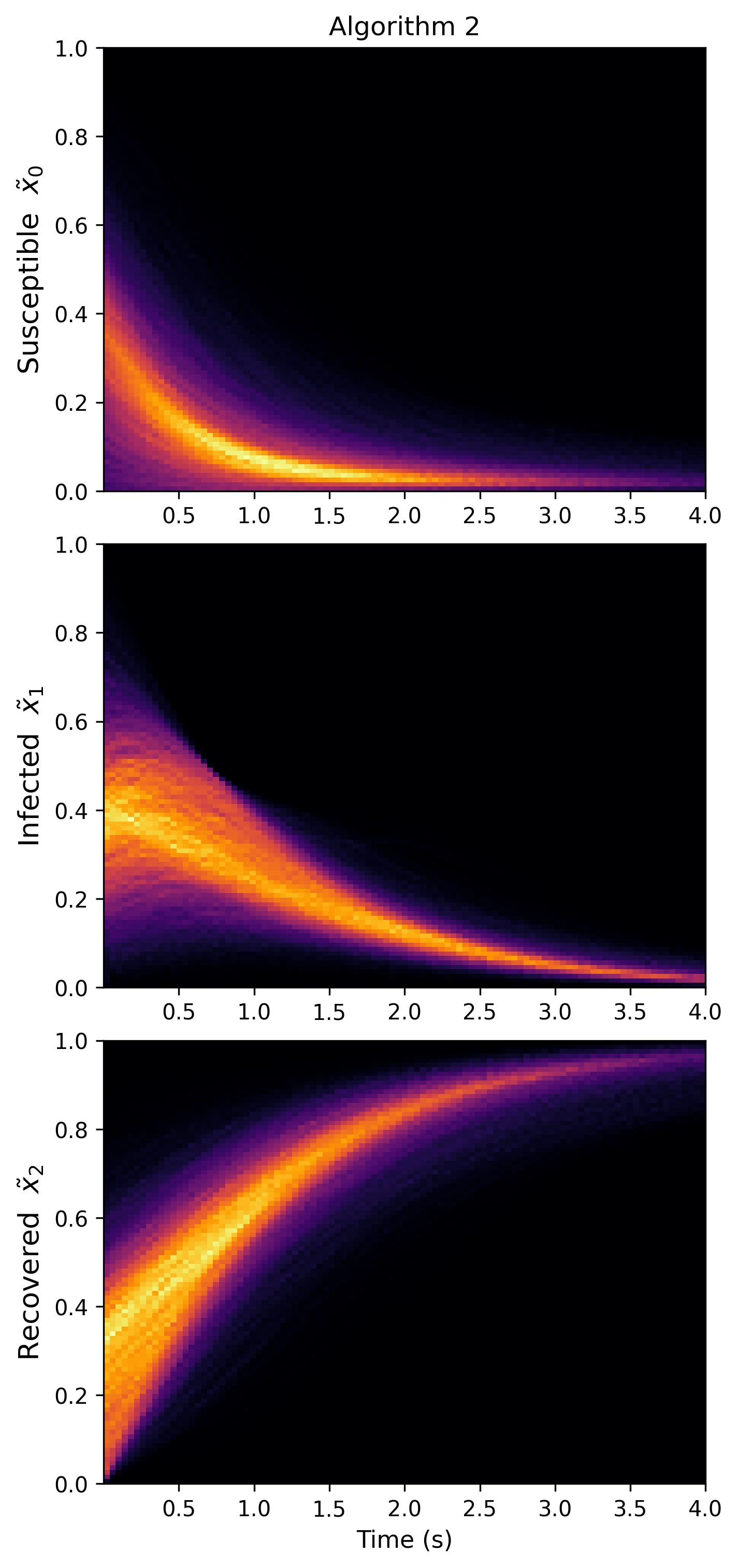}
        \end{subfigure}
        \begin{subfigure}[t]{0.35\textwidth}
            \centering
            \includegraphics[width=\textwidth]{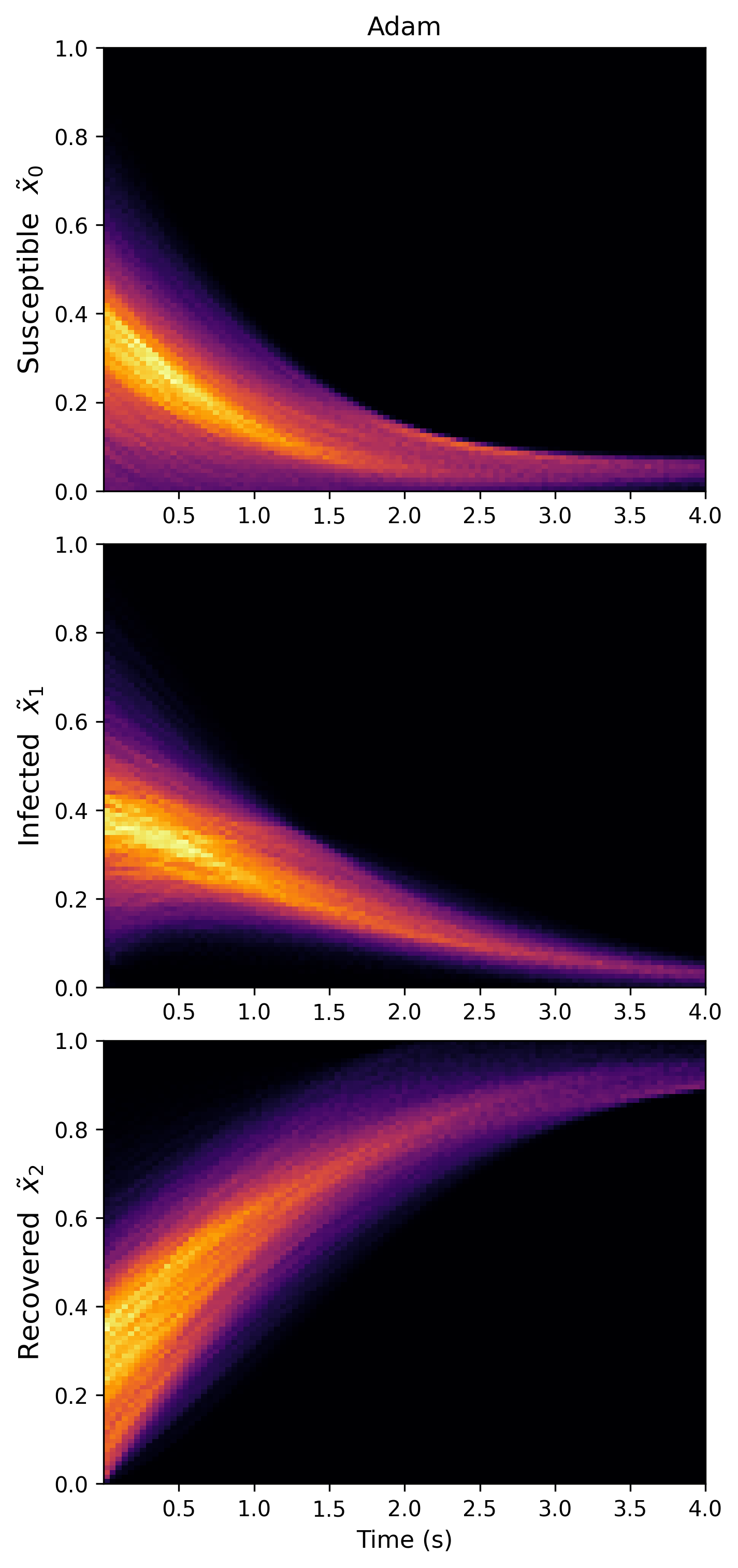}
        \end{subfigure}
    }
    \caption{Comparison of the trajectories generated using the true and trained functions for Experiment 6: susceptible, infected, recovered.}
    \label{fig:Exp_6_histogram}
\end{figure}

Figures \ref{fig:Exp_1_histogram} to \ref{fig:Exp_6_histogram} display the statistical distribution of 10,000 simulated trajectories generated using the ground truth and trained functions. Again the trained functions are the network configurations that resulted in the minimum validation loss in the training procedure.

The trajectories were initialized according to the initial distributions of the training datasets specified in Section \ref{sec:computational_experiments}. The simulation steps were performed using the EM scheme (\ref{eq:euler maruyama}), with a step size equal to the data generation step size. 

For Experiments 1 to 4 (Figures \ref{fig:Exp_1_histogram} to \ref{fig:Exp_4b_histogram}), networks trained using Algorithm \ref{alg:ARFF_SDE} and the Adam optimizer both generate trajectories almost indistinguishable from the true system dynamics - neither method is obviously superior. However, for Experiments 5 and 6 (Figures \ref{fig:Exp_5_histogram} and \ref{fig:Exp_6_histogram}), Algorithm \ref{alg:ARFF_SDE} provides the more accurate replication of the true system dynamics. 

\subsection{Comparison with Deep Neural Networks and ARFF without resampling}\label{sec:further_comparisions}

This section introduces and evaluates the performance impact of modifications made to Algorithm \ref{alg:ARFF_SDE} and Adam optimization. 

First, the role of the resampling step within Algorithm \ref{alg:ARFF} is considered. Research in \cite{kammonen2024adaptiverandomfourierfeatures} identifies performance improvements when implementing a resampling step within ARFF training. This section investigates how the omission of the resampling step impacts loss minimization. 

Second, the impact of network depth on Adam optimization is explored. Gradient descent-based methods typically achieve superior performance with deep compared to shallow network architectures \cite{krizhevsky2012imagenet}. It is worth comparing Algorithm \ref{alg:ARFF_SDE} with more the more widely regarded deep network approach.

Testing is conducted on the dataset from Experiment 3a. The non-sampling variant of Algorithm \ref{alg:ARFF_SDE} maintains the same network parameters and architecture from Experiment 3a: $\lambda = 0.002$, $\delta = 0.1$, $\gamma = 1$ and $K = 2^7$. The deep network architecture consists of two layers of size $K = 2^6$ (same total number of nodes) and employs ReLU activation functions. The hyperparameters are carried over from the shallow implementation in Experiment 3a: a learning rate of 0.001 and a batch size of 32.

\begin{figure}[ht!]
    \centering
    \includegraphics[width=0.5\linewidth]{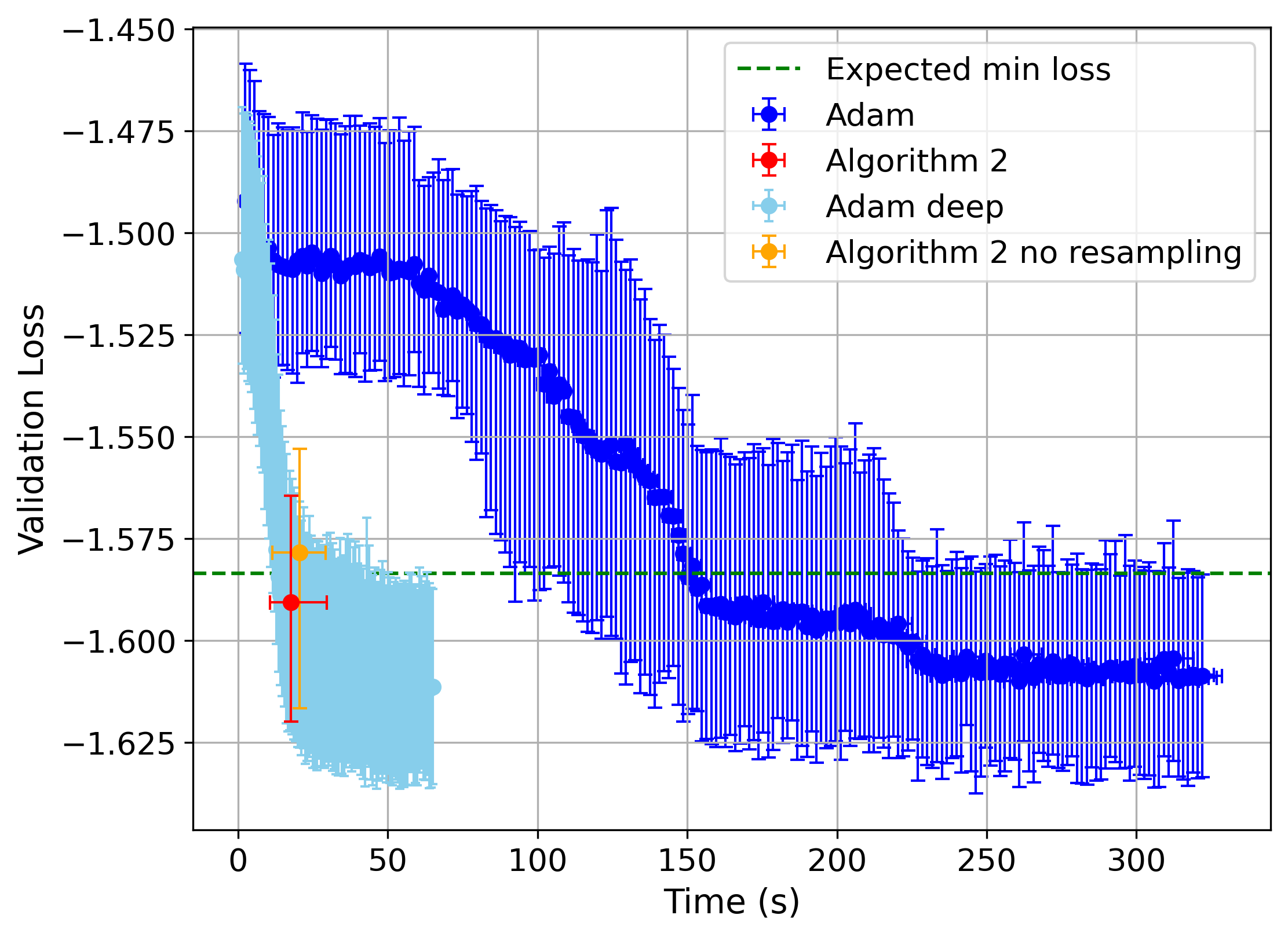}
    \caption{Mean validation loss vs mean training time across 10 runs of Experiment 3a* with error bars indicating $\pm$ one standard deviation. For the Adam optimizer, values for each epoch are plotted.}
    \label{fig:loss_v_time_deep}
\end{figure}

The results are summarized in Table \ref{tab:results_deep} and Figure \ref{fig:loss_v_time_deep}.

\begin{table}[ht!]
\centering
\begin{adjustbox}{max width=\textwidth, center}
\begin{tabular}{|c|c|cccc|cccc|}
\hline
\multirow{2}{*}{Exp.} & \multirow{2}{*}{\begin{tabular}[c]{@{}c@{}}Expected \\ min loss\end{tabular}} & \multicolumn{4}{c|}{Min validation loss}                                                                                                                                                                                                                                                                                 & \multicolumn{4}{c|}{Training time (s)}                                                                                                                                                                                                                                                                                   \\ \cline{3-10} 
                      &                                                                               & \multicolumn{1}{c|}{\begin{tabular}[c]{@{}c@{}}\footnotesize{Algorithm 2}\\ \footnotesize{(resample)}\end{tabular}} & \multicolumn{1}{c|}{\begin{tabular}[c]{@{}c@{}}\footnotesize{Algorithm 2}\\ \footnotesize{(no resample)}\end{tabular}} & \multicolumn{1}{c|}{\begin{tabular}[c]{@{}c@{}}Adam\\ \footnotesize{(shallow)}\end{tabular}} & \begin{tabular}[c]{@{}c@{}}Adam\\ (deep)\end{tabular} & \multicolumn{1}{c|}{\begin{tabular}[c]{@{}c@{}}\footnotesize{Algorithm 2}\\ \footnotesize{(resample)}\end{tabular}} & \multicolumn{1}{c|}{\begin{tabular}[c]{@{}c@{}}\footnotesize{Algorithm 2}\\ \footnotesize{(no resample)}\end{tabular}} & \multicolumn{1}{c|}{\begin{tabular}[c]{@{}c@{}}Adam\\ \footnotesize{(shallow)}\end{tabular}} & \begin{tabular}[c]{@{}c@{}}Adam\\ (deep)\end{tabular} \\ \hline
3a*                   & 0.025636                                                                      & \multicolumn{1}{c|}{-1.5906}                                                          & \multicolumn{1}{c|}{-1.5784}                                                             & \multicolumn{1}{c|}{-1.5121}                                                  & -1.6135                                               & \multicolumn{1}{c|}{17.683}                                                           & \multicolumn{1}{c|}{20.453}                                                              & \multicolumn{1}{c|}{18.946}                                                   & 34.354                                                \\ \hline
\end{tabular}
\end{adjustbox}
\caption{Comparison of the mean minimum validation loss and mean training time across 10 runs of Experiment 3a*. For Adam, the minimum validation loss is defined as the lowest value before stagnation, which is characterized as five consecutive epochs with no decrease in the moving average.}
\label{tab:results_deep}
\end{table}

Table \ref{tab:results_deep} and Figure \ref{fig:loss_v_time_deep} show that the non-resampling variant of Algorithm \ref{alg:ARFF_SDE} marginally underperforms in both loss reduction and training time compared to its resampling counterpart, consistent with expectations from \cite{kammonen2024adaptiverandomfourierfeatures}. The deep network implementation, significantly reduces training time for Adam optimization. Despite this, Algorithm \ref{alg:ARFF_SDE} still reaches the same loss for the same elapsed time. However, continued training for the deep network beyond this point exceeds the loss achieved by Algorithm \ref{alg:ARFF_SDE}.

\vspace{1em}

Another modification to Algorithm \ref{alg:ARFF_SDE} was considered. Prior research by Dridi et al. \cite{dridi2021learningstochasticdynamicalsystems} identified SDE examples that exhibit accuracy reductions when learning the drift independently of the diffusion, specifically when minimizing the problem (\ref{eq:drift_minimisation_problem}). To address this potential limitation, an additional drift training step, conditioned on the previously trained diffusion $\Sigma_{\theta'}$, was appended to Algorithm \ref{alg:ARFF_SDE},

\begin{equation*}
    \min _{\Sigma_{\theta'}^{-1/2} f_{\theta^*} \in \mathcal{N}_K}\left\{ \mathbb{E}_{\hat{\rho}} \left[ | \Sigma_{\theta'}(x_0)^{-1/2} h^{-1}(x_1 - x_0) - \Sigma_{\theta'}(x_0)^{-1/2}f_{\theta^*}(x_0) |^2\right] + \lambda \sum_{k=1}^K |\hat{\beta}_k|^2 \right\}.
\end{equation*}

Here the function $g(x_0) = \Sigma_{\theta'}(x_0)^{-1/2} f_{\theta^*}(x_0)$ is learned and the drift is recovered as $f_{\theta^*}(x_0) = \Sigma_{\theta'}(x_0)^{1/2} g(x_0)$. 

This approach was evaluated across multiple examples, including those that appear in \cite{dridi2021learningstochasticdynamicalsystems}. Testing revealed no loss or training time improvements compared to the non-modified Algorithm \ref{alg:ARFF_SDE}. 

\section{Summary}\label{sec:summary}

This work introduces a novel algorithm based on ARFF for learning SDEs and evaluates its performance against Adam-based neural network training. The aim is to assess whether the benefits of RFF translate effectively into the domain of SDE learning.

The experimental results reveal three key findings that establish potential advantages of ARFF-based SDE learning. First, Algorithm \ref{alg:ARFF_SDE} demonstrates superior computational efficiency, consistently achieving loss thresholds faster than Adam-based training across equivalent network architectures. Second, Algorithm \ref{alg:ARFF_SDE} provides more accurate approximations of the underlying stochastic dynamics across almost all problems considered. Third, Algorithm \ref{alg:ARFF_SDE} maintains competitive performance even when compared against deep Adam-optimized networks with greater representational capacity.

Ablation studies also confirm two important methodological insights: the resampling step provides measurable performance improvements, consistent with findings in \cite{kammonen2024adaptiverandomfourierfeatures}, while relearning the drift function conditioned on the trained diffusion yields no meaningful benefits. 

Whilst these findings present ARFF as a compelling alternative for data-driven SDE learning, the conclusions are drawn from a narrow set of computational experiments. Specifically, the evaluations are exclusively derived from CPU-based training and low-dimensional datasets (up to 3D), which limits the generalization of the timing and performance conclusions. Testing on higher-dimensional datasets and GPU-based training will be critical for establishing the broader practical utility of ARFF-based SDE learning methods.

At present, ARFF-based training is restricted to single hidden layer architectures, preventing direct incorporation of deep network advantages. A promising direction for future research involves developing hybrid architectures where Algorithm \ref{alg:ARFF_SDE} provides rapid initialization of the first layer, followed by gradient-based fine-tuning of the deeper layers. This approach could harness ARFF's computational efficiency for initial feature learning while leveraging the representational power of deep networks for complex pattern capture.

This research establishes Algorithm \ref{alg:ARFF_SDE} as a promising alternative for SDE learning, warranting further investigation of ARFF-based methods within stochastic modeling applications.
\clearpage
\bibliographystyle{plain}   
\bibliography{references}
\end{document}